\documentclass{bmvc2k}
\usepackage[T1]{fontenc}
\usepackage[latin9]{inputenc}
\usepackage{array}
\usepackage{wrapfig}
\usepackage{units}
\usepackage{multirow}
\usepackage{graphicx}

\makeatletter

\providecommand{\tabularnewline}{\\}
\newcommand{\lyxdot}{.}

\addauthor{Jan Hosang}{http://mpi-inf.mpg.de/~jhosang}{1}

\addauthor{Rodrigo Benenson}{http://mpi-inf.mpg.de/~benenson}{1}

\addauthor{Bernt Schiele}{http://mpi-inf.mpg.de/~schiele}{1}

\addinstitution{
MPI Informatics\\
Saarbr\"{u}cken, Germany
}

\runninghead{J. Hosang et al.}{How good are detection proposals, really?}

\title{Selecting selective search}

\usepackage{tweaklist}
\renewcommand{\enumhook}{
}

\@ifundefined{showcaptionsetup}{}{%
 \PassOptionsToPackage{caption=false}{subfig}}
\usepackage{subfig}
\makeatother

\begin{document}

\title{How good are detection proposals, really?}

\maketitle
\vspace{-1em}

\begin{abstract}
Current top performing Pascal VOC object detectors employ detection
proposals to guide the search for objects thereby avoiding exhaustive
sliding window search across images. Despite the popularity of detection
proposals, it is unclear which trade-offs are made when using them
during object detection. We provide an in depth analysis of ten object
proposal methods along with four baselines regarding ground truth
annotation recall (on Pascal VOC 2007 and ImageNet 2013), repeatability,
and impact on DPM detector performance. Our findings show common weaknesses
of existing methods, and provide insights to choose the most adequate
method for different settings.\vspace{-0em}

\end{abstract}

\section{\label{sec:Introduction}Introduction}

\begin{wrapfigure}{r}{0.41\textwidth}%
\begin{centering}
\vspace{-4.5em}
\includegraphics[bb=0bp 0bp 365bp 190bp,clip,width=0.4\textwidth]{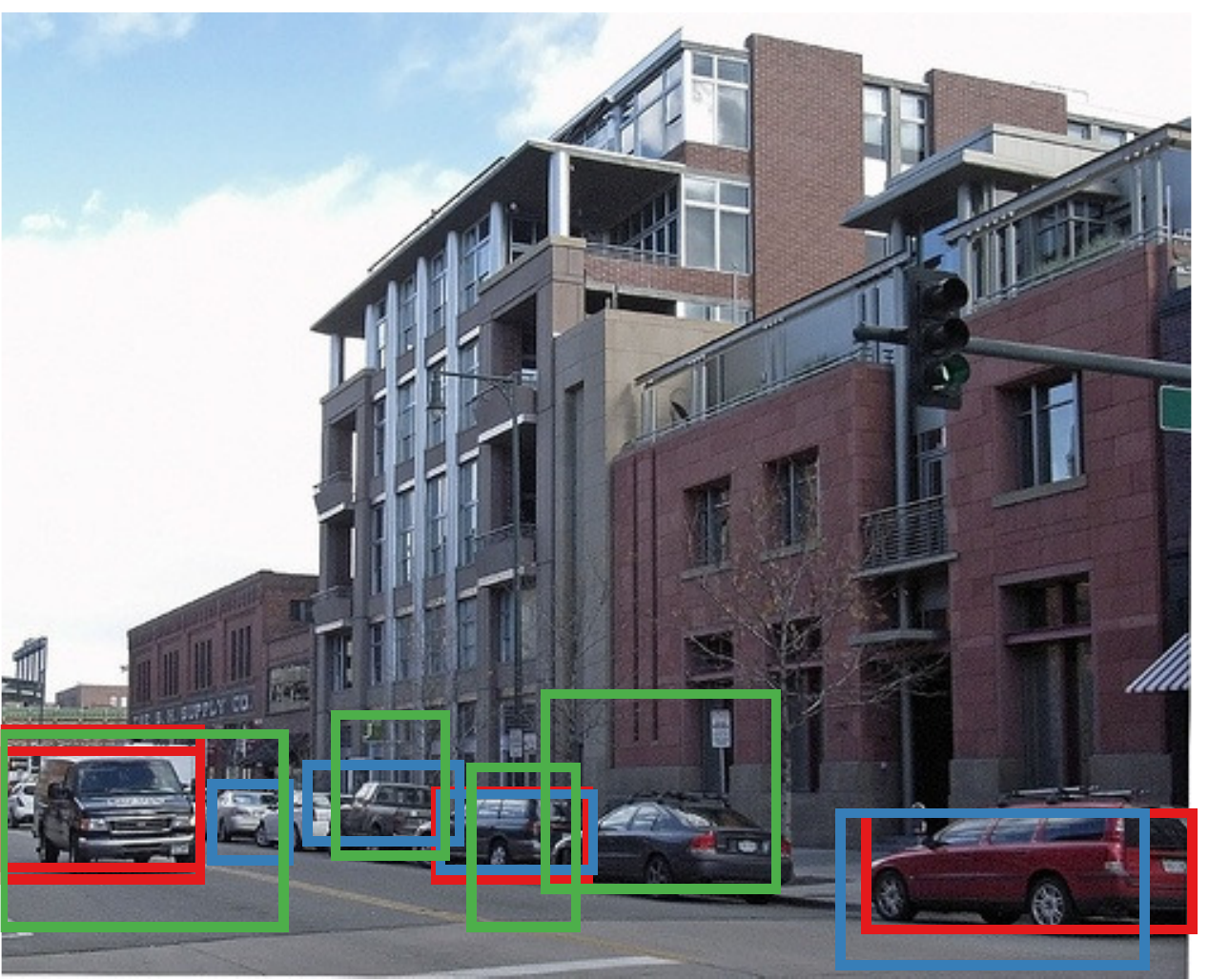}
\par\end{centering}

\protect\caption{\label{fig:teaser}How to evaluate the quality of such detection proposals?}
\vspace{-1.5em}
\end{wrapfigure}%
Object detection is traditionally formulated as a classification
problem in the well known ``sliding window'' paradigm where the
classifier is evaluated over an exhaustive list of positions, scales,
and aspect ratios. Steadily increasing the sophistication of the core
classifier has led to increased detector performance \cite{Felzenszwalb2010Pami,Wang2013Iccv,Girshick2014Cvpr}.

A typical sliding window detector requires $\sim10^{6}$ classifier
evaluations per image.  One approach to overcome the tension between
computational tractability and high detection quality, is the notion
of ``detection proposals'' (sometimes called ``objectness'' or
``selective search''). Under the assumption that all objects of
interest share common visual properties that distinguish them from
the background, one can train a method that, given an input image,
outputs a set of detection window proposals that are likely to contain
the objects. If high recall can be reached with $\sim10^{4}$ or less
windows, significant speed-ups can be achieved, enabling the use of
even more sophisticated classifiers. 

Besides the potential to improve speed, the use of detection proposals
changes the data distribution that the classifier handles, thus also
potentially improving detection quality (reduce false positives).
It is interesting to note that the current two top performing detection
methods for Pascal VOC \cite{Everingham2014Ijcv} use detection proposals
\cite{Wang2013Iccv,Girshick2014Cvpr}.\\
\label{sub:Contributions}\textbf{Contributions}$\quad$In contrast
to previous work that typically performs a partial evaluation when
introducing a novel method, this paper aims to revisit existing work
on detection proposals and compare all publicly available methods
in a common framework to better understand their benefits and limitations.
We attempt to provide the necessary insights to understand when to
choose which method.

Our contributions are the following. (1) We review existing detection
proposal methods (\S\ref{sub:Related-work}) and compare them in
a unified evaluation. Our evaluation aims to provide an unbiased
view of existing methods. (2) We introduce the notion of repeatability
for detection proposals, discuss why it matters when considering detection
proposals for object detection, and compare repeatability of existing
methods (\S\ref{sec:Repeatibility}). (3) We evaluate overlap with
annotated ground truth objects on the Pascal VOC 2007 test set, and
for the first time, over the larger and more diverse ImageNet 2013
validation set (\S\ref{sec:Proposal-recall}). This latter experiment
aims at detecting possible biases towards the Pascal VOC objects categories.
(4) Finally we evaluate the influence on detector performance using
different proposal methods (\S\ref{sec:Using-detection-proposals}).
(5) All bounding boxes from our experiments and the evaluation scripts
will be released with the paper. The results presented summarise more
than 500 experiments over different data sets, totalling to more
than 2.5 months of CPU computation.

\section{\label{sub:Related-work}Detection proposal methods}

Interestingly, the spirit of detection proposals is similar to the
idea of interest point detection \cite{Tuytelaars2008FoundationsandTrends,Mikolajczyk2005Ijcv}.
Interest points were proposed at a time when computing feature descriptors
densely was computationally too expensive and some selection of interest
points was important. Feature descriptors computed around such interest
points were used successfully for classification, retrieval, and detection.
Today however, with the increase of computing power, it is standard
practice to use dense feature extraction instead \cite{Tuytelaars2010Cvpr}.
Thus we may ask: do object proposals help detection quality, or are
they just a transition technology until we have sufficient computing
power?

Detection proposal methods are based on low-level image features to
generate candidate windows. One can reinterpret this process as a
discriminative one; given the low-level features the method quickly
decides whether a window should be considered for detection or not.
In this sense detection proposal methods are related to cascade methods
\cite{Viola2004Ijvc,Harzallah2009Iccv,Dollar2012Eccv}, which use
a fast (but inaccurate) classifier to discard the vast majority of
unpromising proposals. Although traditionally used for class specific
detection, cascade methods also apply to sets of categories \cite{Torralba2007Pami},
and in principle can be applied to a very large set of categories
(see \texttt{Bing} method below).

Since their introduction \cite{Gu2009Cvpr,Alexe2010Cvpr} multiple
works have explored the idea of generating detection proposals. We
briefly review all methods we are aware of in chronological order.

\noindent \texttt{\textbf{gPbUCM}}\texttt{\,}\cite{Gu2009Cvpr} is
a leading method for grouping pixels into objects. Given an input
image it hierarchically groups segments and  uses these segments
(or bounding boxes around them) as object detection candidates \cite{Gu2009Cvpr,Gu2012Eccv}.
This paper evaluates the bounding box proposals since not all methods
generate a segmentation of the proposals.\\
\texttt{\textbf{Objectness}}\texttt{\,}\cite{Alexe2010Cvpr,Alexe2012Pami}
is a term that refers to a measure of how likely a detection window
contains an object (of any category). \cite{Alexe2012Pami} estimates
this score based on a combination of multiple cues such as saliency,
colour contrast, edge density, location and size statistics, and how
much such windows overlap with superpixel segments.  \\
\texttt{\textbf{CPMC}}\texttt{\,}\cite{Carreira2010Cvpr,Carreira2012Pami}:
Most methods are built upon some form of hierarchical segmentation.
\texttt{CPMC} avoids this by generating a set of overlapping segments.
Each proposal segment is the solution of a binary segmentation problem,
initialised with diverse seeds. Up to $10^{4}$ segments are generated
per image, which are subsequently ranked by objectness using a trained
regressor. \\
\texttt{\textbf{Endres2010}}\texttt{\,}\cite{Endres2010Eccv,Endres2014Pami}:
The approach from \cite{Endres2014Pami} mixes a large set of cues.
It uses a hierarchical segmentation, a learned regressor to estimate
boundaries between surfaces with different orientations, graph cuts
with different seeds, and parameters to generate diverse segments
(similar to \texttt{CPMC}). It also learns a ranking function for
the proposed segments. \\
\texttt{\textbf{SelectiveSearch}}\texttt{\,}\cite{Sande2011Iccv,Uijlings2013Ijcv}
is a method where no parameters are learned. The authors carefully
engineer features and score functions that greedily merge low-level
superpixels. The authors  obtain state of the art object detections
on Pascal VOC and ILSVRC2011.\\
\texttt{\textbf{Rahtu2011}}\texttt{\,}\cite{Rahtu2011Iccv} revisits
\cite{Alexe2012Pami} and obtains a significant improvement by proposing
new objectness cues and a different method to learn how to combine
them.\\
\texttt{\textbf{RandomizedPrim's}}\texttt{\,}\cite{Manen2013Iccv}
is similar to \texttt{SelectiveSearch} in terms of features to merge
low-level superpixels. The key difference is that the weights of the
merging function are learned and the merging process is randomised.\\
\texttt{\textbf{Bing}}\texttt{\,}\cite{Cheng2014Cvpr} is  one of
the only two methods that are not based on segmentation. A simple
linear classifier over edge features is trained and applied in a sliding
window manner. Using adequate approximations a very fast class agnostic
detector is obtained ($1\,\nicefrac{\mbox{ms}}{\mbox{image}}$ on
CPU). \\
\texttt{\textbf{MCG}}\texttt{\,}\cite{Arbelaez2014Cvpr} is one of
the most recent methods combining \texttt{gPbUCM} and \texttt{CPMC}.
The authors propose an improved multi-scale hierarchical segmentation
(similar to \texttt{gPbUCM}), a new strategy to generate proposals
by merging up to 4 segments, and (similar to \texttt{CPMC}) a new
ranking procedure to select the final detection proposals.\\
\texttt{\textbf{Rantalankila2014}}\texttt{\,}\cite{Rantalankila2014Cvpr},
also recently proposed, combines \texttt{SelectiveSearch} and \texttt{CPMC}.
Starting from low-level superpixels the authors propose a merging
strategy, similar to \texttt{SelectiveSearch} but using different
features. These merged segments are then used as seeds for a \texttt{CPMC}-like
process to generate larger segments.\\
\texttt{\textbf{Rigor}}\texttt{\,}\cite{Humayun2014Cvpr} is a variant
of \texttt{CPMC} obtaining higher quality by using different low-level
superpixels and different features for merging segments. They also
obtain higher speed by minimizing redundant computations. \\
\texttt{\textbf{EdgeBoxes\,}}\cite{Zitnick2014Eccv} is similar in
spirit to \texttt{Bing}, a scoring function is evaluated in a sliding
window fashion. This method uses object boundaries estimates (obtained
via structured decision forests) as feature for the scoring. Interestingly,
the authors propose tuning parameters to optimize recall at a desired
overlap threshold (see section \ref{sec:Proposal-recall}).

Recently, Kang et al.~\cite{Kang2014Pami} proposed a ``data-driven
objectness'' approach. Although it showed promising results for their
indoor application scenario, the method seems of limited applicability
to other datasets, and thus we do not consider it in our evaluation. 

The majority of the methods is based on some low-level segmentation:
five use \cite{Felzenszwalb2004IJCV}, two use a variant of the \texttt{gPbUCM}
segmentation method, and\texttt{ Endres2010} uses its own custom
low-level segmentation (also used by \texttt{Rigor}). Only three methods
work without computing low-level segments (\texttt{CPMC}, \texttt{Bing},
\texttt{EdgeBoxes}). At the time of writing no code is available
for \texttt{gPbUCM} and \texttt{Rigor}%
\footnote{Recently published at CVPR 2014.%
}, thus we do not consider these in our evaluation. In total seven
out of the ten methods evaluated in this paper use some low-level
segmentation, five out of these eight use \cite{Felzenszwalb2004IJCV}.

It should also be noted that since \cite{Alexe2010Cvpr}, all related
work evaluates the quality of detection proposal methods based on
the overlap with ground truth annotations of the Pascal VOC dataset
(2007 or 2010) \cite{Everingham2014Ijcv}. Although a relevant metric,
its repeated use opens the doors for over-fitting to this specific
dataset. In this paper we also consider other evaluations.\\
\textbf{Number of windows}$\quad$The different methods listed above
provide different numbers of detection candidates. Some provide rather
few windows ($\sim10^{2}$), others provide a large number ($\sim10^{5}$).
Some methods do provide sorted (scored) windows, others do not. Having
more windows increases the chance for high recall, thus for each method
in all experiments we do our best effort to generate a similar average
number of candidate windows per image. See supplementary material
for details on how this is achieved for each method.

\subsection{Baselines}

Besides the above methods, we also consider a set of baselines that
serve as quantitative reference points. All of the above candidate
methods and the following baselines are class independent.\\
\texttt{\textbf{Uniform}}: To generate detection proposals, we uniformly
sample the bounding box parameters centre position, square root area,
and log aspect ratio. We estimate the range of these parameters on
the Pascal VOC 2007 training set after discarding 0.5\% of the smallest
and largest values, so that our estimated distribution covers 99\%
of the data.\\
\texttt{\textbf{Gaussian}}: We also estimate a multivariate Gaussian
distribution for the bounding box parameters centre position, square
root area, and log aspect ratio. After calculating mean and covariance
on the training set we sample proposals from this distribution.\\
\texttt{\textbf{SlidingWindow}} places windows on a regular grid as
common for sliding window object detectors. The contingent of the
requested number of proposals is distributed across different windows
sizes. For each window size, we place the windows uniformly across
the image. The procedure is inspired by the implementation of \texttt{Bing}~\cite{Cheng2014Cvpr}.\texttt{ }
 \\
\texttt{\textbf{Superpixels}}: As discussed in the next sections,
low-level superpixels have an important influence on the behaviour
of the detection proposal methods. As five of the compared methods
build on \cite{Felzenszwalb2004IJCV}, we use it as a baseline: each
low-level segment is used as a detection proposal. We should expect
this method to have low recall for objects (\S \ref{sec:Proposal-recall}),
but high repeatability (\S \ref{sec:Repeatibility}).

\section{\label{sec:Repeatibility}Proposal repeatability}

\begin{wrapfigure}{r}{0.5\columnwidth}%
\vspace{-6em}
\hspace*{\fill}\includegraphics[width=0.15\textwidth]{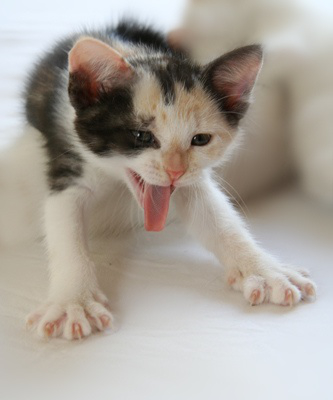}\hspace*{\fill}\includegraphics[width=0.15\textwidth]{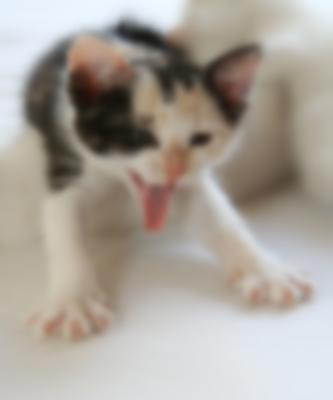}\hspace*{\fill}\includegraphics[width=0.15\textwidth]{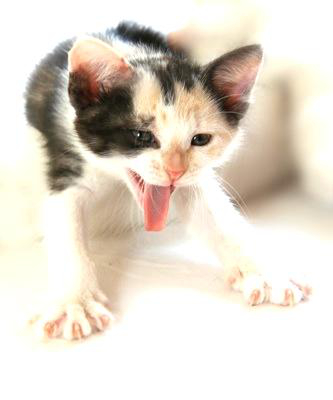}\hspace*{\fill}

\begin{centering}
\hspace*{\fill}\includegraphics[width=0.15\textwidth]{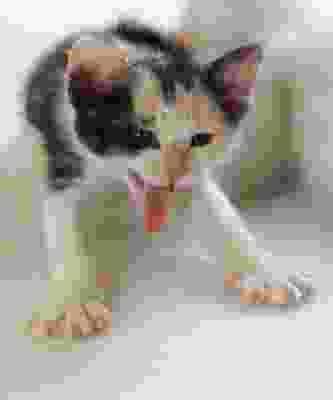}\hspace*{\fill}\includegraphics[width=0.15\textwidth]{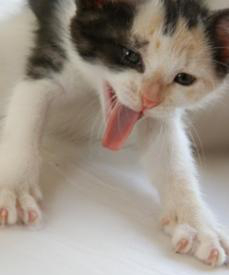}\hspace*{\fill}\includegraphics[width=0.15\textwidth]{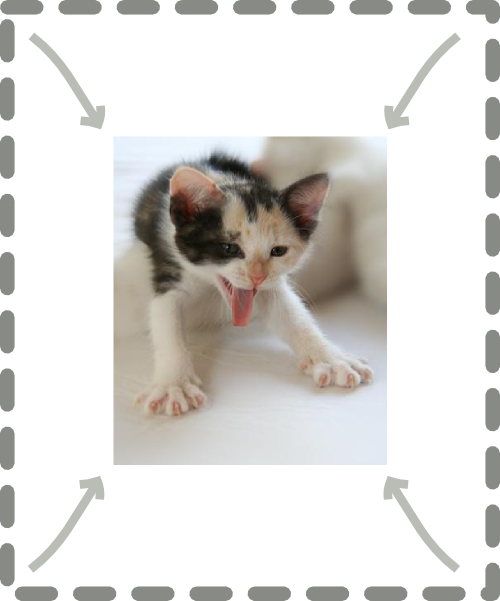}\hspace*{\fill}
\par\end{centering}

\vspace{-0.5em}

\protect\caption{\label{fig:perturbations-example}Example of the image perturbations
considered. Top to bottom, left to right: original, then blur, illumination,
JPEG artefact, rotation, and scale perturbations.}
\vspace{-1em}

\end{wrapfigure}%
Before looking into how well the different object proposals overlap
with ground truth annotations of objects, we want to answer a more
basic question. Training a detector on detection proposals rather
than on all sliding windows modifies the distribution of negative
windows that the classifier is trained on. If the proposal method
does not consistently propose windows on similar image content without
objects or with partial objects, the classifier cannot produce useful
scores on negative windows on the test set. We call the property of
proposals being placed on similar image content the repeatability
of a proposal method. Intuitively the detection proposals should be
repeatable on slightly different images with the same image content.
To evaluate repeatability we project proposals from one image into
another slightly modified image. Pascal does not contain suitable
images. An alternative is the dataset of \cite{Mikolajczyk2005Ijcv},
but it only consists of 54 images and only few of the images contain
objects. Instead, we opt to apply synthetic transformations to Pascal
images, so we know how to project proposals from one image into another.

\subsection{Evaluation protocol}

Our evaluation protocol is inspired by \cite{Mikolajczyk2005Ijcv},
which evaluates interest point repeatability. For each image in the
Pascal VOC 2007 test set, we generate several perturbed versions.
We consider changes in scale, blur, small rotation, illumination,
and JPEG compression (see figure~\ref{fig:perturbations-example}).
The details of the transformations and additional examples of the
perturbed images are provided in the supplementary material.

For each pair of reference and perturbed images we compute detection
proposals with a given algorithm (requesting $1\,000$ windows per
image). The proposals are projected back from the perturbed into the
reference image and then matched to the proposals in the reference
image. All proposals whose centre lies outside the image after projection
are removed before matching (which can only happen for rotation).
Matching is done greedily according to the intersection over union
(IoU) criterion. Given the matching, we plot the recall for every
IoU threshold and define the area under this ``recall versus IoU
threshold'' curve to be the repeatability. Methods that propose windows
at similar locations at high IoU are more repeatable, since the area
under the curve is larger.

One issue regarding such proposal matching is that large windows are
more likely to match than smaller ones. This effect is important to
consider since different methods have quite different distributions
of window areas (see figure~\ref{fig:distribution-of-windows-sizes}).
To reduce the impact of this effect, we bin the original image windows
by area (into 10 groups), and evaluate the area under the recall versus
IoU curve per size group. The plots in figure \ref{fig:repeatability}
show the unweighted average across size bins.

Figure~\ref{fig:distribution-of-windows-sizes} shows that the different
proposal methods exhibit a variety of different distributions of proposal
window sizes and in general different from the ground truth annotation
distribution. This confirms the need for normalisation based on the
window size. Figure~\ref{fig:fluctuation-per-size-group} shows the
recall versus intersection over union (IoU) curve for a blur perturbation.
This example shows the effect, that large rectangles have higher repeatability,
which is why we use the average across size bins to analyse repeatability
in the following.

We omit the slowest two methods, \texttt{CPMC} and \texttt{Endres2010},
from this experiment because it involves running the candidate detector
over the Pascal test set \textasciitilde{}50 times (once for every
perturbation).

\subsection{Repeatability results}

\begin{figure}[t]
\begin{centering}
\vspace{-1em}
\hspace*{\fill}\subfloat[\label{fig:distribution-of-windows-sizes}Histogram of proposal window
sizes on Pascal VOC 2007 test.]{\begin{centering}
\includegraphics[width=0.29\textwidth]{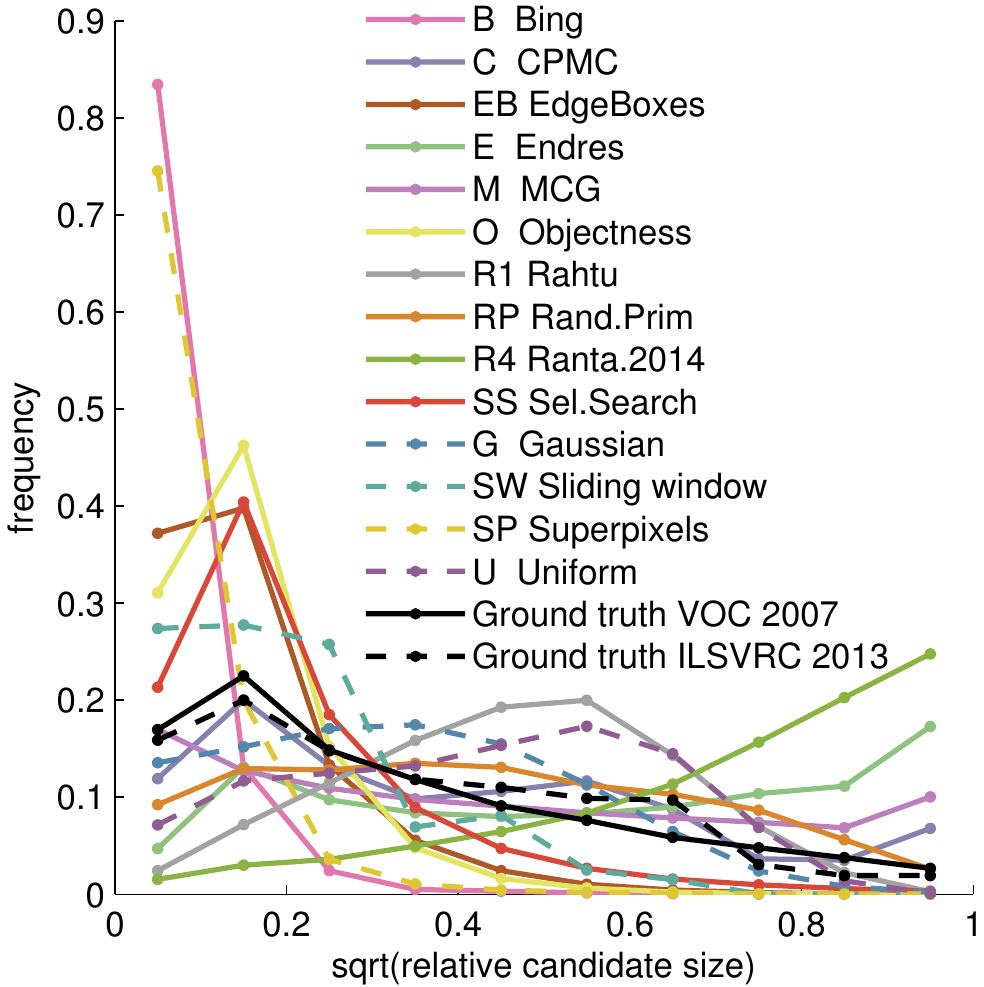}
\par\end{centering}

}\hspace*{\fill}\subfloat[\label{fig:fluctuation-per-size-group}Example of recall for different
candidate window sizes. ]{\begin{centering}
\includegraphics[width=0.3\textwidth]{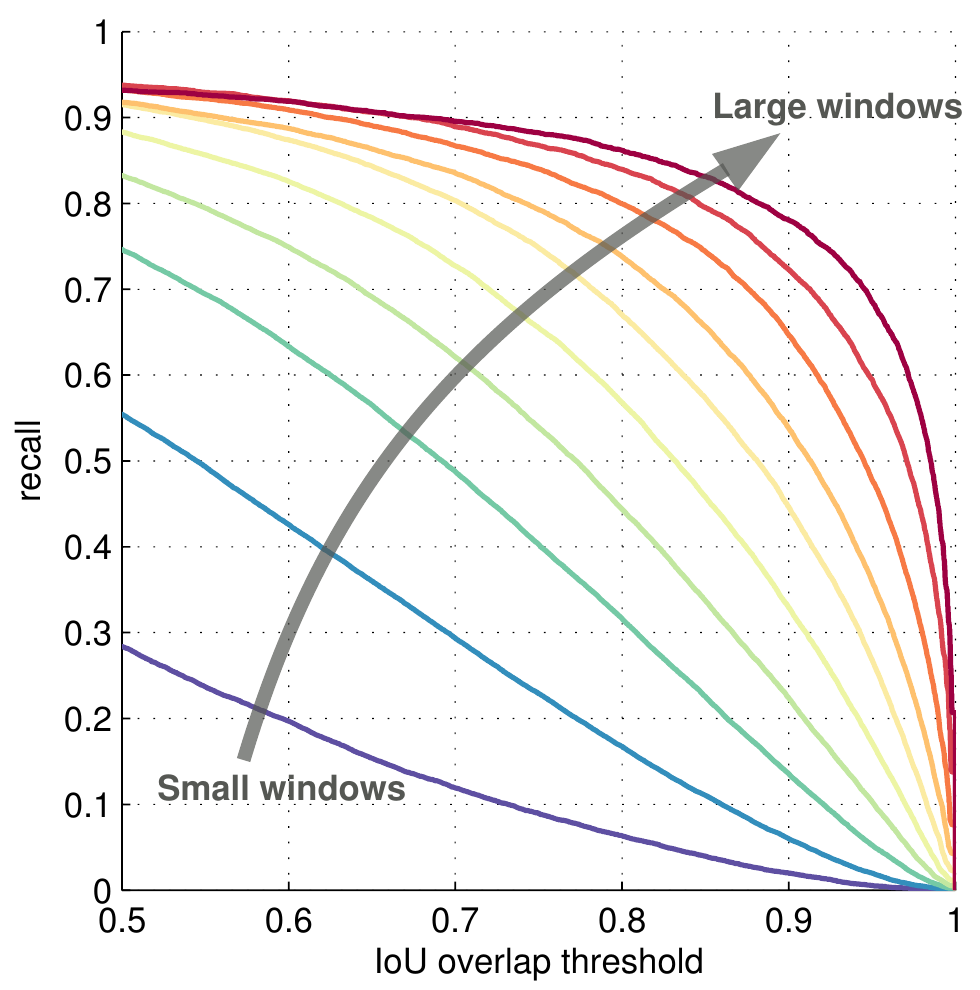}
\par\end{centering}

}\hspace*{\fill}\subfloat[\label{fig:repeatiblity-scale-change}Scale change.]{\begin{centering}
\includegraphics[width=0.3\textwidth]{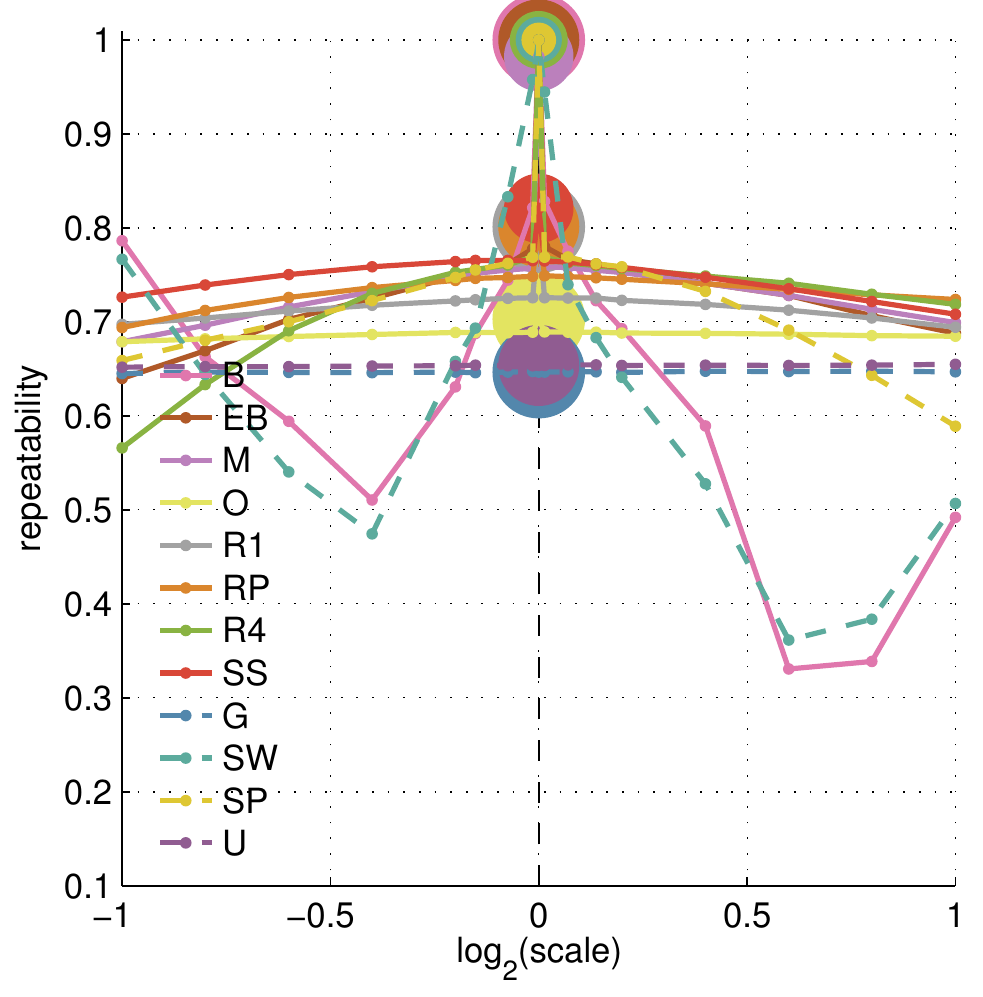}
\par\end{centering}

}\hspace*{\fill}\vspace{-1.25em}

\par\end{centering}

\begin{centering}
\hspace*{\fill}\subfloat[\label{fig:repeatiblity-jpeg-change}JPEG artefacts.]{\begin{centering}
\includegraphics[width=0.3\textwidth]{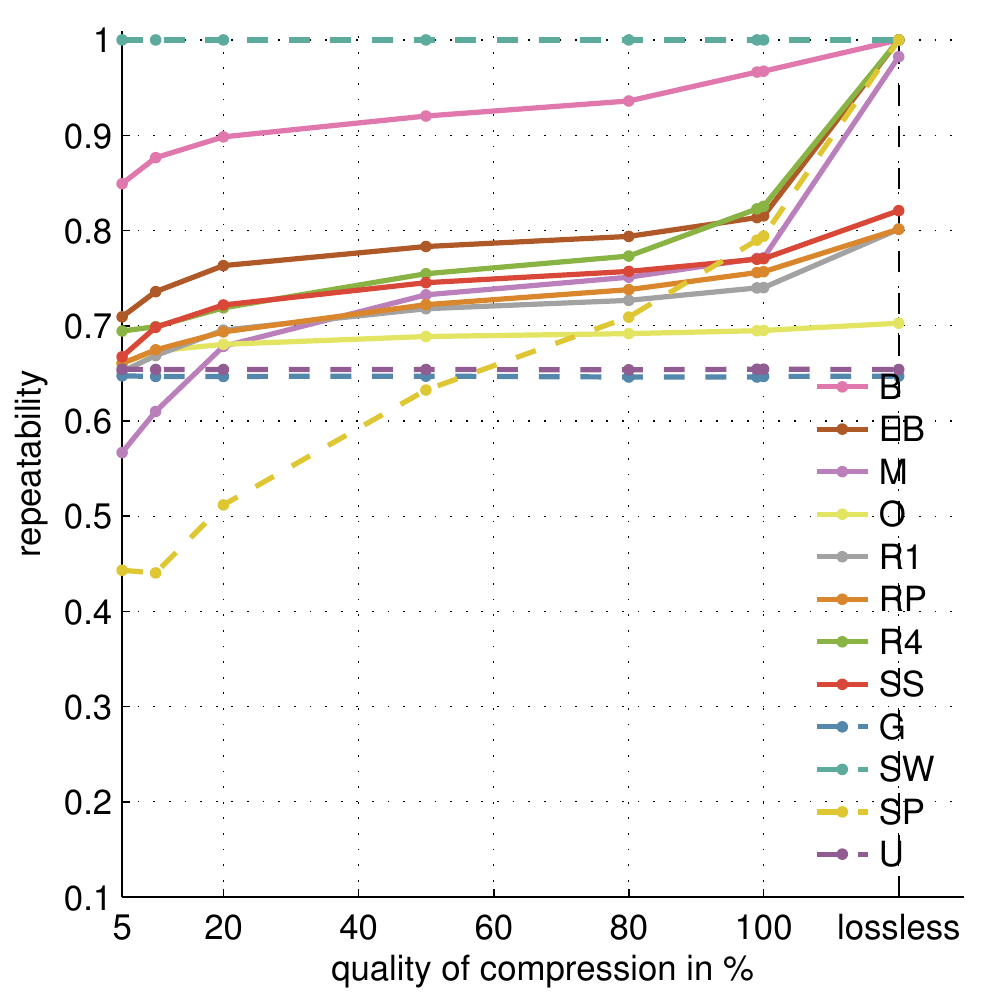}
\par\end{centering}

}\hspace*{\fill}\subfloat[\label{fig:repeatiblity-rotation-change}Rotation.]{\begin{centering}
\includegraphics[width=0.3\textwidth]{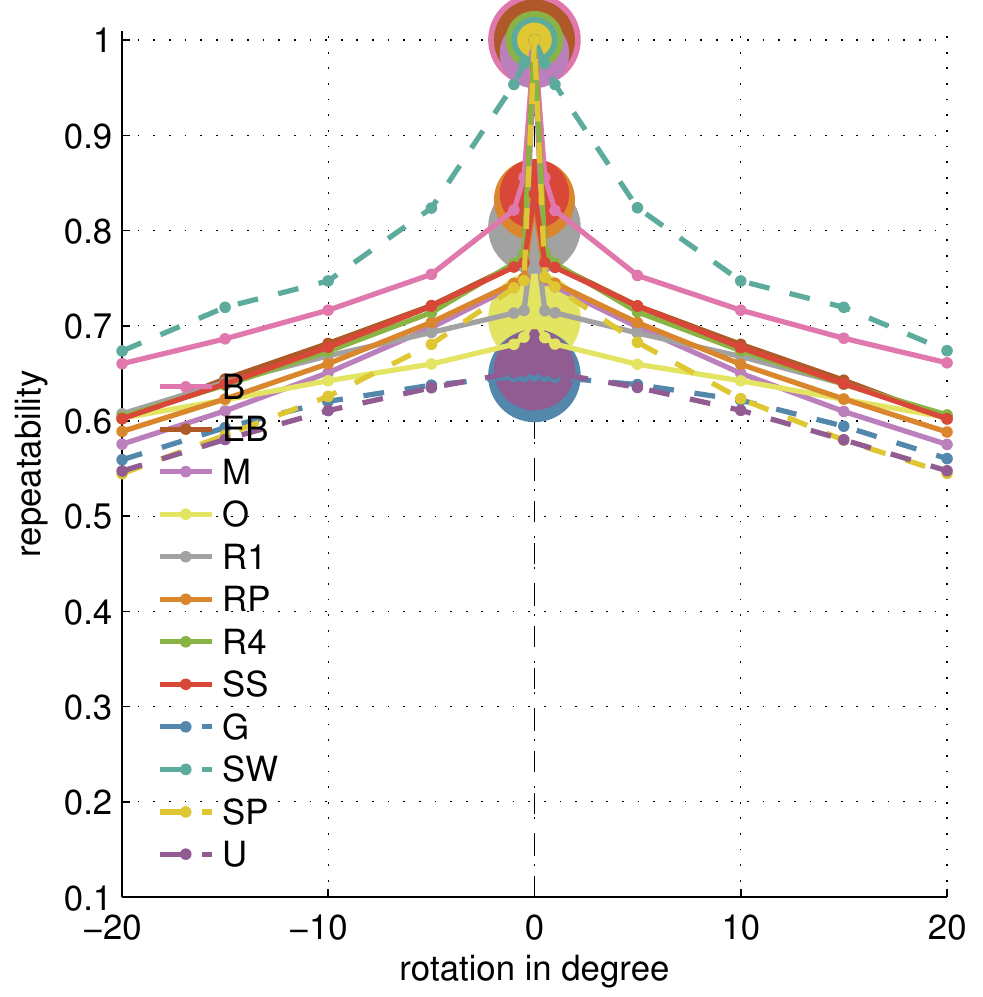}
\par\end{centering}

}\hspace*{\fill}\subfloat[\label{fig:repeatiblity-illumination-change}Illumination.]{\begin{centering}
\includegraphics[width=0.3\textwidth]{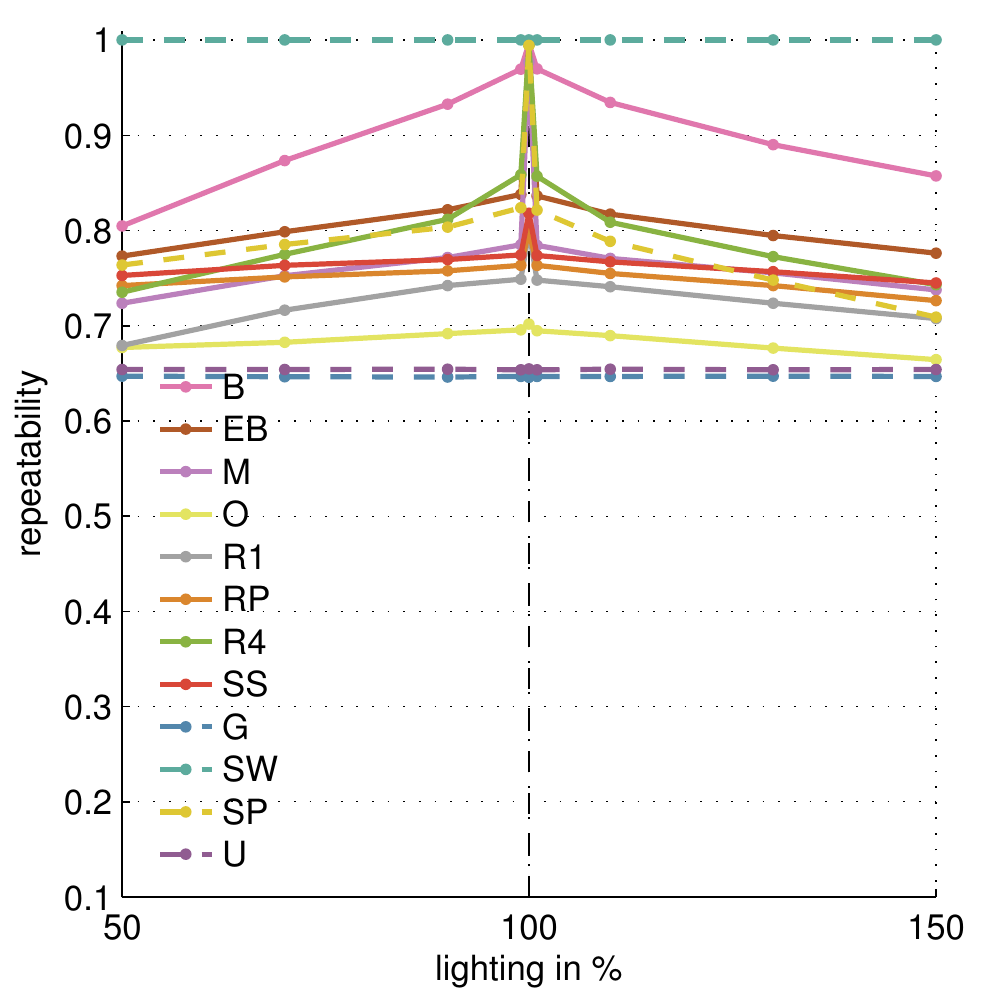}
\par\end{centering}

}\hspace*{\fill}
\par\end{centering}

\vspace{0.5em}

\protect\caption{\label{fig:repeatability}Repeatability results under perturbations.
See also supplementary material.}
\vspace{-1.5em}
\end{figure}
There are some salient aspects of the result curves in figure \ref{fig:repeatability}
that need some explanation. First, not all methods have $100\%$ repeatability
when there is no perturbation. This is due to random components in
the selection of proposals for several methods (\texttt{Bing} being
the most notable exception). It is possible to control the random
aspect by fixing the random seed, but this would not be effective
when perturbations are present and the underlying segmentation changes:
a drop in performance could be due to pseudo-random samples or to
an actual change in behaviour. Another approach would be to remove
the random aspect from all methods, but this would denature the method
being evaluated and we would actually evaluate a different algorithm.
By keeping the random aspect when no transformation is applied to
the image, we can compare values along the change in perturbation.

A second important aspect is the large drop of repeatability for
most methods, even for very subtle image changes. We have created
the \texttt{Superpixels} baseline as a means to validate our hypothesis
regarding this effect. Most methods use low-level superpixels as a
building block for the proposal windows. We suspect that the superpixels
themselves are unstable under small perturbations. This is supported
by the fact that the \texttt{Superpixels} baseline also shows a strong
drop (since it is a direct reflection of the superpixels themselves).
Since proposals on larger perturbations are all matched against the
same proposals in the reference (unperturbed) image, after the very
first small perturbation most windows are ``lost'', the remaining
windows that are correctly matched are progressively lost as the perturbation
level increases. Inversely we notice that methods that are not based
on superpixels are most robust to small image changes (\texttt{Bing}
and the baselines that ignore image content being the most noticeable
ones).

We now briefly discuss the effects under each perturbation, shown
in figure \ref{fig:repeatability}:\\
\textbf{Scale~change} \ref{fig:repeatiblity-scale-change}: All methods
except \texttt{Bing} show a drastic drop with small scale changes,
and minor relative degradation for larger changes. \texttt{Bing}'s
drop is less extreme for very small scale changes. For larger changes,
however, it presents a non monotonous behaviour because it uses a
coarse set of box sizes while searching for detection candidates.
Such coarse set impedes \texttt{Bing} to detect the same object at
two slightly different scales. Logically, our \texttt{SlidingWindow}
baseline suffers from the same effect.\\
\textbf{JPEG~artefacts} \ref{fig:repeatiblity-jpeg-change}: Similar
to scale change, even small JPEG artefacts have a large effect and
more aggressive JPEG compression factors show monotonic degradation.
Despite using gradient information \texttt{Bing} is the method most
robust to these kind of changes. It should be noted that JPEG images
with $100\%$ quality are still lossy. Only when requesting lossless
JPEG the superpixels are preserved because the image content does
not change.\\
\textbf{Rotation} \ref{fig:repeatiblity-rotation-change}: Here all
methods are equally affected by the perturbation. The drop of the
\texttt{Uniform} and \texttt{Gaussian} baselines indicate the repeatability
loss due to the fact that we are matching rotated bounding boxes.\\
\textbf{Illumination~change} \ref{fig:repeatiblity-illumination-change}
shows a similar trend to JPEG artefacts. We notice here that the gradient
information used by \texttt{Bing} is more robust to small changes
than the superpixels that are used by other methods. Changes in illumination
reduce the difference between neighbour pixels, and in turn, affect
how they are merged into larger superpixels.\\
\textbf{Blurring} exhibits a similar trend to JPEG artefacts, although
the drop is stronger for a small $\sigma$. The corresponding plot
can be found in the supplementary material.

Overall it seems that \texttt{Bing} and \texttt{EdgeBoxes} are more
repeatable than other methods, possibly because both use machine components
(SVM classifier for scoring and decision forests for features computation,
respectively). We also conclude that the sensitivity of superpixels
to image perturbations is a major cause for degradation in repeatability
of several detection proposal methods.

To better understand how much repeatability influences detector performance,
one should train detectors on proposals and compare the detection
performance of the detector using different proposal methods. This
is, however, a more complicated procedure that involves many new parameters
(see e.g. \cite{Wang2013Iccv,Girshick2014Cvpr}) which is why we leave
this for future work.

\section{\label{sec:Proposal-recall}Proposal recall}

When using detection proposals it is important to have a good coverage
of the true objects in the test image, since missed objects will never
be recovered. Thus it is common practice to evaluate the quality of
proposals based on ground truth annotations.

\subsection{Evaluation protocol}

The protocol introduced in \cite{Alexe2010Cvpr} has served as a
guideline for most other evaluations in the literature. Typically
the detection proposal method is trained on a subset of the Pascal
VOC categories, and tested on the test set, including unseen categories.
The metric of interest we use is the fraction of ground truth annotation
covered above a certain intersection over union (IoU) threshold. Note
that evaluating (class agnostic) detection proposals is quite different
from traditional class-specific detection \cite{Hoiem2012Eccv}, since
most of the metrics (class confusion, background confusion, precision,
etc.) do not apply.

\begin{figure}[t]
\begin{centering}
\vspace{-1em}
\hspace*{\fill}\subfloat[\label{fig:recall-vs-iuo-at-100-windows}$100$ proposals per image.]{\begin{centering}
\includegraphics[width=0.32\textwidth]{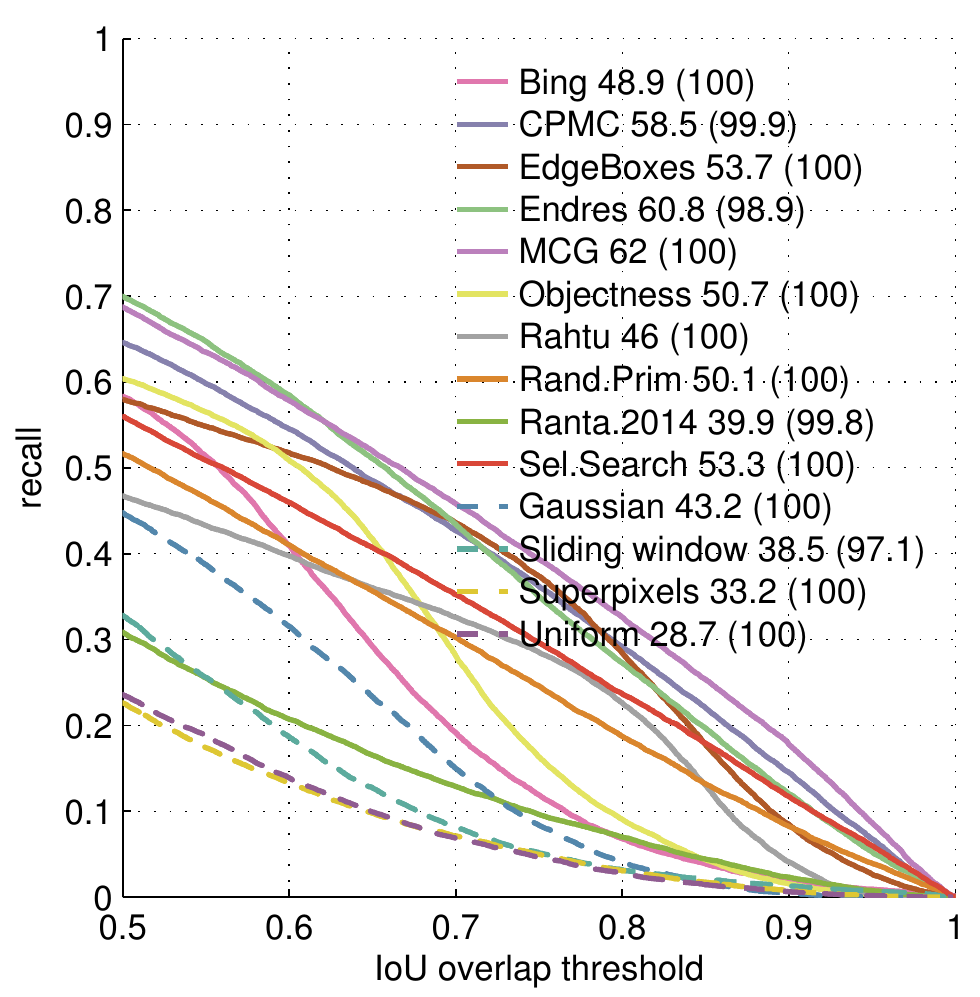}
\par\end{centering}

}\hspace*{\fill}\subfloat[\label{fig:recall-vs-iuo-at-1000-windows}$1\,000$ proposals per
image. ]{\begin{centering}
\includegraphics[width=0.32\textwidth]{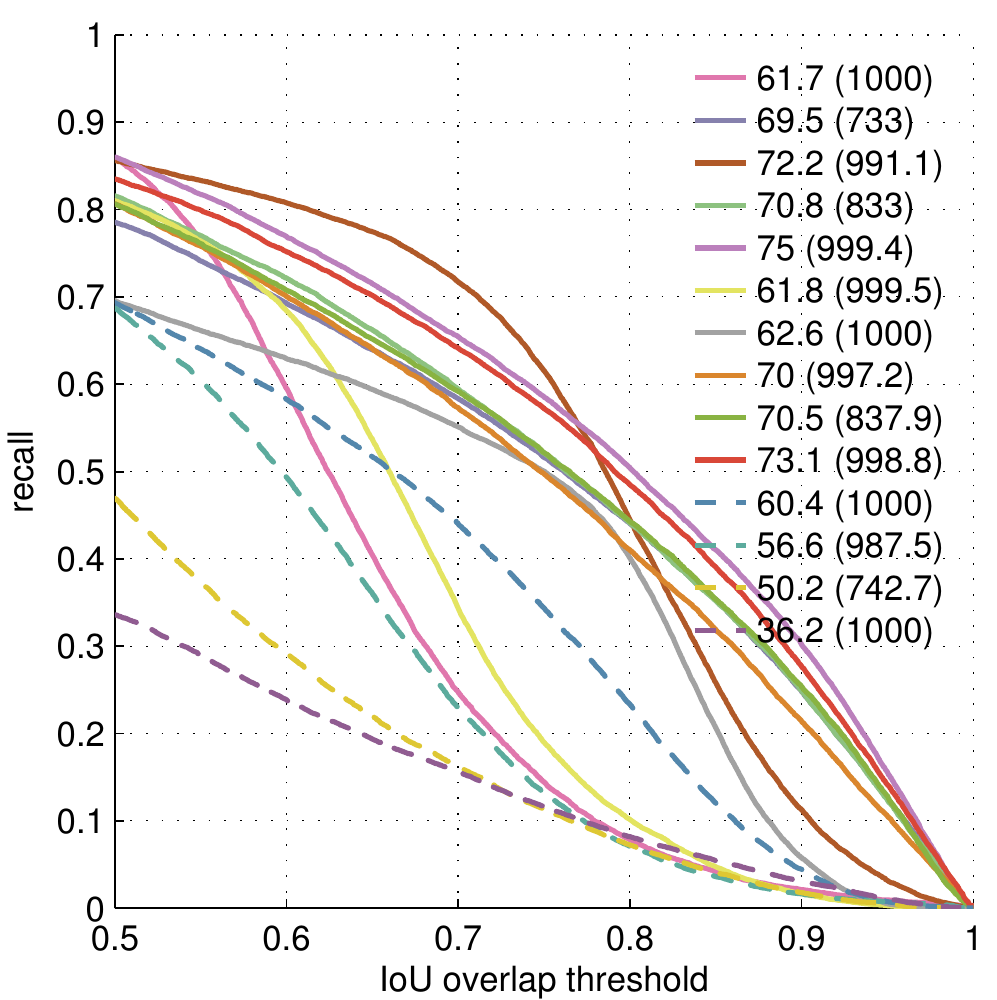}
\par\end{centering}

}\hspace*{\fill}\subfloat[\label{fig:recall-vs-iuo-at-10000-windows}$10\,000$ proposals per
image. ]{\begin{centering}
\includegraphics[width=0.32\textwidth]{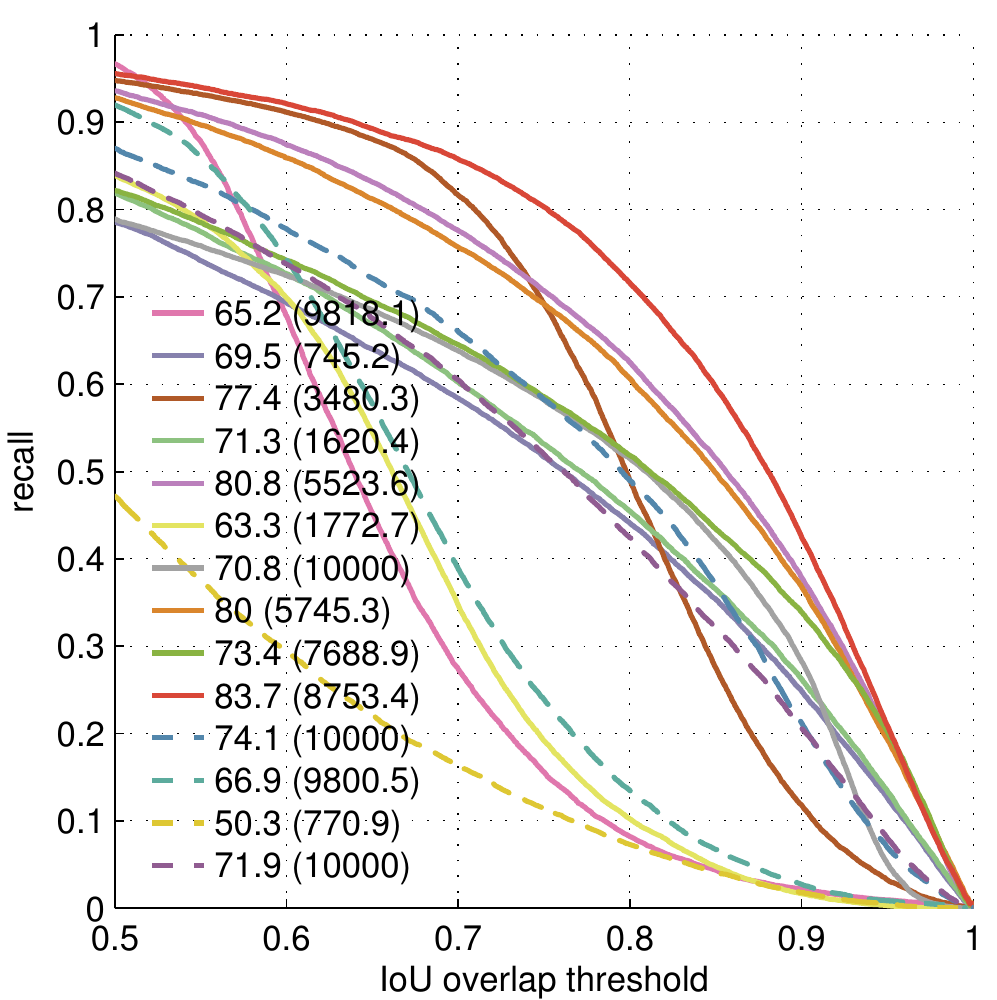}
\par\end{centering}

}\hspace*{\fill}
\par\end{centering}

\vspace{0.5em}

\protect\caption{\label{fig:recall-versus-iou-threshold-pascal}Recall versus IoU threshold
on the Pascal VOC 2007 test set. Numbers next to label indicate area
under the curve and average number of windows per image, respectively.}
\vspace{-1em}
\end{figure}

\begin{figure}[t]
\vspace{-1em}
\hspace*{\fill}\subfloat[\label{fig:recall-vs-iuo-area-vs-number-of-proposals-pascal}Area
under ``recall versus IoU threshold'' curves.]{\begin{centering}
\includegraphics[width=0.32\textwidth]{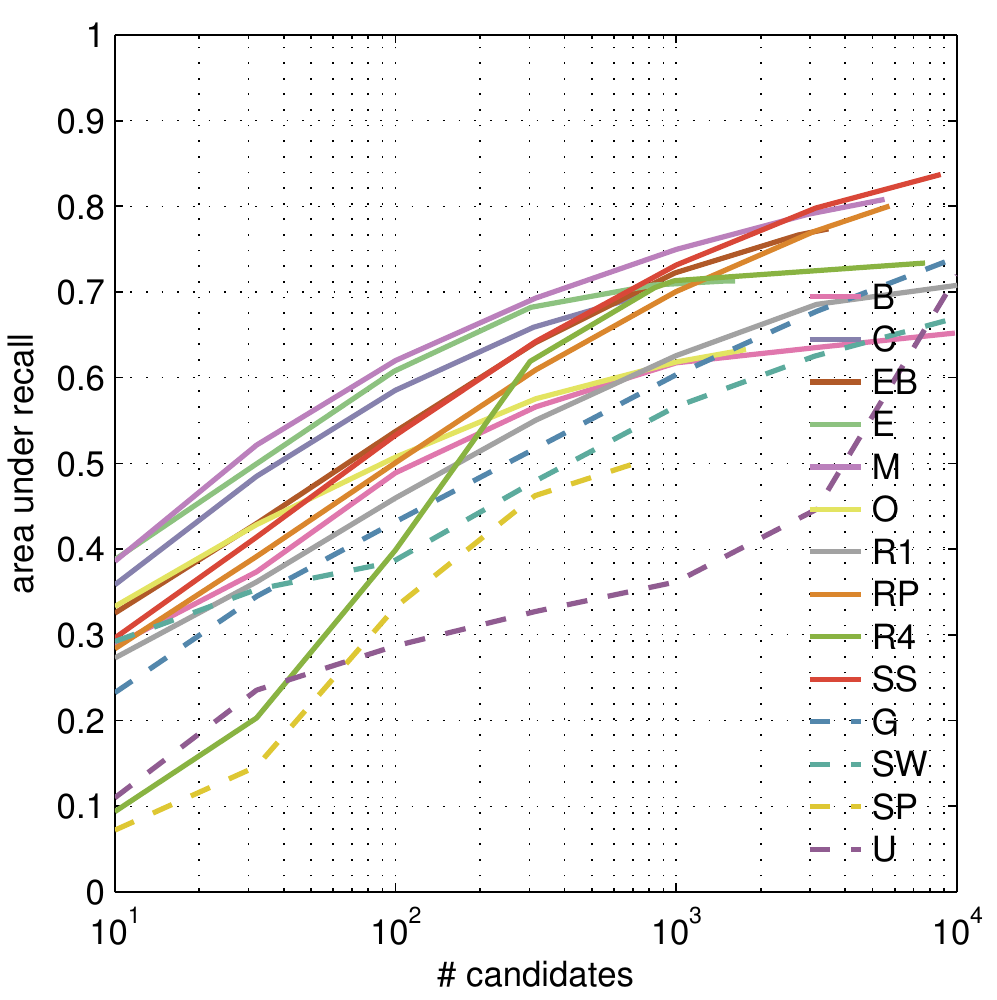}
\par\end{centering}

}\hspace*{\fill}\subfloat[\label{fig:recall-at-iou-0.5-vs-number-of-proposals-pascal}Recall
at IoU above 0.5.]{\begin{centering}
\includegraphics[width=0.32\textwidth]{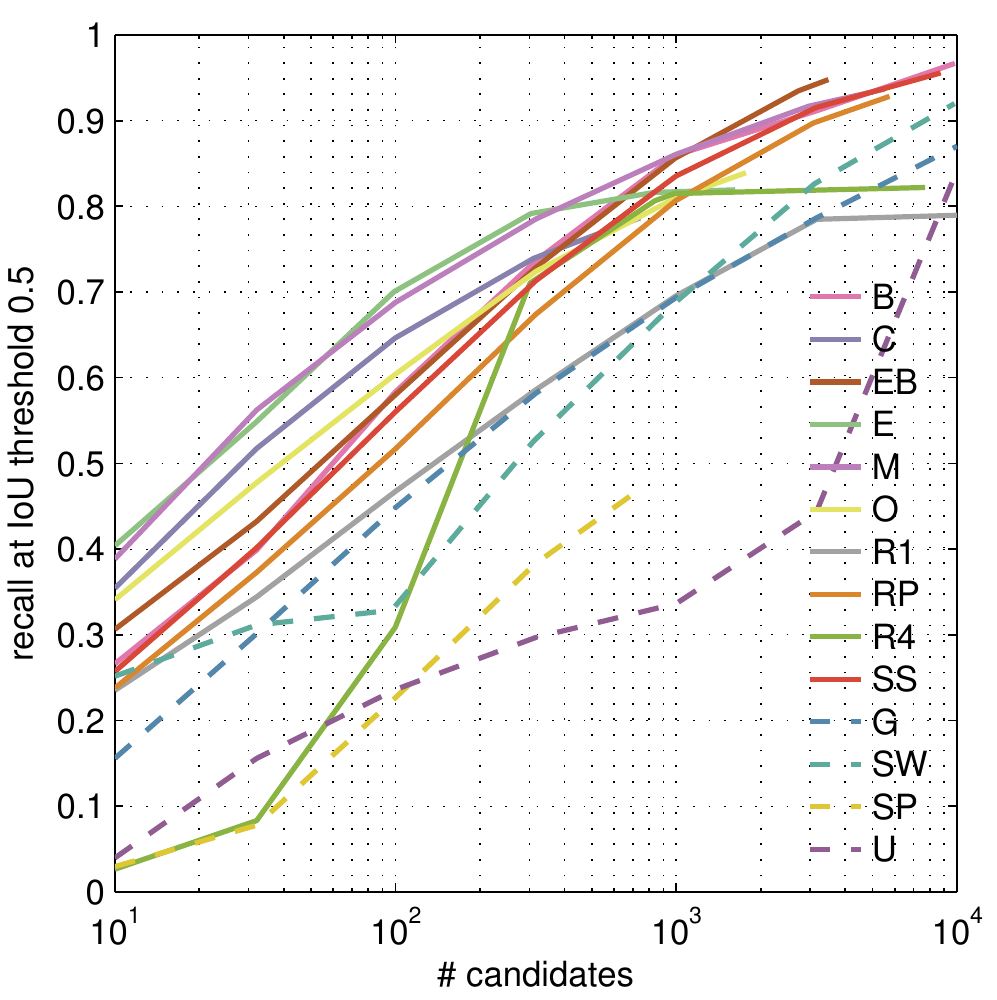}
\par\end{centering}

}\hspace*{\fill}\subfloat[\label{fig:recall-at-iou-0.8-vs-number-of-proposals-pascal}Recall
at IoU above 0.8.]{\begin{centering}
\includegraphics[width=0.32\textwidth]{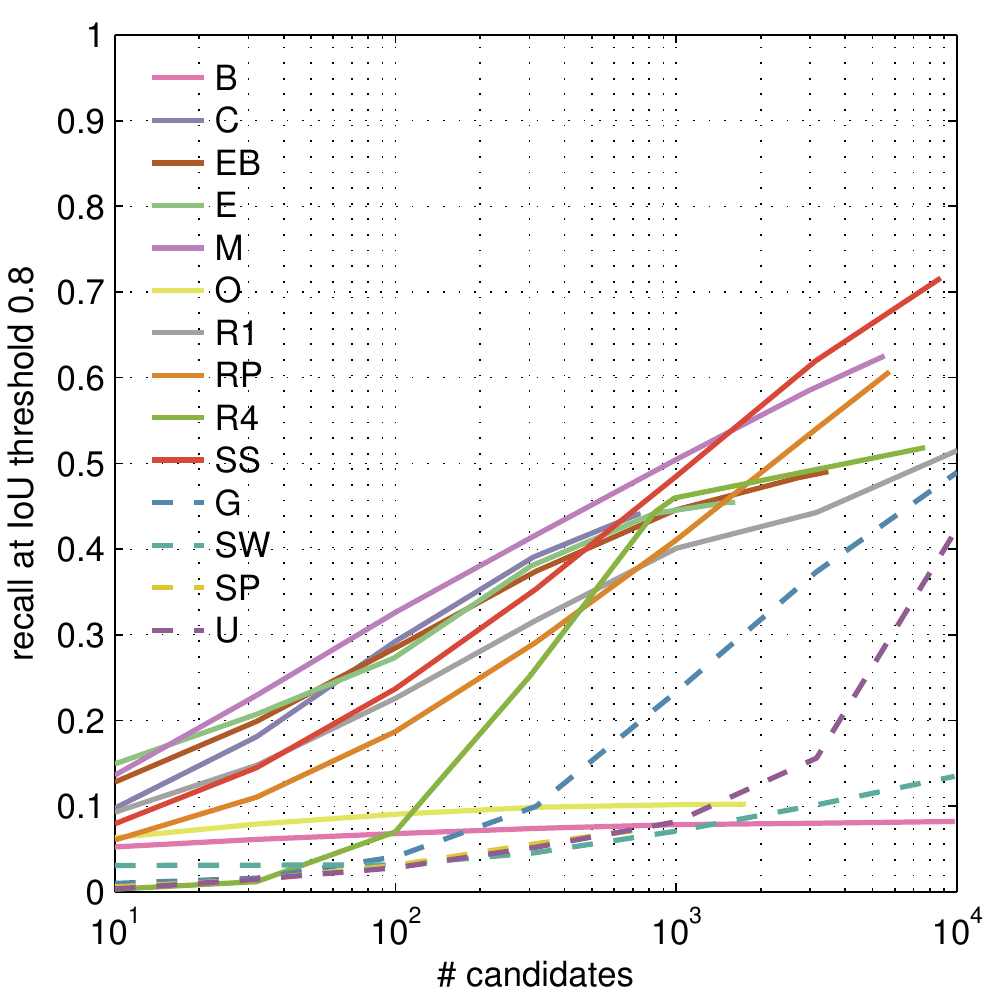}
\par\end{centering}

}\hspace*{\fill}\vspace{0.5em}
\protect\caption{\label{fig:recall-versus-num-windows-pascal}Recall versus number
of proposed windows on the Pascal VOC 2007 test set.}
\vspace{-1em}
\end{figure}

Previous work includes comparisons, however the train and test sets
vary amongst papers, and the metrics shown tend to favour different
methods. We provide an extensive unified evaluation, that shows how
a different perspective can offer different ranking of the methods
(see figure \ref{fig:recall-vs-iuo-at-1000-windows} versus \ref{fig:recall-at-iou-0.5-vs-number-of-proposals-pascal}).

We use the methods as provided by the authors. Different methods may
be trained on different sets, but we think this is still fair since
we care about absolute quality. Some methods do not have training
at all (\texttt{SelectiveSearch}), yet provide competitive results.
To mitigate this effect we evaluate on the full Pascal VOC 2007 test
set, including all categories.

The Pascal VOC 2007 test set has only $20$ categories, present in
a diverse set of $\sim5\,000$ unconstrained images, yet detection
proposal methods attempt to predict ``any'' object. On principle,
methods which are particularly good at detecting non-Pascal categories
will be penalised compared to the ones that are good at these $20$
categories but nothing else. To investigate this dataset bias, we
also evaluate methods on the larger ImageNet \cite{Deng2009Cvpr}
2013 validation set, which contains annotations for $200$ categories
over $\sim20\,000$ images. It should be noted that these $200$ categories
are \emph{not} fine grained versions of the Pascal ones. It includes
additional types of animals (e.g. crustaceans), food items (e.g. hot-dogs),
house hold items (e.g. diapers), and other diverse object categories.

\subsection{Recall results}

\begin{wrapfigure}{r}{0.35\columnwidth}%
\begin{centering}
\vspace{-6.5em}
\includegraphics[width=0.35\textwidth]{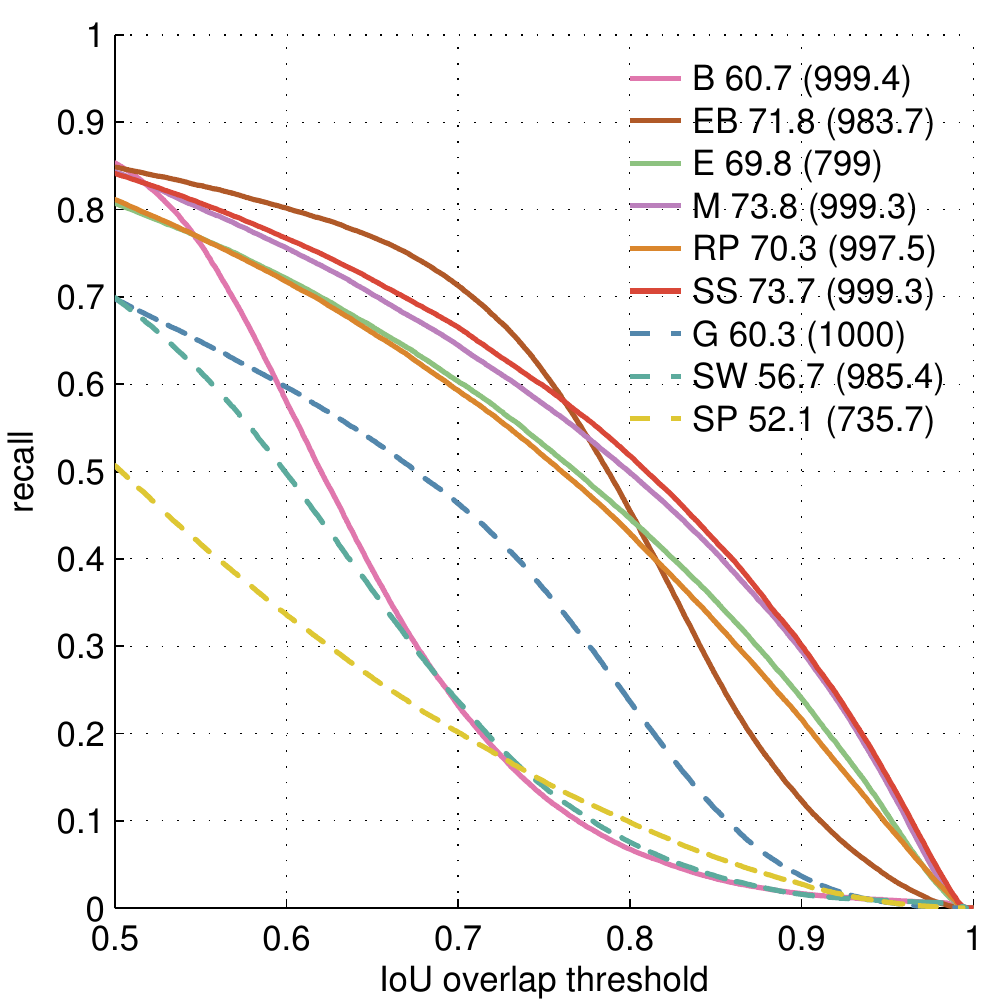}\vspace{-0.5em}

\par\end{centering}

\protect\caption{\label{fig:imagenet-quality-results}\label{fig:recall-vs-iuo-at-1000-windows-imagenet}Recall
versus IoU threshold, for $1\,000$ proposals per image on the ImageNet
2013 validation set. See figure \ref{fig:recall-versus-iou-threshold-pascal}
and supplementary material.}
\vspace{-1em}
\end{wrapfigure}%
Results in figure \ref{fig:recall-versus-iou-threshold-pascal} present
a consistent trend across the different metrics. \texttt{MCG},\texttt{
Edge\-Boxes}, and \texttt{Se\-lec\-tive\-Search} seem to be the
best methods across different number of proposals. 

We show \texttt{EdgeBoxes} when tuned for IoU at $0.7$, which results
in a clear bump in recall. When tuned for $\mbox{IoU}=0.9$, for $10^{3}$
proposals, \texttt{Edge\-Boxes }is below \texttt{MCG}, and when tuned
for $\mbox{IoU}=0.5$ it reaches about $93\%$ recall. \texttt{MCG}
shows the most consistent results when varying the number of proposals.
\texttt{Se\-lective\-Search} is surprisingly effective despite being
a fully hand-crafted method (no machine learning involved). When considering
less than $10^{3}$ proposals, \texttt{MCG}, \texttt{Endres\-2010},
and \texttt{CPMC} provide strong results.

Overall, we can group the methods into two groups: well localized
methods that gradually lose recall as the IoU threshold increases
and methods that only provide approximate bounding box locations,
so their recall drops dramatically. All baseline methods, as well
as \texttt{Bing}, \texttt{Rahtu2011}, and \texttt{Edge\-Boxes} fall
into the latter category. \texttt{Bing} in particular, while providing
high repeatability, only provides high recall at $\mbox{IoU}=0.5$
and drops dramatically when requiring higher overlap.

When inspecting figure \ref{fig:recall-versus-iou-threshold-pascal}
from left to right, one notices that with only few proposal windows
our baselines provide low recall compared to non-baseline methods
(figure \ref{fig:recall-vs-iuo-at-100-windows}). However as the number
of proposals increases baselines such as \texttt{Gaussian} or \texttt{Sli\-ding\-Win\-dow}
become competitive (figure \ref{fig:recall-vs-iuo-at-1000-windows},
$\mbox{IoU}>0.7$). Eventually for enough windows even simple random
baselines such as \texttt{Uniform} become competitive (figure \ref{fig:recall-vs-iuo-at-10000-windows}).
In relative gain, detection proposal methods have most to offer for
low numbers of windows.

Figure \ref{fig:imagenet-quality-results} presents the results over
the ImageNet validation dataset. Compared to Pascal VOC 2007 test
set, it includes $10\times$ ground truth classes and $4\times$ images.
Somewhat surprisingly the results are almost identical to the ones
in \ref{fig:recall-vs-iuo-at-1000-windows}. A more detailed inspection
reveals that, by design, the statistics of ImageNet match the ones
of Pascal VOC. In particular the typical image size, and the mean
number of object annotation per image (three) is identical in both
datasets. This explains why the recall behaviour is similar, and why
\texttt{Se\-lective\-Search},\texttt{ Edge\-Boxes}, and \texttt{MCG}
still perform well despite being crafted for Pascal VOC.

\begin{wraptable}{r}{0.24\columnwidth}%
\begin{centering}
\vspace{-0.5em}
{\scriptsize{}}%
\begin{tabular}{c|c}
Method & mAP\tabularnewline
\hline 
\hline 
{\scriptsize{}LM-LLDA} & \textbf{\scriptsize{}33.5/34.4}\tabularnewline
\hline 
{\scriptsize{}Objectness} & {\scriptsize{}25.0/25.4}\tabularnewline
{\scriptsize{}CPMC} & {\scriptsize{}29.9/30.7}\tabularnewline
{\scriptsize{}Endres2010} & {\scriptsize{}31.2/31.6}\tabularnewline
{\scriptsize{}Sel.Search} & {\scriptsize{}31.7/32.3}\tabularnewline
{\scriptsize{}Rahtu2011} & {\scriptsize{}29.6/30.4}\tabularnewline
{\scriptsize{}Rand.Prim} & {\scriptsize{}30.5/30.9}\tabularnewline
{\scriptsize{}Bing} & {\scriptsize{}21.8/22.4}\tabularnewline
{\scriptsize{}MCG} & \textbf{\scriptsize{}32.4/32.7}\tabularnewline
{\scriptsize{}Ranta.2014} & {\scriptsize{}30.7/31.3}\tabularnewline
{\scriptsize{}EdgeBoxes} & {\scriptsize{}31.8/32.2}\tabularnewline
\hline 
{\scriptsize{}Uniform} & {\scriptsize{}16.6/16.9}\tabularnewline
{\scriptsize{}Gaussian} & \textbf{\scriptsize{}27.3/28.0}\tabularnewline
{\scriptsize{}Slid.Window} & {\scriptsize{}20.7/21.5}\tabularnewline
{\scriptsize{}Superpixels} & {\scriptsize{}11.2/11.3}\tabularnewline
\end{tabular}\vspace{1em}

\par\end{centering}

\protect\caption{\label{tab:detection-results}Detection results on Pascal 2007 (left/right
mAP is before/after bounding box regression).}
\vspace{-4.5em}
\end{wraptable}%
Although the curves look very similar, ImageNet covers $200$ object
categories, many of them unrelated to Pascal VOC. The good news is
that having almost identical results in both datasets confirms that
these methods do transfer adequately amongst object classes, and thus
can be considered as ``true objectness'' measures. In other words,
the experiments of figure \ref{fig:imagenet-quality-results} indicate
that there is no visible over-fitting of the candidate detection methods
towards the $20$ Pascal VOC categories.

For researchers looking to benchmark their detection proposal method,
figure \ref{fig:recall-vs-iuo-area-vs-number-of-proposals-pascal}
serves as good overall summary. It puts aside the IoU threshold, because
detection proposal users can typically estimate their computation
constrains (maximum number of proposals) better than the needed detection
proposals localization (IoU threshold). If a specific IoU threshold
is desired, then summary figure such as  \ref{fig:recall-at-iou-0.5-vs-number-of-proposals-pascal}
or \ref{fig:recall-at-iou-0.8-vs-number-of-proposals-pascal} are
suitable.

\section{\label{sec:Using-detection-proposals}Using the detection proposals}

This section analyses the detection proposals in the context of a
DPM object detector, namely the LM-LLDA variant \cite{Girshick2013ICCV}.
We use a pre-trained model and different proposals to filter its detections
at test time. This experiment does not speed-up the detection, but
enable evaluating the effect of proposals on the detection quality.
A priori detections might get worse (by losing recall), but can also
improve if the detection proposal method discards windows that would
otherwise be false positives.\\
\textbf{Implementation details}$\quad$We take the raw detections
of an LM-LLDA model before non-maximum suppression and filter them
with the detection proposals of each method. We keep all detections
that overlap more than 0.8 IoU with a candidate; the surviving detections
are then non-maxima suppressed. As final step we do bounding box regression,
as common for DPM models \cite{Felzenszwalb2010Pami}. Note that with
this procedure the detector is not only evaluated inside each proposal
window, but also around it.\\
\textbf{Results}$\quad$Table \ref{tab:detection-results} shows that
using $1\,000$ detection proposals decreases detection quality compared
to sliding window.   When we compare these results with figure~\ref{fig:recall-vs-iuo-at-1000-windows},
we see that methods with high area under the curve (above $69\%$),
also have high mAP. Although \texttt{Rahtu} and the \texttt{Gaussian}
baseline have a similar area under the curve as \texttt{Objectness}
and \texttt{Bing}, the latter have are lower mAP than the former.
We attribute this to higher recall in the high precision area ($\mbox{IoU}\geq0.7$).
These results show that better localisation of proposals leads to
increased detection quality and support the analysis in \S\ref{sec:Proposal-recall}.

The per class recall versus precision plots (in supplementary material)
can be grouped into three cases: 1) the best proposal methods do not
gain or lose much; 2) proposals sometimes clearly hurt performance
(bicycle, bottle, car, chair, motorbike, person, potted plant); 3)
proposals improve performance (aeroplane, cat, dining table, dog,
sofa). In the case of (2) we see reduced recall but also reduced precision,
probably because bad localization decreases scores of strong detections.

\section{\label{sec:Conclusion}Conclusion}

\begin{wraptable}{r}{0.555\columnwidth}%
\begin{centering}
\vspace{-6em}
\begin{tabular}{lc||c|c|c|c}
\multirow{2}{*}{Method} & \multirow{2}{*}{} & {\scriptsize{}Ti-} & {\scriptsize{}Repea-} & {\scriptsize{}Re-} & {\scriptsize{}Detec-}\tabularnewline
 &  & {\scriptsize{}me} & {\scriptsize{}tability} & {\scriptsize{}call} & {\scriptsize{}tion}\tabularnewline
\hline 
\texttt{\scriptsize{}Objectness}{\scriptsize{}\cite{Alexe2010Cvpr}} & \texttt{\scriptsize{}O} & {\footnotesize{}3} & $\cdot$ & $\star$ & $\cdot$\tabularnewline
\texttt{\scriptsize{}CPMC}{\scriptsize{}\cite{Carreira2010Cvpr}} & \texttt{\scriptsize{}C} & {\footnotesize{}250} & - & $\star\star$ & $\star$\tabularnewline
\texttt{\scriptsize{}Endres2010}{\scriptsize{}\cite{Endres2010Eccv}} & \texttt{\scriptsize{}E} & {\footnotesize{}100} & - & $\star\star$ & $\star\star$\tabularnewline
\texttt{\scriptsize{}Sel.Search}{\scriptsize{}\cite{Sande2011Iccv}} & \texttt{\scriptsize{}SS} & {\footnotesize{}10} & $\star\star$ & $\star\star\star$ & $\star\star$\tabularnewline
\texttt{\scriptsize{}Rahtu2011}{\scriptsize{}\cite{Rahtu2011Iccv}} & \texttt{\scriptsize{}R1} & {\footnotesize{}3} & $\cdot$ & $\cdot$ & $\star$\tabularnewline
\texttt{\scriptsize{}Rand.Prim}{\scriptsize{}\cite{Manen2013Iccv}} & \texttt{\scriptsize{}RP} & {\footnotesize{}1} & $\star$ & $\star$ & $\star$\tabularnewline
\texttt{\scriptsize{}Bing}{\scriptsize{}\cite{Cheng2014Cvpr}} & \texttt{\scriptsize{}B} & {\footnotesize{}0.2} & $\star\star\star$ & $\star$ & $\cdot$\tabularnewline
\texttt{\scriptsize{}MCG}{\scriptsize{}\cite{Arbelaez2014Cvpr}} & \texttt{\scriptsize{}M} & {\footnotesize{}30} & $\star$ & $\star\star\star$ & $\star\star$\tabularnewline
\texttt{\scriptsize{}Ranta.2014}{\scriptsize{}\cite{Rantalankila2014Cvpr}} & \texttt{\scriptsize{}R4} & {\footnotesize{}10} & $\star\star$ & $\cdot$ & $\star$\tabularnewline
\texttt{\scriptsize{}EdgeBoxes}{\scriptsize{}\cite{Zitnick2014Eccv}} & \texttt{\scriptsize{}EB} & {\footnotesize{}0.3} & $\star\star$ & $\star\star\star$ & $\star\star$\tabularnewline
\hline 
\texttt{\scriptsize{}Uniform} & \texttt{\scriptsize{}U} & {\footnotesize{}0} & $\cdot$ & $\cdot$ & $\cdot$\tabularnewline
\texttt{\scriptsize{}Gaussian} & \texttt{\scriptsize{}G} & {\footnotesize{}0} & $\cdot$ & $\cdot$ & $\star$\tabularnewline
\texttt{\scriptsize{}SlidingWindow} & \texttt{\scriptsize{}SW} & {\footnotesize{}0} & $\star\star\star$ & $\cdot$ & $\cdot$\tabularnewline
\texttt{\scriptsize{}Superpixels} & \texttt{\scriptsize{}SP} & {\footnotesize{}1} & $\star$ & $\cdot$ & $\cdot$\tabularnewline
\end{tabular}\vspace{1em}

\par\end{centering}

\protect\caption{\label{tab:methods-comparison}Overview of detection proposal methods.\protect \\
Time is in seconds. Repeatability, quality, and detection rankings
are provided as rough qualitative overview; ``-'' indicates no data,
``$\cdot$'', ``$\star$'', ``${\star\star}$'', ``${\star\star\star}$''
indicate progressively better results. See paper's text for details
and quantitative evaluations.}
\vspace{-1em}
\end{wraptable}%
We evaluated ten detection proposal methods, among which \texttt{Se\-lec\-ti\-ve\-Search}
and \texttt{Edge\-Boxes} strike by their consistently good performance
both in ground truth recall, reasonable repeatability, and tolerable
evaluation speed (see table \ref{tab:methods-comparison}). If fast
proposals are required properly tuned \texttt{Edge\-Boxes} seem to
provide the best compromise in speed versus quality. When requiring
less than $10^{3}$ proposals, \texttt{MCG} is the method of choice
for high recall if speed is not a concern. \texttt{MCG} also obtains
top results regarding pure detection quality (table \ref{tab:detection-results}).

The extensive evaluation enables researchers to make more informed
decisions when considering to use detection proposals. We hope that
the provided code and data will encourage authors of detection proposal
methods to provide more complete comparisons to related work. Our
study also has revealed that most method suffer from low repeatability
due to unstable superpixels, even for the slightest of image perturbation.
We foresee room for improvement by using more robust superpixel (or
boundary estimation) methods. Finally, our ImageNet experiments have
validated that most methods do indeed generalise to different unseen
categories (beyond Pascal VOC), and as such can be considered true
``objectness'' methods.

In future work we plan to study the issue of missing recall in more
detail and do further experiments regarding the relation between detection
proposals and object detection.

\bibliographystyle{bmvc2k}
\bibliography{2014_bmvc_selective_search_references}
\clearpage{}

\appendix
\appendix

\part*{Supplementary Material}

\section{\label{sec:Overview}Method overview}

Table \ref{tab:methods-comparison-supp} presents an overview of all
eleven methods listed in \S2 of the paper. For three of the methods,
no code is available, so we evaluate the remaining eight. The table
also shows the four baseline methods.

Some of the methods have a parameter to ask for a certain number of
proposals and either return roughly that number of candidates (denoted
by ``R'') or return significantly less (denoted by ``LR''). Methods
without a parameter to directly ask for a number of proposals return
a number of proposals that depends on the image content and size.
These proposals are either sorted by likelihood of containing an object
(denoted by ``Fs'') or in an arbitrary order (denoted by ``F'').

\begin{table}[th]
\begin{centering}
\begin{tabular}{cc||c|c|c|c|c|c|c|c}
\multirow{2}{*}{} & \multirow{2}{*}{{\footnotesize{}Method}} & {\footnotesize{}Low } & {\footnotesize{}Output} & \multirow{2}{*}{{\footnotesize{}Score}} & {\footnotesize{}\#win-} & \multirow{2}{*}{{\footnotesize{}Time }} & {\footnotesize{}Repea-} & {\footnotesize{}Qua-} & {\footnotesize{}Detec-}\tabularnewline
 &  & {\footnotesize{}level} & {\footnotesize{}type} &  & {\footnotesize{}dows} &  & {\footnotesize{}tability} & {\footnotesize{}lity} & {\footnotesize{}tion}\tabularnewline
\hline 
\hline 
\texttt{Pb} & \texttt{gPbUCM} & {\footnotesize{}Own} & Sg & No & F & - & - & - & -\tabularnewline
\texttt{O} & \texttt{\scriptsize{}Objectness} & sp & Bx & Yes & R & 3 & $\cdot$ & $\star$ & $\cdot$\tabularnewline
\texttt{C} & \texttt{CPMC} & {\footnotesize{}Own} & Sg & Yes & Fs & 250 & - & $\star\star$ & $\star$\tabularnewline
\texttt{E} & \texttt{\scriptsize{}Endres2010} & sp & Sg & Yes & Fs & 100 & - & $\star\star$ & $\star\star$\tabularnewline
\texttt{SS} & \texttt{\scriptsize{}SelectiveSearch} & sp & Sg & No & F & 10 & $\star\star$ & $\star\star\star$ & $\star\star$\tabularnewline
\texttt{R1} & \texttt{Rahtu2011} & sp & Sg & Yes & R & 3 & $\cdot$ & $\cdot$ & $\star$\tabularnewline
\texttt{RP} & \texttt{\scriptsize{}RandomizedPrim} & sp & Sg & No & LR & 1 & $\star$ & $\star$ & $\star$\tabularnewline
\texttt{B} & \texttt{Bing} & {\footnotesize{}Own} & Bx & Yes & $\cdot$Fs & 0.2 & $\star\star\star$ & $\star$ & $\cdot$\tabularnewline
\texttt{M} & \texttt{MCG} & \texttt{C{*}} & Sg & Yes & R & 30 & $\star$ & $\star\star\star$ & $\star\star$\tabularnewline
\texttt{R4} & \texttt{\scriptsize{}Rantalankila2014}\texttt{} & {\footnotesize{}C+sp} & Sg & No & F & 10 & $\star\star$ & $\cdot$ & $\star$\tabularnewline
\texttt{R} & \texttt{Rigor} & \texttt{C{*}} & Sg & No & LR & - & - & - & -\tabularnewline
\texttt{EB} & \texttt{EdgeBoxes} & {\footnotesize{}Own} & Bx & Yes & Fs & 0.3 & $\star\star$ & $\star\star\star$ & $\star\star$\tabularnewline
\hline 
\texttt{U} & \texttt{Uniform} & $\emptyset$ & Bx & No & R & 0 & $\cdot$ & $\cdot$ & $\cdot$\tabularnewline
\texttt{G} & \texttt{Gaussian} & $\emptyset$ & Bx & No & R & 0 & $\cdot$ & $\cdot$ & $\star$\tabularnewline
\texttt{SW} & \texttt{\scriptsize{}SlidingWindow} & $\emptyset$ & Bx & No & R & 0 & $\star\star\star$ & $\cdot$ & $\cdot$\tabularnewline
\texttt{SP} & \texttt{Superpixels} & {\footnotesize{}Own} & Sg & No & F & 1 & $\star$ & $\cdot$ & $\cdot$\tabularnewline
\end{tabular}
\par\end{centering}

\begin{centering}
\vspace{1em}

\par\end{centering}

\protect\caption{\label{tab:methods-comparison-supp}Comparison of different detection
proposal methods.\protect \\
 ``sp'' indicates ``super pixels'' (of diverse type), \texttt{C{*}}
indicates variants of \texttt{CPMC}, and $\emptyset$ indicates ``no
low-level component''. \protect \\
Sg indicates that the method outputs segments, Bx that it outputs
only bounding boxes. F~indicates ``fix set'', Fs ``fix sorted
set'', R ``as requested'', and LR ``less than requested''. \protect \\
Time is indicated in seconds. \protect \\
Repeatability, quality, and detection rankings are provided as very
rough, subjective, guidelines; ``-'' indicates no data, ``$\cdot$'',
``$\star$'', ``$\star\star$'', ``$\star\star\star$'' indicate
progressively better results.}
\end{table}

\section{\label{sec:Controlling-num-windows}Controlling the average number
of detection proposals}

This section gives the details on how exactly we run the different
detection proposal methods and how we obtain the desired number of
proposals.
\begin{description}
\item [{\texttt{gPbUCM}}] has no code available.
\item [{\texttt{Objectness}}] scores every detection proposal, so for selecting
a subset, we use the highest scored ones.
\item [{\texttt{CPMC}}] has no parameter that directly controls the number
of produced proposals, but provides a score. We use the highest scoring
proposals if we need less than were produced.
\item [{\texttt{Endres2010}}] has no parameter to control the number of
proposals. The list is, however, presorted so we use the first proposals
to select subsets.
\item [{\texttt{Selective~Search}}] returns a randomized priority, which
can be used to select a subset of proposals. Low priorities are selected
first.
\item [{\texttt{Rahtu2011}}] provides a score that is not monotonically
decreasing in the order of the returned proposals. Using the first
$n$, instead of the best $n$ proposals resulted in better results.
\item [{\texttt{RandomizedPrim's}}] provides no scores, since the output
is the result of a sampling procedure, that takes learned probabilities
into account. After discussion with the authors, we request 20~000
proposals for the quality experiments and 5~000 proposals for the
repeatability experiments and filter by using those, that are returned
first.
\item [{\texttt{Bing}}] provide a score and we use the highest scoring
detection proposals.
\item [{\texttt{MCG}}] We use the first $n$ candidates. Sorting by the
returned score worsens results.
\item [{\texttt{Rantalankila2014}}] has no parameter that controls the
number of proposal directly. We run the method with both slic and
Felsenzwalb segmentation and control the number of proposals by adapting
the parameter \texttt{gc\_branches}. The parameter settings for the
different number of proposals were determined experimentally by running
the method with different parameters over 50 images and averaging
the number of proposals we get per image. If we need to filter we
use the first proposals.
\item [{\texttt{Rigor}}] No code available at time of writing.
\item [{\texttt{EdgeBoxes}}] scores all candidates. We use highest scoring
candidates first.
\item [{\texttt{Uniform}}] proposals are sampled independently, so we can
sample 10~000 proposals and use the first ones if we need a subset.
\item [{\texttt{Gaussian}}] same as for Uniform.
\item [{\texttt{SlidingWindow}}] always gives at most as many proposals
as we ask for, returns no scores, and the samples are not independent.
So we rerun the method and ask for an increasing amount of proposals
to produce the quality curve.
\item [{\texttt{Superpixels}}] runs the Felsenzwalb segmentation method
with 4 different parameters, like Randomized Prims does, and returns
the bounding boxes of every superpixel. If we need to filter, we use
the first proposals. This could, on principle, hurt the repeatability
performance, but on average we get 771 proposals on the Pascal test
set, so this is not an issue.
\end{description}

\section{\label{sec:Perturbations}Perturbations for repeatability}

In this section, we explain the details of the image perturbations
and give examples.

\subsection{Scale}

We uniformly sample the scale factor in scale space, from one octave
down to one octave up, i.e.~0.5 to 2. Do get an idea of the drop
around the identity transform, we add the scaling factors 0.9, 0.95,
0.99, 1.01, 1.05, and 1.1. The scaling operations have been done with
the default parameter setting of Matlab's imresize, which means upscaling
with bicubic interpolation and downscaling with antialiasing.

\subsection{Blur}

\begin{figure}[t]
\begin{centering}
\hspace*{\fill}\subfloat[\label{fig:blur-sigma-1}$\sigma=1$]{\begin{centering}
\includegraphics[width=0.25\textwidth]{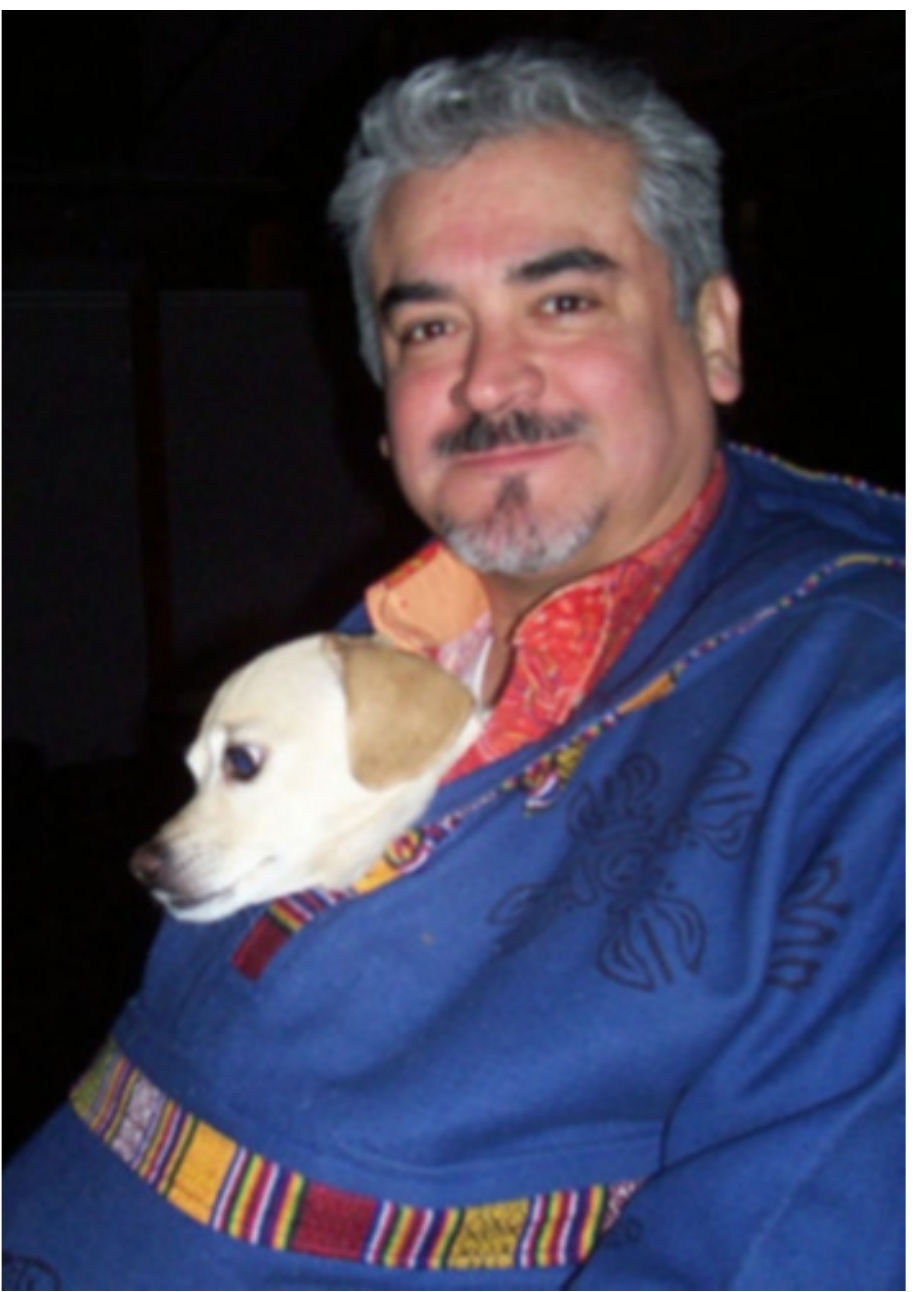}
\par\end{centering}

}\hspace*{\fill}\subfloat[\label{fig:blur-sigma-4}$\sigma=4$]{\begin{centering}
\includegraphics[width=0.25\textwidth]{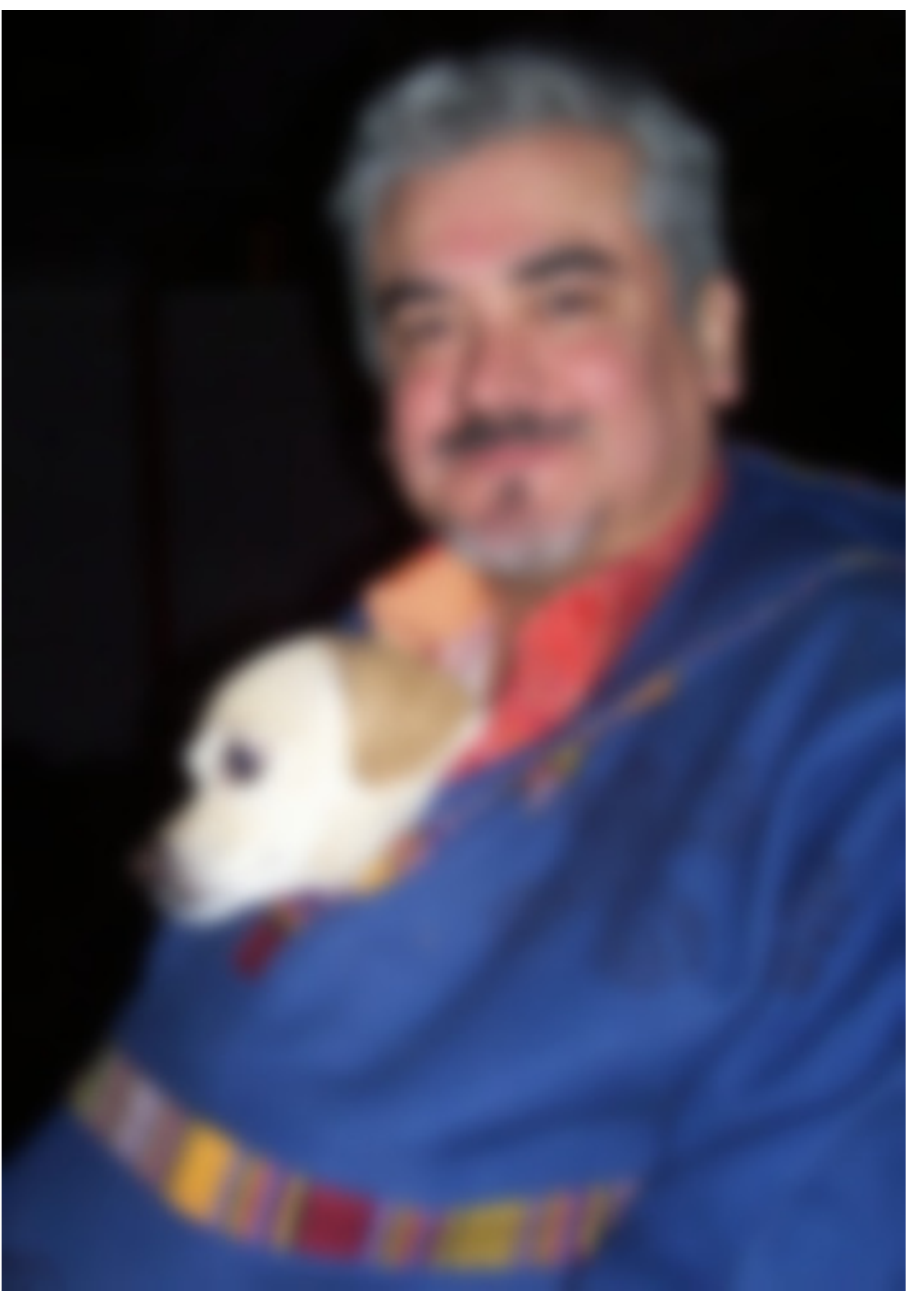}
\par\end{centering}

}\hspace*{\fill}\subfloat[\label{fig:blur-sigma-8}$\sigma=8$]{\begin{centering}
\includegraphics[width=0.25\textwidth]{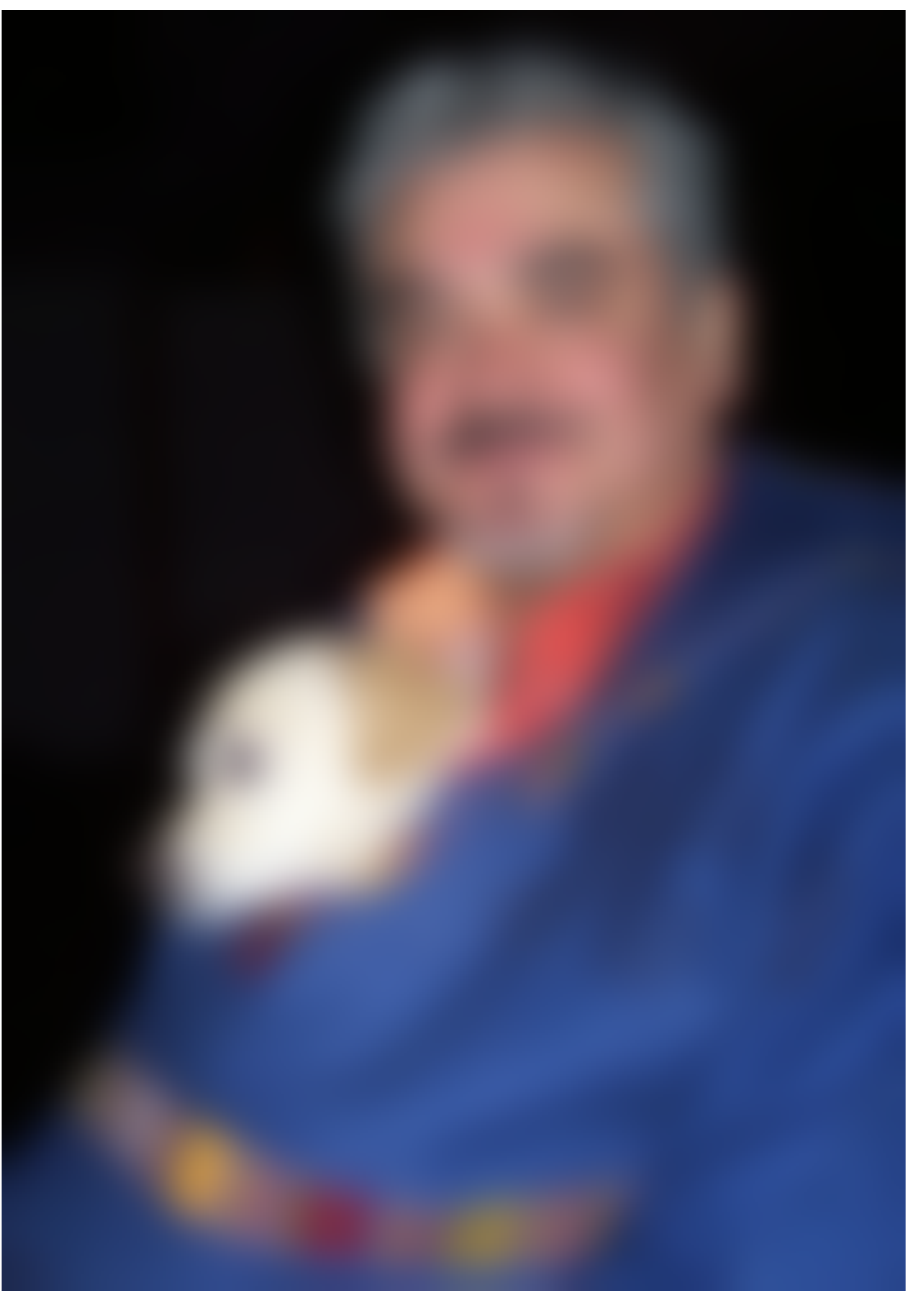}
\par\end{centering}

}\hspace*{\fill}
\par\end{centering}

\medskip{}

\protect\caption{\label{fig:blur-examples}Image blurring examples for different values
of $\sigma$.}
\end{figure}

We blur the images with a Gaussian kernel with different standard
deviations $\sigma$. In detail, we construct a Gaussian kernel of
size $20\cdot\sigma$ with the Matlab function fspecial. We use imfilter
with symmetric image padding to get an output image of same size.
Figure~\ref{fig:blur-examples} shows some examples.

\subsection{Rotation}

\begin{figure}[t]
\begin{centering}
\hspace*{\fill}\subfloat[\label{fig:rotation-20-degrees}The biggest rectangle with the same
aspect ratio as the original image, that can fit into the image with
20 degrees rotation.]{\begin{centering}
\includegraphics[width=0.3\textwidth]{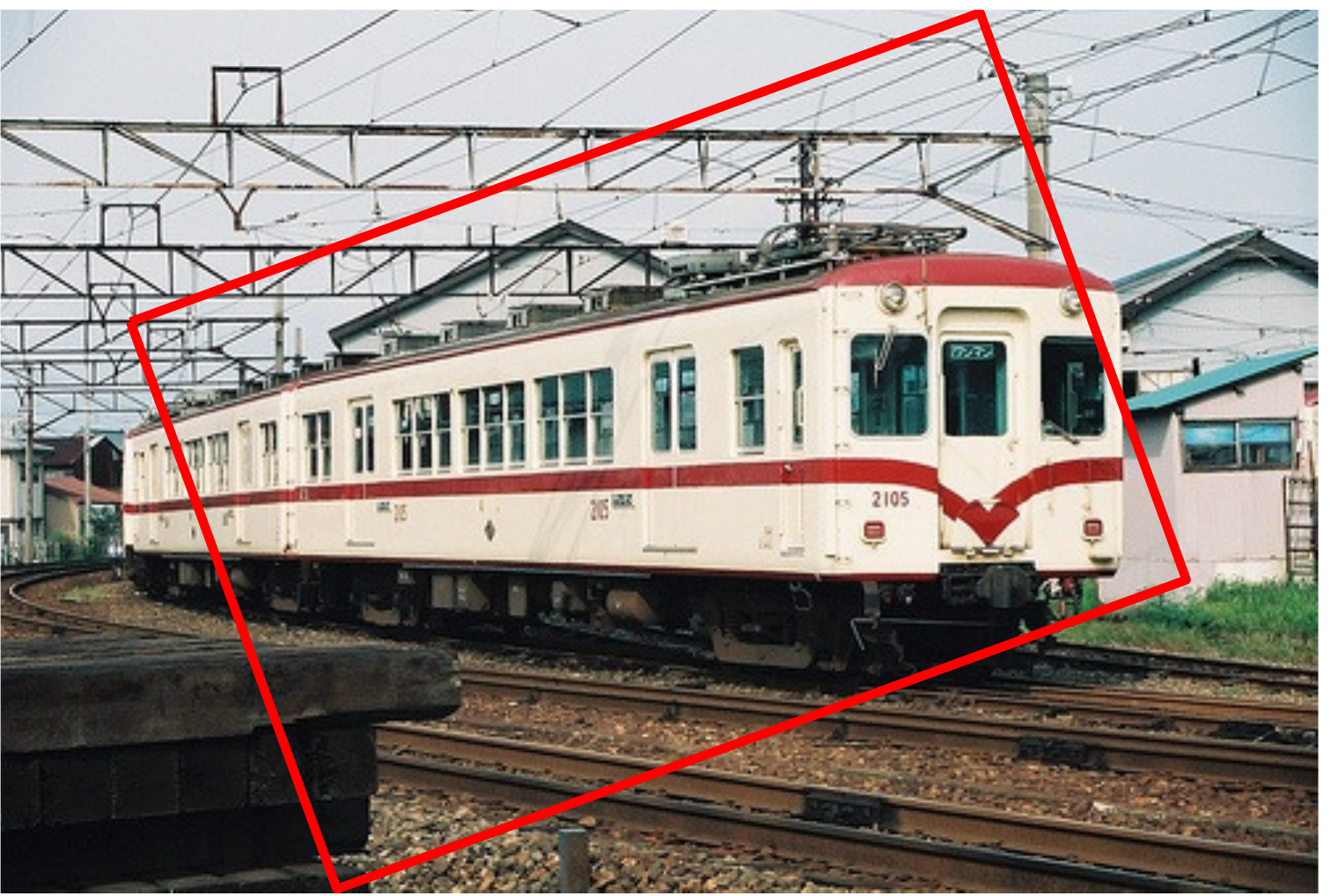}
\par\end{centering}

}\hspace*{\fill}\subfloat[\label{fig:rotation-20-degrees-crop}The resulting crop from \ref{fig:rotation-20-degrees}.]{\begin{centering}
\includegraphics[width=0.3\textwidth]{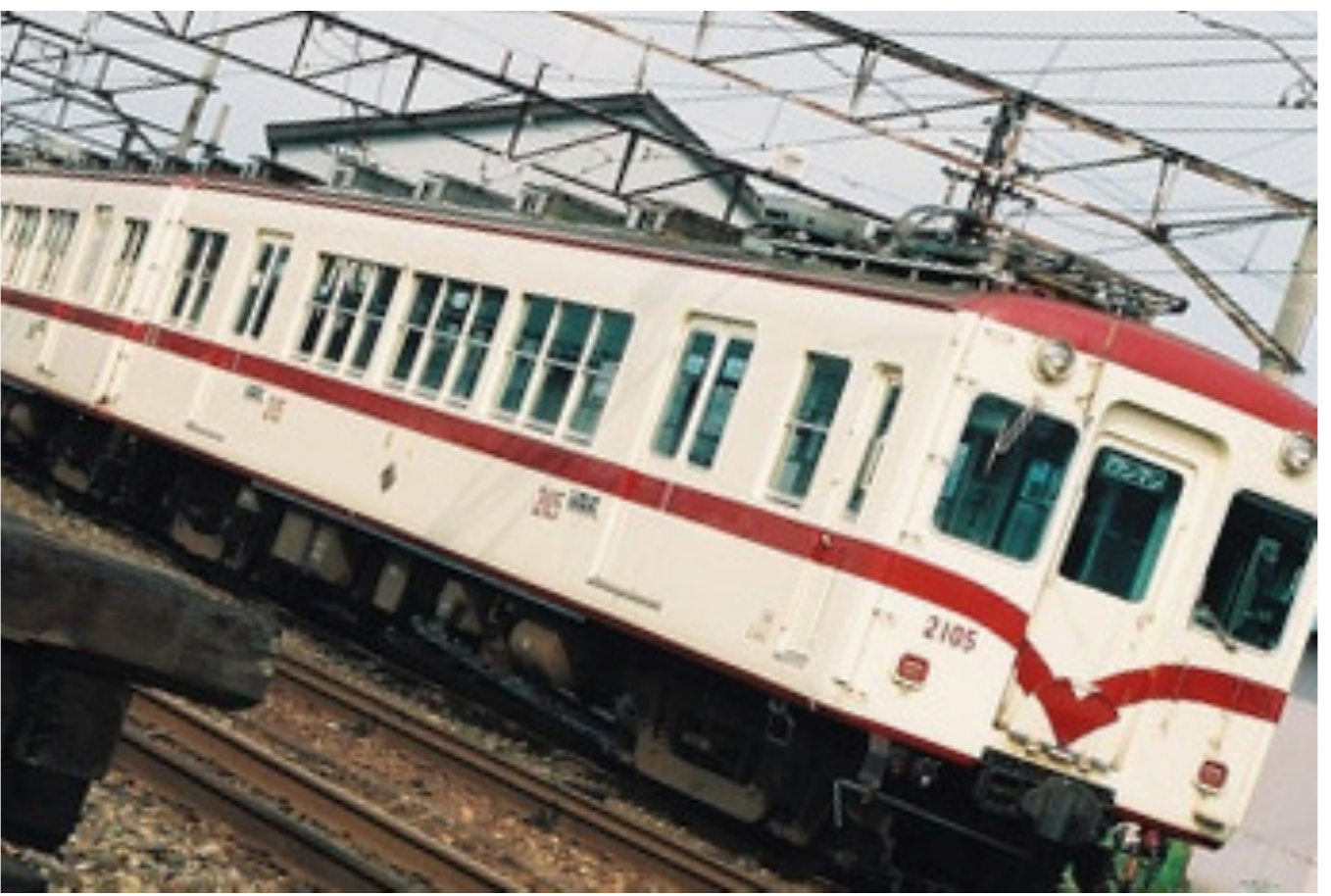}
\par\end{centering}

}\hspace*{\fill}
\par\end{centering}

\begin{centering}
\hspace*{\fill}\subfloat[\label{fig:rotation-5-degrees}Example rotation with of -5 degrees.]{\begin{centering}
\includegraphics[width=0.3\textwidth]{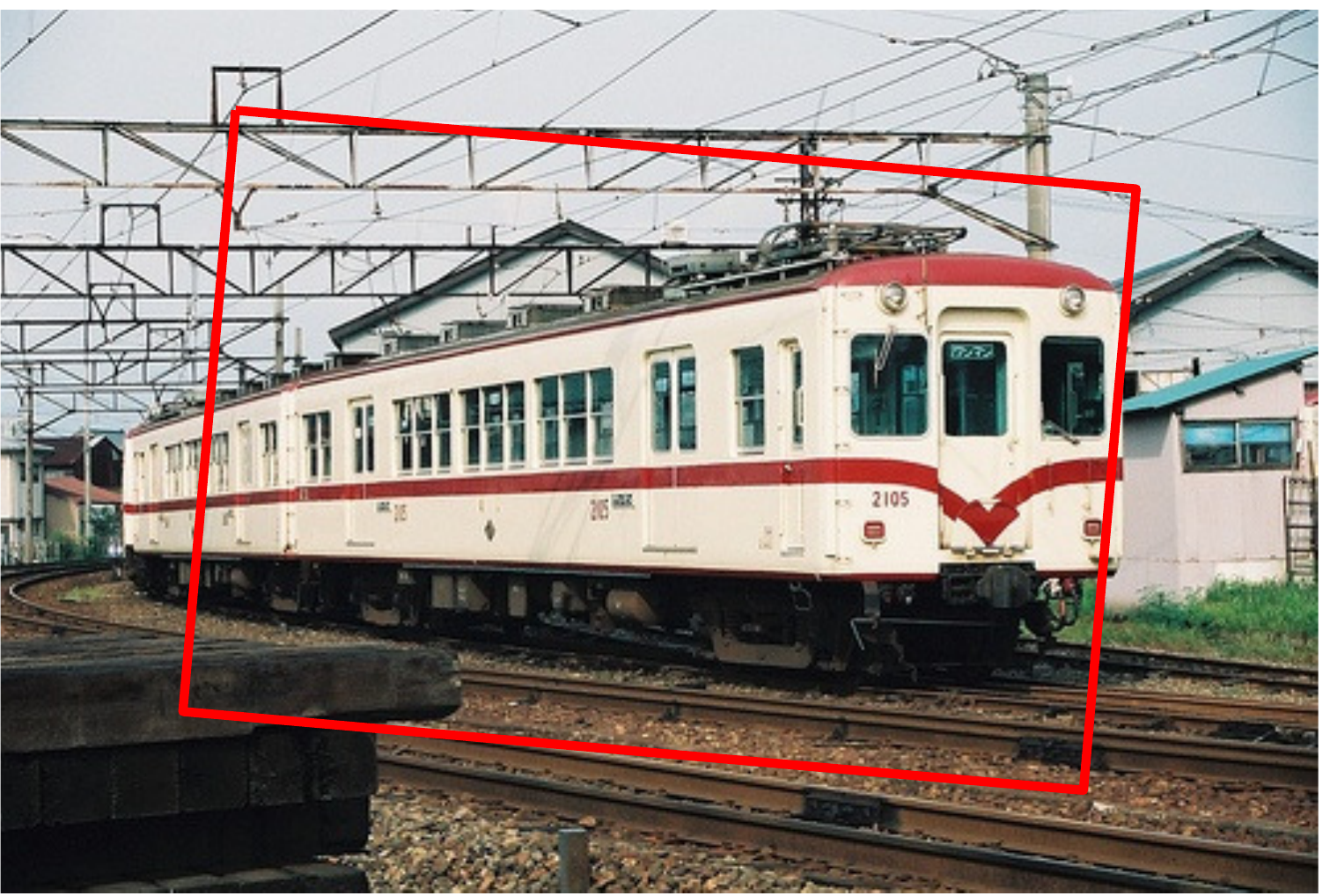}
\par\end{centering}

}\hspace*{\fill}\subfloat[\label{fig:rotation-5-degrees-crop}The resulting crop of \ref{fig:rotation-5-degrees}.]{\begin{centering}
\includegraphics[width=0.3\textwidth]{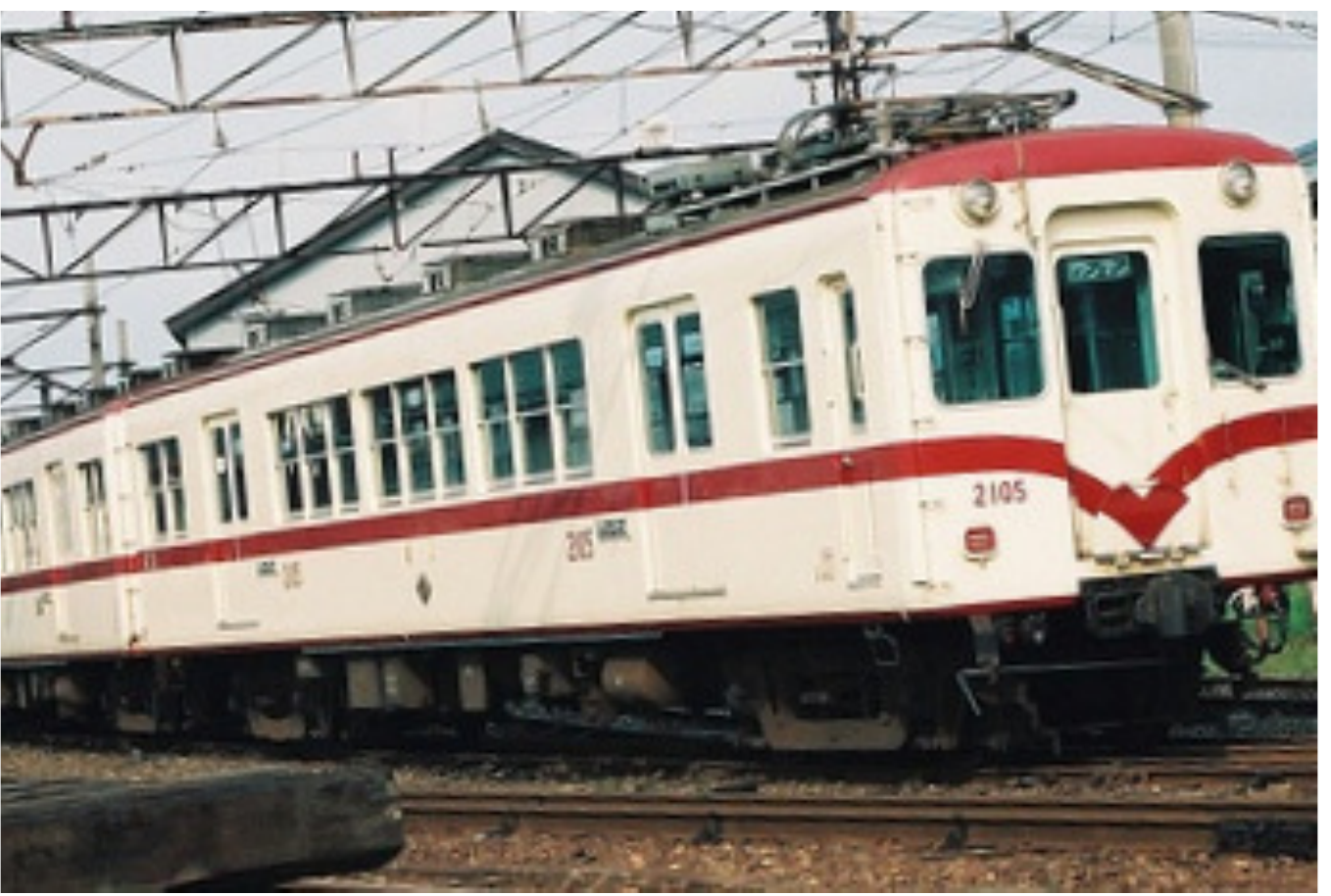}
\par\end{centering}

}\hspace*{\fill}
\par\end{centering}

\vspace{0.5em}

\protect\caption{\label{fig:rotation-examples}Examples of how the rotation transformation
is defined.}
\end{figure}

We rotate the image in 5 degree steps between -20 and 20 degrees.
To avoid both 1)~padding the image with fake image content and 2)~having
too many detection proposals in areas that are cropped by the transformation,
we do the following: We first construct the biggest bounding box with
the same aspect ratio as the original image that can fit into the
original image with the most extreme setting of rotation. Figure~\ref{fig:rotation-20-degrees}
shows such a rectangle on an example image, while the resulting cropped
can be seen in figure~\ref{fig:rotation-20-degrees-crop}. This defeats
problem 1). To limit problem 2), we use the size of the previous crop
for all rotations, even if there is enough content to make the crop
bigger. See figure~\ref{fig:rotation-5-degrees} and \ref{fig:rotation-5-degrees-crop}
for an example.

\subsection{Illumination}

\begin{figure}[th]
\begin{centering}
\hspace*{\fill}\subfloat[\label{fig:illumination-darkest}Illumination change to 50\%.]{\begin{centering}
\includegraphics[width=0.25\textwidth]{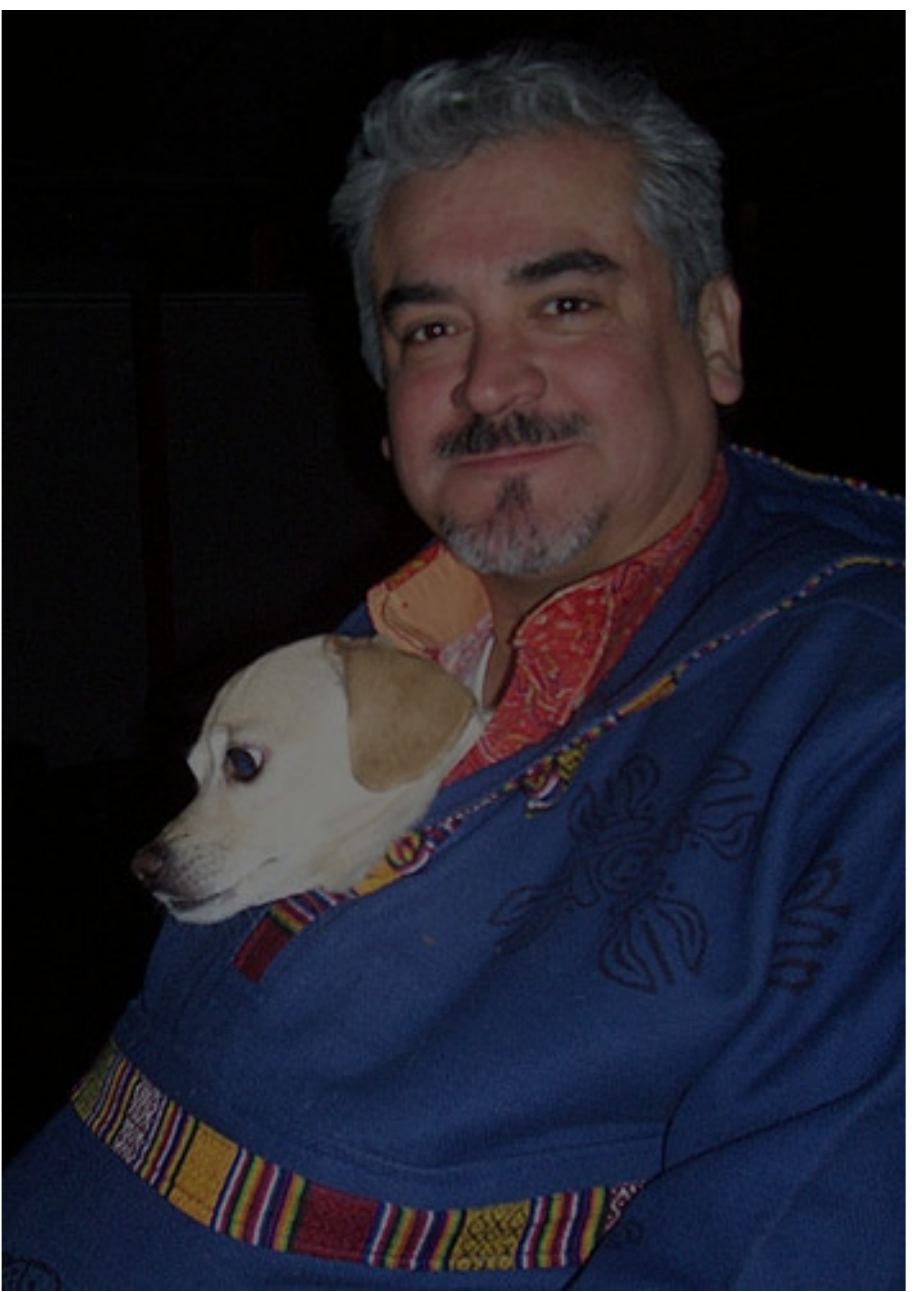}
\par\end{centering}

}\hspace*{\fill}\subfloat[\label{fig:illumination-lightest}Illumination change to 150\%.]{\begin{centering}
\includegraphics[width=0.25\textwidth]{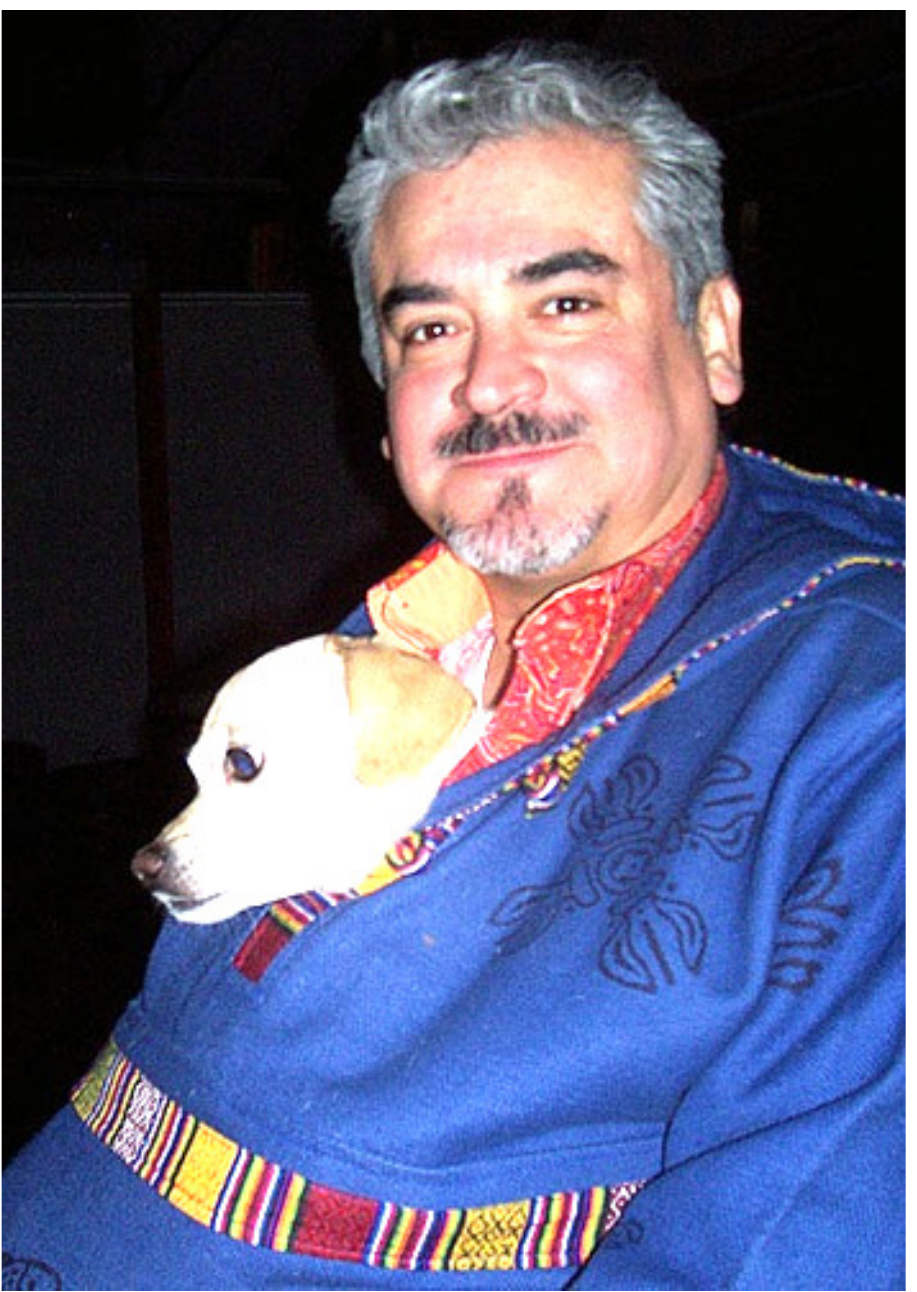}
\par\end{centering}

}\hspace*{\fill}
\par\end{centering}

\vspace{0.5em}

\protect\caption{\label{fig:illumination-examples}Extreme examples for illumination
changes.}
\end{figure}

To synthetically change the illumination of the images we changed
the brightness channel in the HSB color space. We do that with the
imagemagick library. We chose the extremes of the transformation so
that we observe some over and under saturation, as can be seen in
figure~\ref{fig:illumination-examples}.

\subsection{JPEG artefacts}

\begin{figure}[th]
\begin{centering}
\hspace*{\fill}\subfloat[\label{fig:jpeg-50}50\% quality]{\begin{centering}
\includegraphics[width=0.25\textwidth]{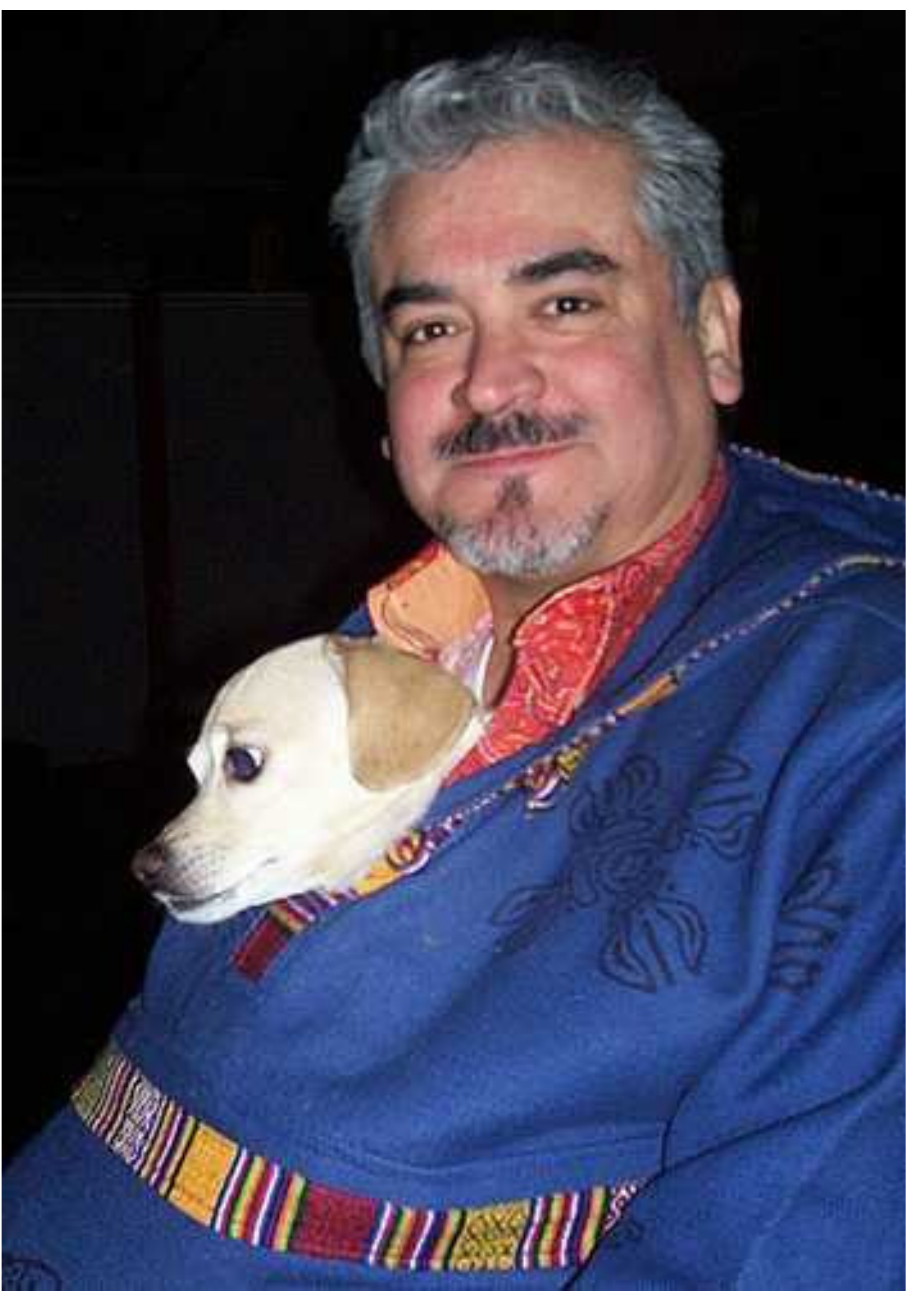}
\par\end{centering}

}\hspace*{\fill}\subfloat[\label{fig:jpeg-10}10\% quality]{\begin{centering}
\includegraphics[width=0.25\textwidth]{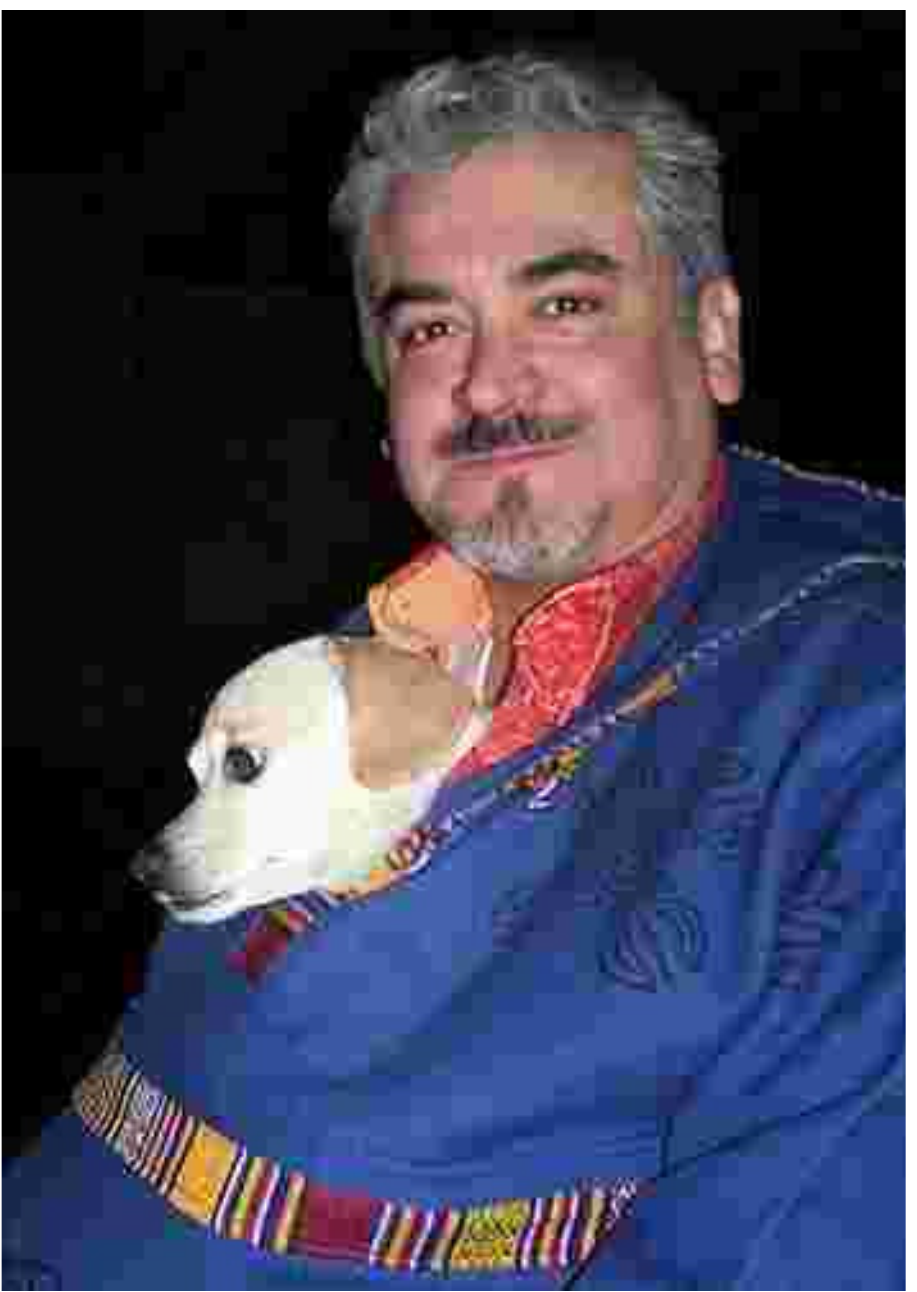}
\par\end{centering}

}\hspace*{\fill}\subfloat[\label{fig:jpeg-5}5\% quality]{\begin{centering}
\includegraphics[width=0.25\textwidth]{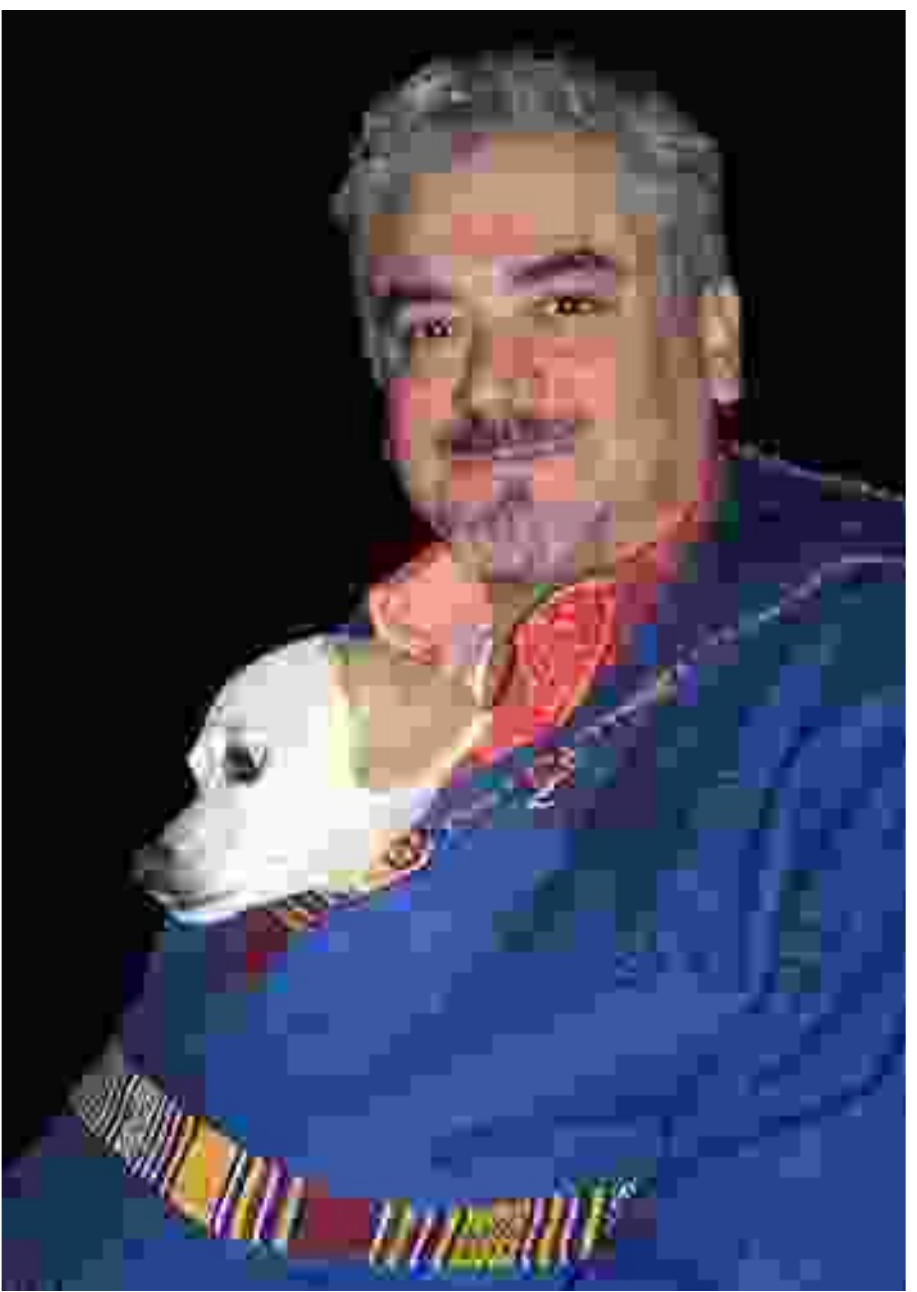}
\par\end{centering}

}\hspace*{\fill}
\par\end{centering}

\vspace{0.5em}

\protect\caption{\label{fig:jpeg-examples}JPEG compression examples.}
\end{figure}

To create JPEG artefacts we write the image to disk with the Matlab
function imwrite, which supports JPEG compression and a quality parameter.
We try very different quality parameters including very low ones and
and the highest setting, 100\%. We also include a lossless parameter
(no image berturbation) setting for comparison.

Figure~\ref{fig:jpeg-examples} shows the low end of the quality
spectrum, as only that is actually easily visible for humans. When
doing a pixel wise comparison between a JPEG compressed image and
the original, we see differences on the entire image, even for the
100\% quality setting.

\section{Repeatability results}

\begin{figure}[h]
\begin{centering}
\hspace*{\fill}\subfloat[\label{fig:distribution-of-windows-sizes-supp}Histogram of proposal
windows sizes for different methods on Pascal VOC 2007 test.]{\begin{centering}
\includegraphics[width=0.29\textwidth]{figures/candidate_size_histogram}
\par\end{centering}

}\hspace*{\fill}\subfloat[\label{fig:fluctuation-per-size-group-supp}Example of recall fluctuation
for different size group (blur perturbation). Each group corresponds
to one point in figure \ref{fig:distribution-of-windows-sizes-supp}.]{\begin{centering}
\includegraphics[width=0.3\textwidth]{figures/explain_repeatability_selective_search}
\par\end{centering}

}\hspace*{\fill}
\par\end{centering}

\begin{centering}
\hspace*{\fill}\subfloat[\label{fig:repeatiblity-scale-change-supp}Repeatability under scale
changes. x-axis in log scale from scale half to double of the original
image size.]{\begin{centering}
\includegraphics[width=0.3\textwidth]{figures/repeatability_scale}
\par\end{centering}

}\hspace*{\fill}\subfloat[\label{fig:repeatiblity-rotation-change-supp}Repeatability under
rotation. x-axis shows rotations from $-20$ to $20$ degrees.]{\begin{centering}
\includegraphics[width=0.3\textwidth]{figures/repeatability_rotate}
\par\end{centering}

}\hspace*{\fill}
\par\end{centering}

\begin{centering}
\hspace*{\fill}\subfloat[\label{fig:repeatiblity-blur-change}Repeatability under blurring.
x-axis shows the gaussian kernel standard deviation from $0$ to 8
pixels.]{\begin{centering}
\includegraphics[width=0.3\textwidth]{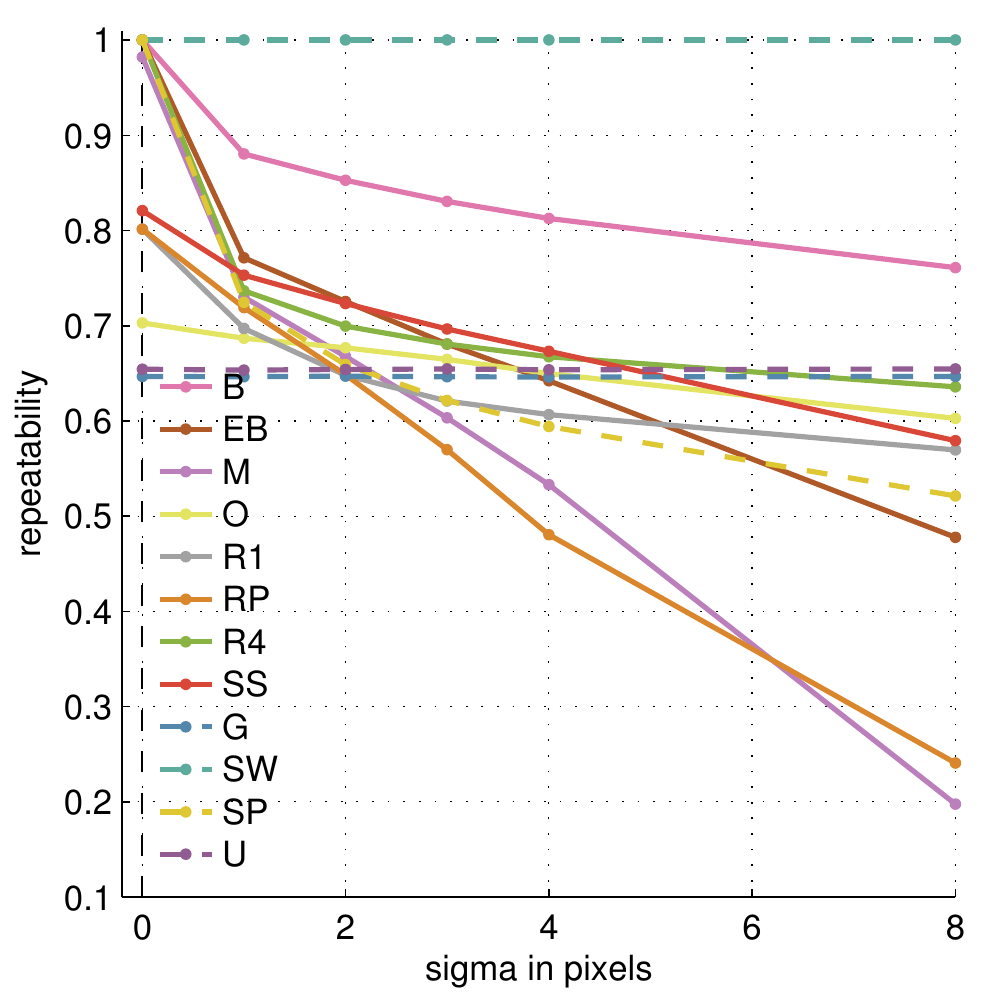}
\par\end{centering}

}\hspace*{\fill}\subfloat[\label{fig:repeatiblity-illumination-change-supp}Repeatability under
illumination. x-axis shows the changes from dark ($50$), to neutral
($100$), to over-exposed image ($150$).]{\begin{centering}
\includegraphics[width=0.3\textwidth]{figures/repeatability_light}
\par\end{centering}

}\hspace*{\fill}\subfloat[\label{fig:repeatiblity-blur-change-3}Repeatability under JPEG artefacts.
x-axis shows the quality preservation from $100\ \%$ (no change)
to $5\%$ (obvious artefacts).]{\begin{centering}
\includegraphics[width=0.3\textwidth]{figures/repeatability_jpeg}
\par\end{centering}

}\hspace*{\fill}\vspace{0.5em}

\par\end{centering}

\protect\caption{\label{fig:repeatability-supp}Repeatability results. }
\end{figure}
We omitted the repeatability plot of the blurring perturbation from
the paper due to space constrains. It can be found in figure~\ref{fig:repeatiblity-blur-change}.
For completeness we repeat all other repeatability results in figure~\ref{fig:repeatability-supp}.

\section{\label{sec:Proposals-quality}Proposal recall}

\subsection{Pascal VOC 2007}

\begin{figure}
\begin{centering}
\hspace*{\fill}\subfloat[\label{fig:recall-vs-iuo-at-100-windows-1}Recall versus IoU threshold,
for $100$ proposals per image. Curves are labelled with number of
proposals/area under recall curve.]{\begin{centering}
\includegraphics[width=0.3\textwidth]{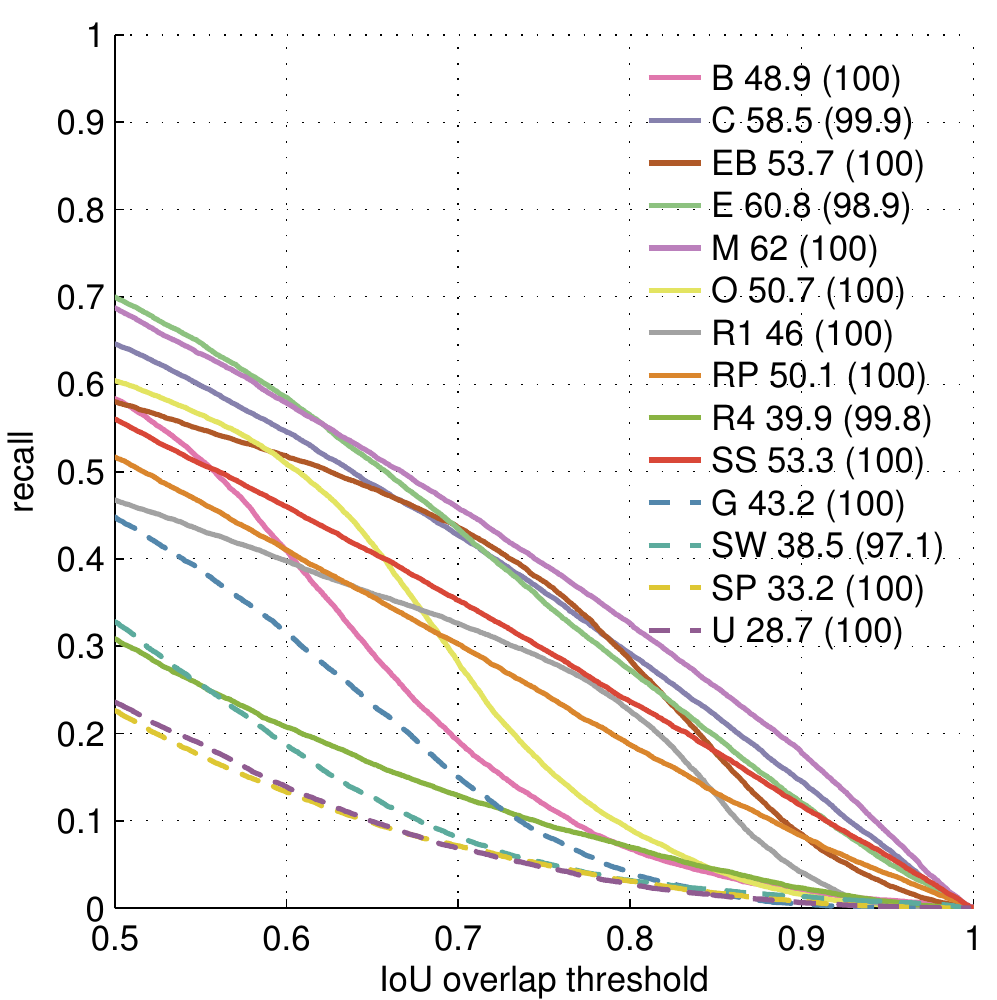}
\par\end{centering}

}\hspace*{\fill}\subfloat[\label{fig:recall-vs-iuo-at-1000-windows-1}Recall versus IoU threshold,
for $1\,000$ proposals per image. ]{\begin{centering}
\includegraphics[width=0.3\textwidth]{figures/recall_1000}
\par\end{centering}

}\hspace*{\fill}\subfloat[\label{fig:recall-vs-iuo-at-10000-windows-1}Recall versus IoU threshold,
for $10\,000$ proposals per image. ]{\begin{centering}
\includegraphics[width=0.3\textwidth]{figures/recall_10000}
\par\end{centering}

}\hspace*{\fill}
\par\end{centering}

\begin{centering}
\hspace*{\fill}\subfloat[\label{fig:recall-vs-iuo-area-vs-number-of-proposals-pascal-supp}Area
under ``recall versus IoU threshold'' curves, for varying number
proposed windows.]{\begin{centering}
\includegraphics[width=0.3\textwidth]{figures/num_candidates_area_under_recall}
\par\end{centering}

}\hspace*{\fill}\subfloat[\label{fig:recall-at-iou-0.5-vs-number-of-proposals-pascal-supp}Recall
at IoU above 0.5 versus number of proposed windows.]{\begin{centering}
\includegraphics[width=0.3\textwidth]{figures/num_candidates_recall_0\lyxdot 5}
\par\end{centering}

}\hspace*{\fill}\subfloat[\label{fig:recall-at-iou-0.8-vs-number-of-proposals-pascal-supp}Recall
at IoU above 0.8 versus number of proposed windows.]{\begin{centering}
\includegraphics[width=0.3\textwidth]{figures/num_candidates_recall_0\lyxdot 8}
\par\end{centering}

}\hspace*{\fill}
\par\end{centering}

\vspace{0.5em}

\protect\caption{\label{fig:pascal-quality-results}Proposals quality over Pascal VOC
2007 test set. On recall versus IoU threshold curves, number indicates
area under the curve, and number in parentheses obtained average number
of windows per image.}
\end{figure}
In the paper we analyse the recall as a function of how well the detection
proposals have to be localised. Another interesting view on the behaviour
of the methods is to plot the recall as a function of the number of
proposals we use. Doing this, we are faced with the problem, that
the IoU threshold has to be removed from the plot. One solution is
to plot the recall for a fixed IoU threshold (figure~\ref{fig:recall-at-iou-0.5-vs-number-of-proposals-pascal-supp}
and \ref{fig:recall-at-iou-0.8-vs-number-of-proposals-pascal-supp}).
An alternative is to plot the area under the ``recall versus IoU
thershold curve'', i.e.~the area under the curves in figure~\ref{fig:recall-vs-iuo-at-100-windows-1},~\ref{fig:recall-vs-iuo-at-1000-windows-1},
and~\ref{fig:recall-vs-iuo-at-10000-windows-1}. You can see the
area under those curves as the first number in the legend. In figure~\ref{fig:recall-vs-iuo-area-vs-number-of-proposals-pascal-supp},
you see the area under the curves as the function of the number of
proposals.

Note how the three different curves (figure~\ref{fig:recall-vs-iuo-area-vs-number-of-proposals-pascal-supp},~\ref{fig:recall-at-iou-0.5-vs-number-of-proposals-pascal-supp},
and~\ref{fig:recall-at-iou-0.8-vs-number-of-proposals-pascal-supp})
favour different methods. Picking one of the curves or showing only
part of a curve does not tell the full story and can be misleading.

\subsection{ImageNet}

All methods were trained on Pascal and also tuned towards the test
set of Pascal it is unclear how they perform on a different dataset.
Thus, we also evaluate the recall on the ImageNet Large Scale Visual
Recognition Challenge (ILSVRC2013) detection task, which has 200 classes
instead of 20 on Pascal. Albeit the difference in size and number
of classes, it has to be noted that the distribution of annotation
sizes (figure\,\ref{fig:distribution-of-windows-sizes-supp}), the
distribution of image sizes, and the average number of annotations
per image is quite similar.

We do the experiments on the ILSVRC2013 validation set (as the test
set annotations are not available). The ILSVRC detection evaluation
protocol blacklists some images for some classes, because of too much
ambiguity in the annotations. We follow this procedure and do not
count those annotations in the recall computation. This does not have
a negative impact on the curves because they ignore false positives.

\begin{figure}[h]
\begin{centering}
\hspace*{\fill}\subfloat[\label{fig:recall-vs-iuo-at-100-windows-imagenet}Recall versus IoU
threshold, for $100$ proposals per image.]{\begin{centering}
\includegraphics[width=0.3\textwidth]{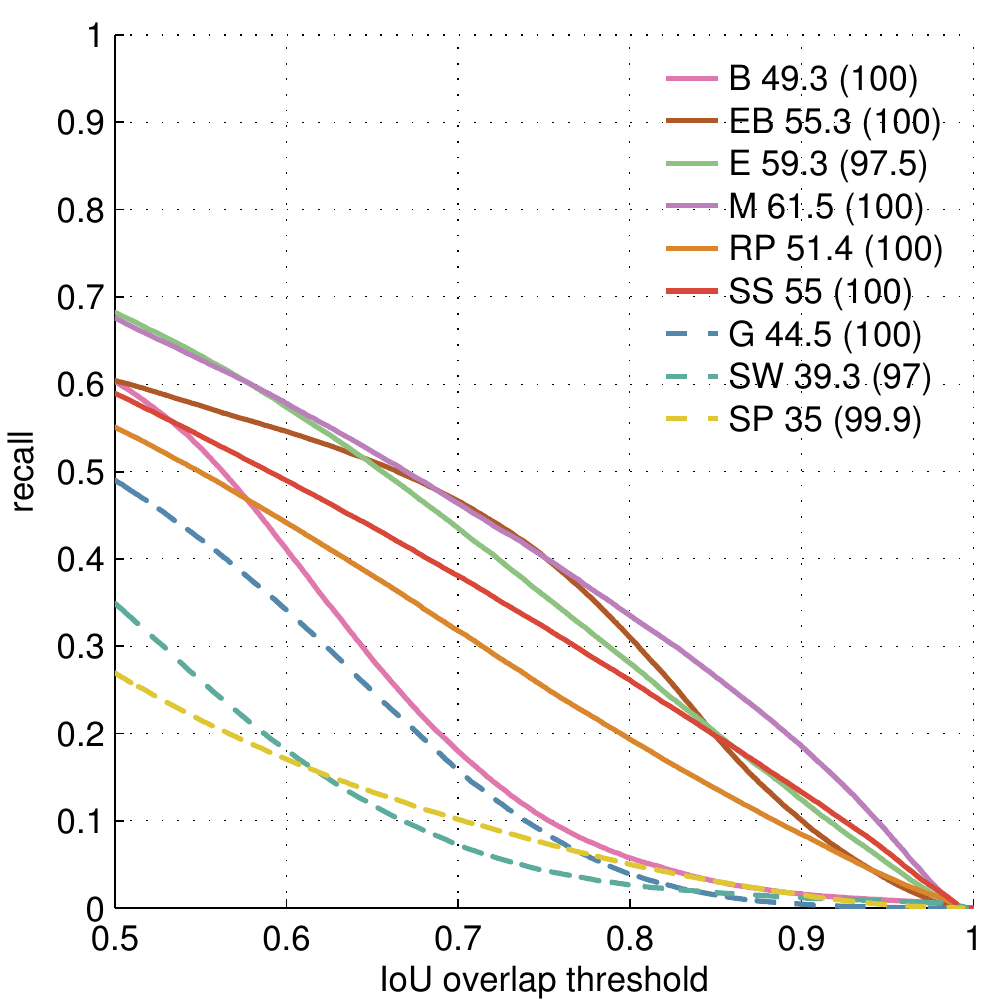}
\par\end{centering}

}\hspace*{\fill}\subfloat[\label{fig:recall-vs-iuo-at-1000-windows-imagenet-supp}Recall versus
IoU threshold, for $1\,000$ proposals per image. ]{\begin{centering}
\includegraphics[width=0.3\textwidth]{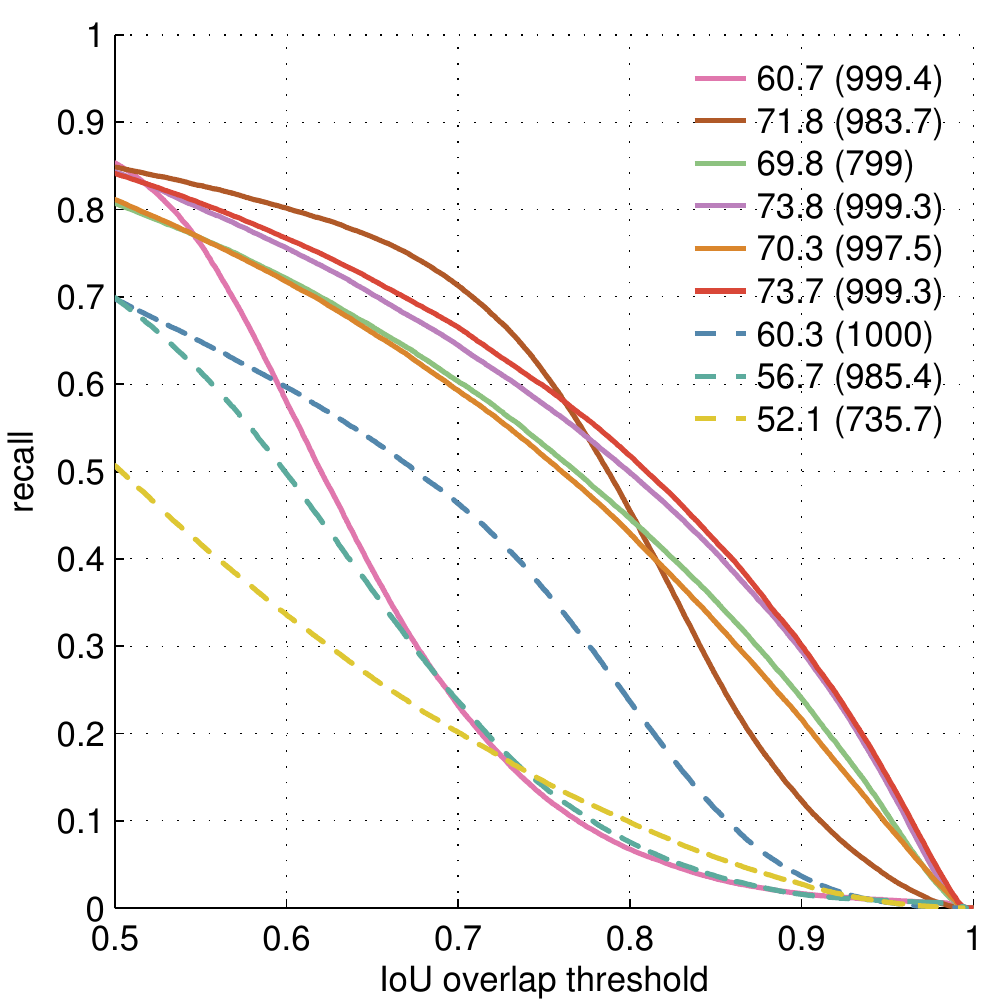}
\par\end{centering}

}\hspace*{\fill}\subfloat[\label{fig:recall-vs-iuo-at-10000-windows-imagenet}Recall versus
IoU threshold, for $10\,000$ proposals per image. ]{\begin{centering}
\includegraphics[width=0.3\textwidth]{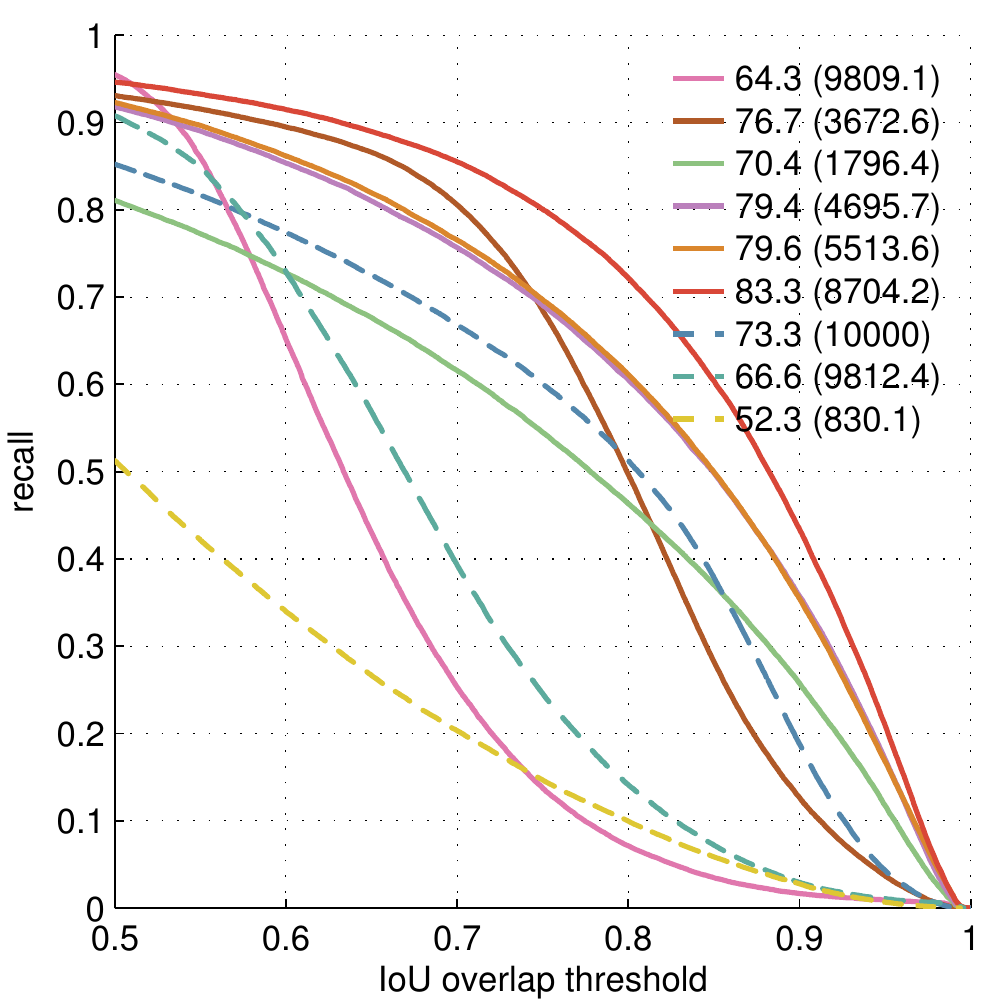}
\par\end{centering}

}\hspace*{\fill}
\par\end{centering}

\begin{centering}
\hspace*{\fill}\subfloat[\label{fig:recall-vs-iuo-area-vs-number-of-proposals-imagenet}Area
under the ``recall versus IoU threshold'' curves, for varying number
proposed windows.]{\begin{centering}
\includegraphics[width=0.3\textwidth]{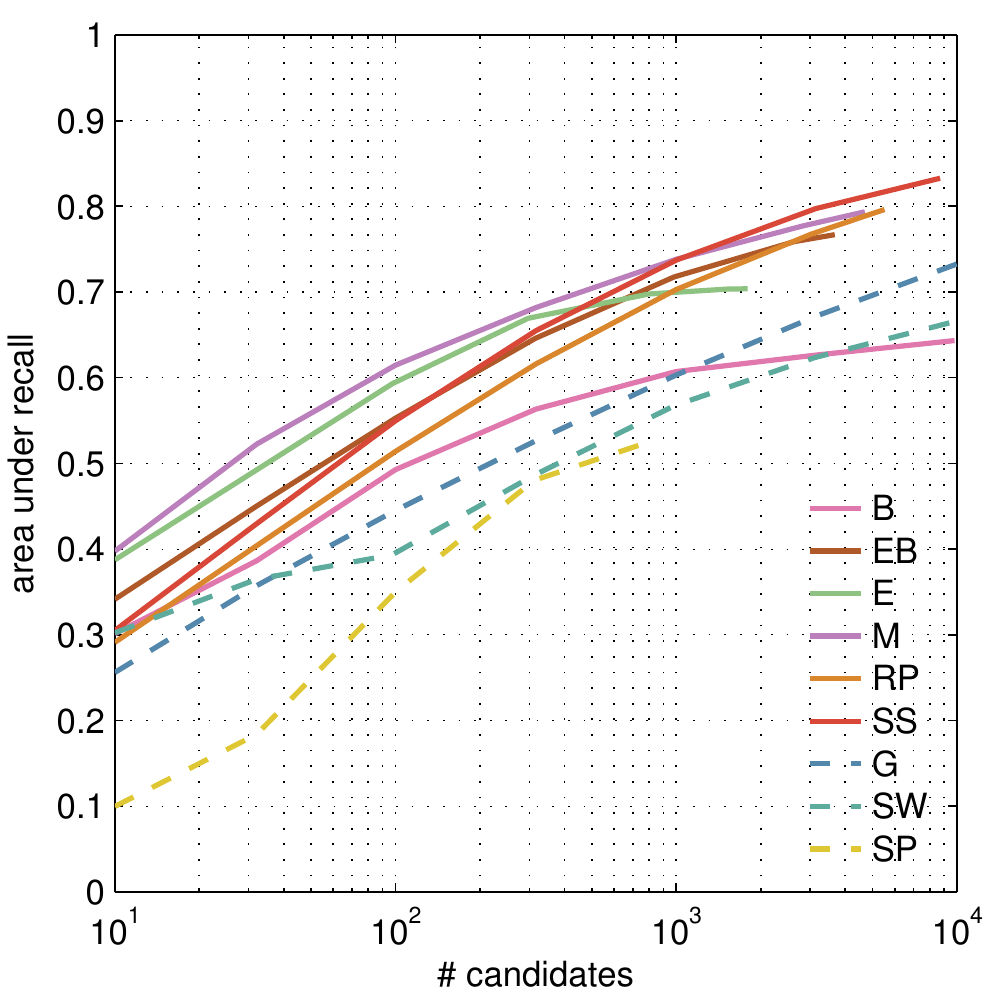}
\par\end{centering}

}\hspace*{\fill}\subfloat[\label{fig:recall-at-iou-0.5-vs-number-of-proposals-imagenet}Recall
at IoU above 0.5 versus number of proposed windows.]{\begin{centering}
\includegraphics[width=0.3\textwidth]{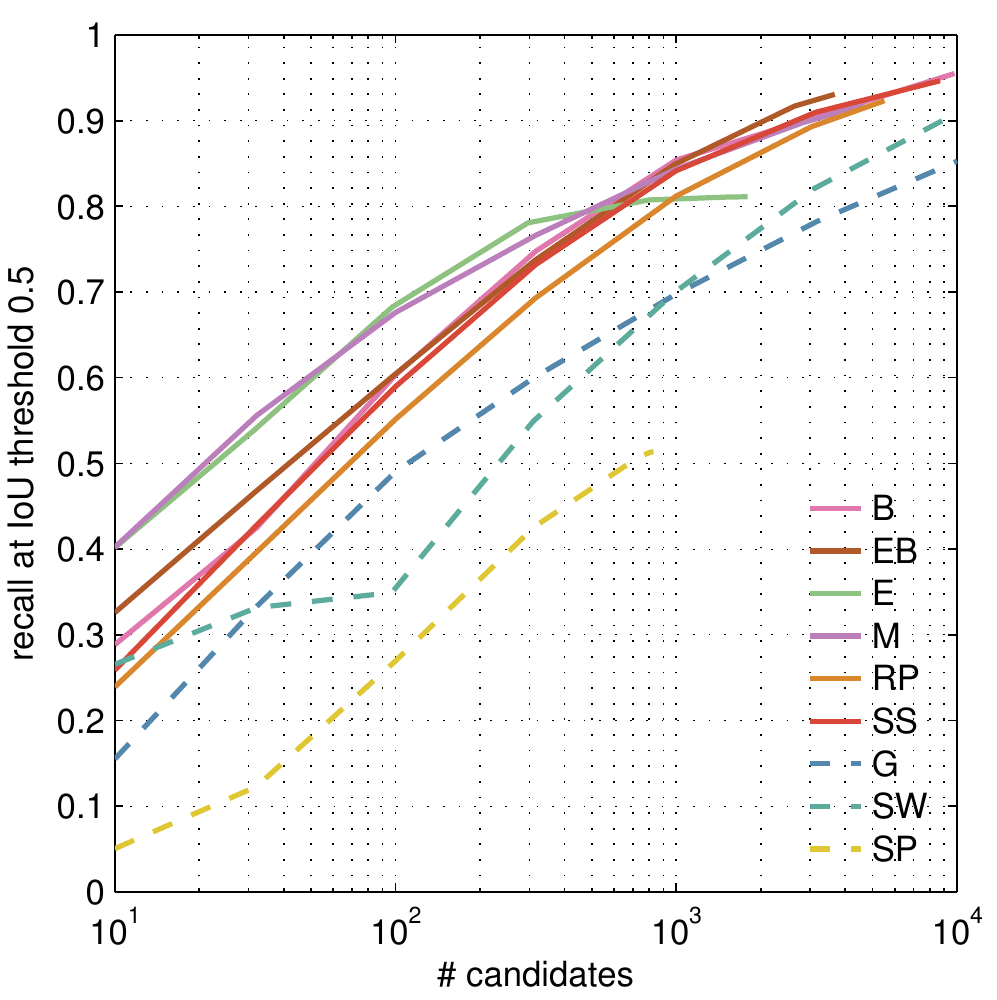}
\par\end{centering}

}\hspace*{\fill}\subfloat[\label{fig:recall-at-iou-0.8-vs-number-of-proposals-imagenet}Recall
at IoU above 0.8 versus number of proposed windows.]{\begin{centering}
\includegraphics[width=0.3\textwidth]{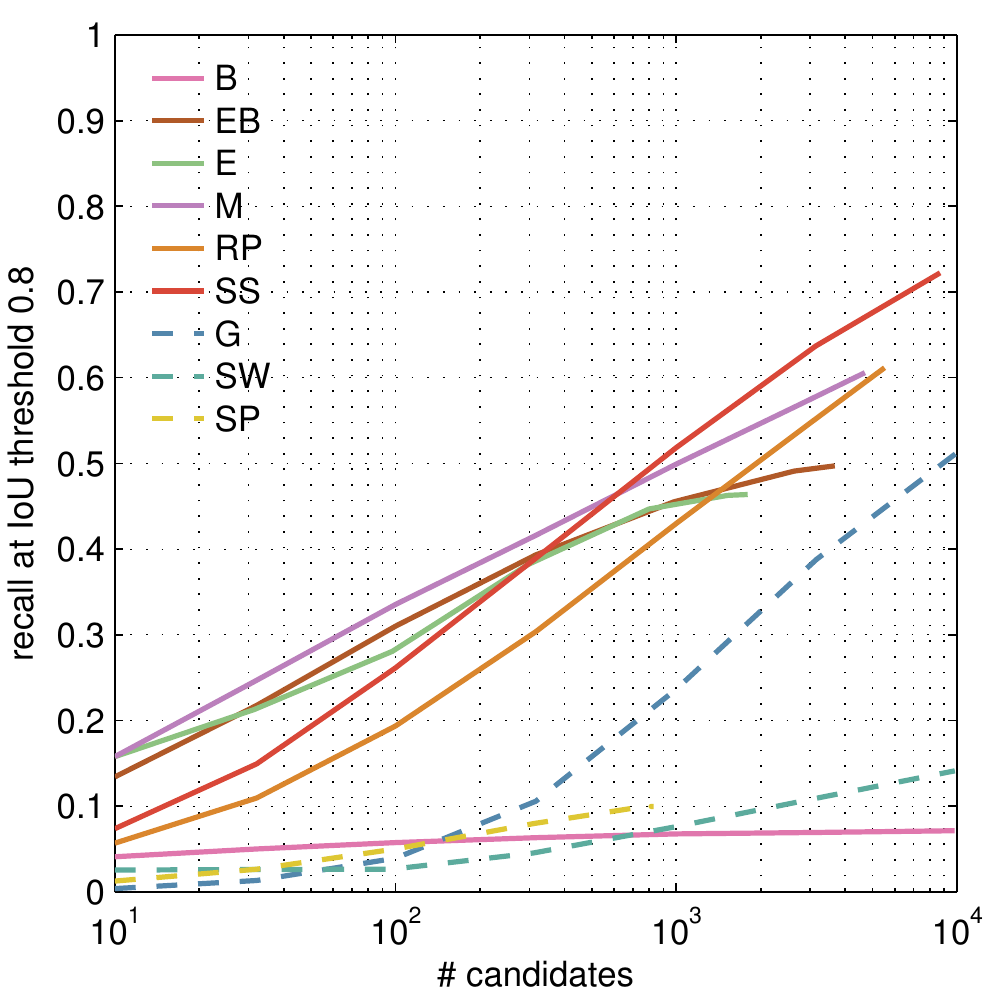}
\par\end{centering}

}\hspace*{\fill}
\par\end{centering}

\vspace{0.5em}

\protect\caption{\label{fig:imagenet-quality-results-supp}Proposals quality over ImageNet
2013 validation set. On recall versus IoU threshold curves, number
indicates area under the curve, and number in parentheses obtained
average number of proposals per image.}
\end{figure}

Figure~\ref{fig:imagenet-quality-results-supp} shows the recall
results of on the ILSVRC2013 detection validation set.

\section{Detection}

\begin{figure}[h]
\begin{centering}
\hspace*{\fill}\subfloat{\centering{}\includegraphics[width=0.45\textwidth]{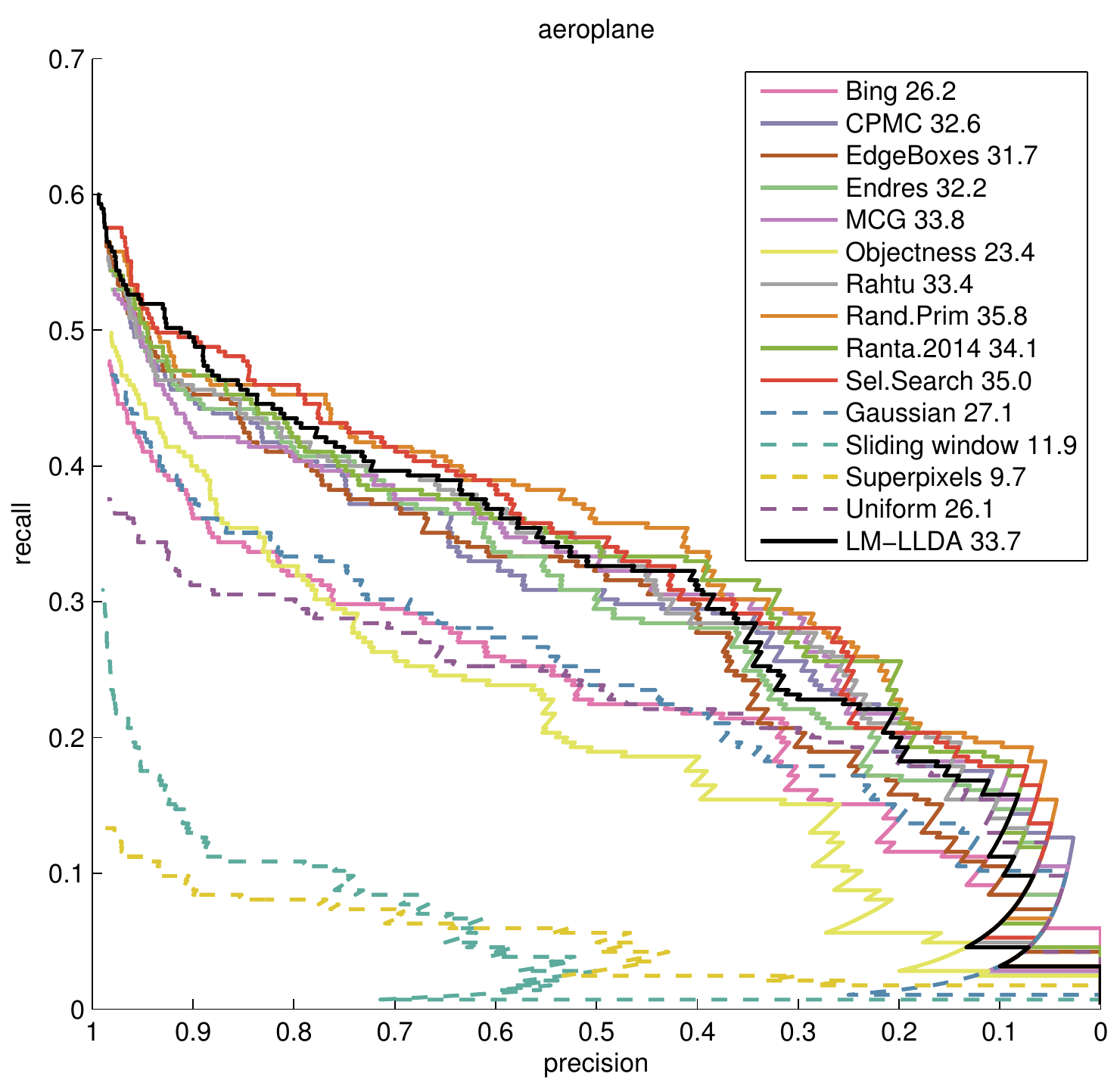}}\hspace*{\fill}\subfloat{\centering{}\includegraphics[width=0.45\textwidth]{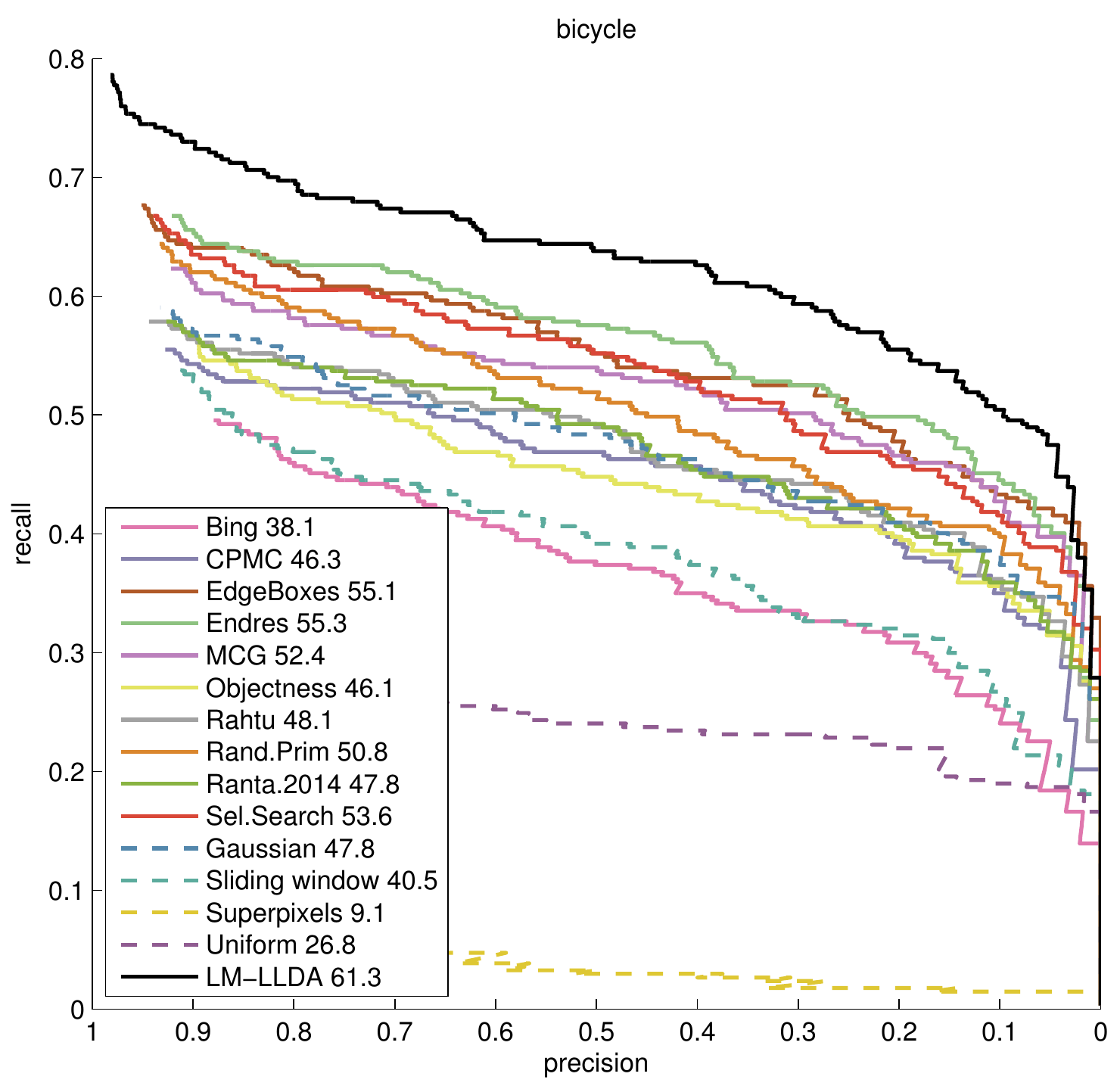}}\hspace*{\fill}
\par\end{centering}

\begin{centering}
\hspace*{\fill}\subfloat{\centering{}\includegraphics[width=0.45\textwidth]{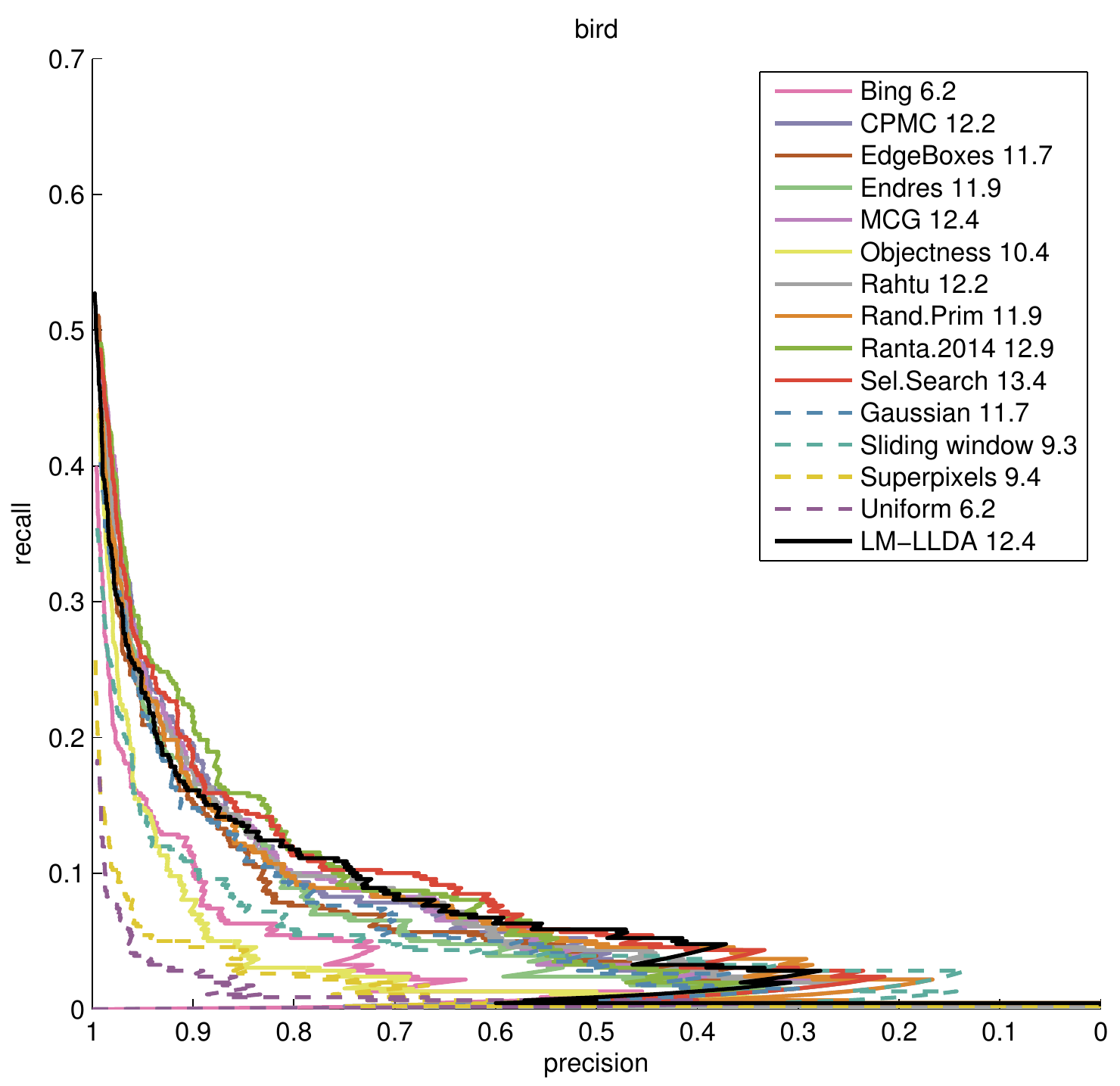}}\hspace*{\fill}\subfloat{\centering{}\includegraphics[width=0.45\textwidth]{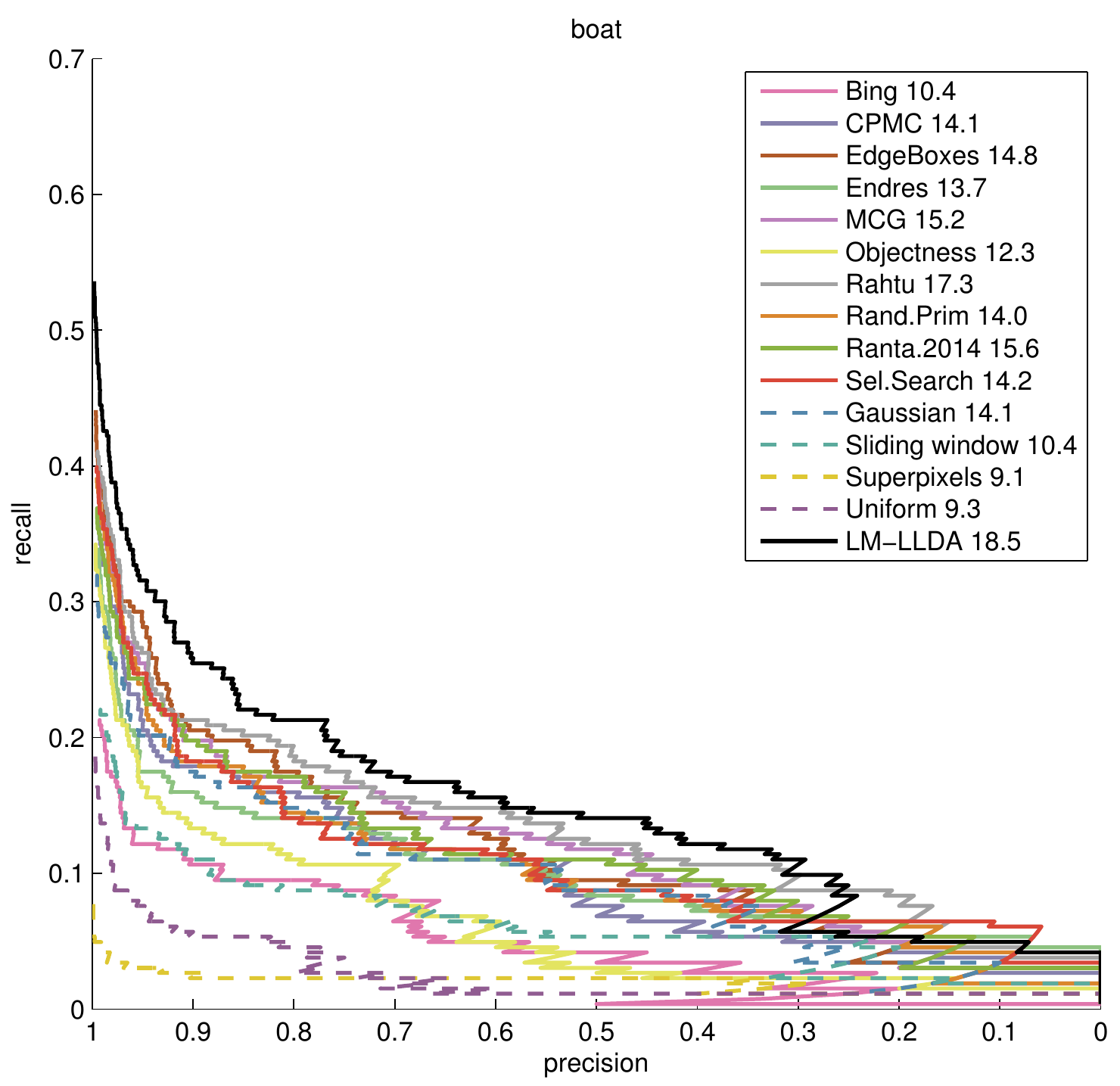}}\hspace*{\fill}
\par\end{centering}

\begin{centering}
\hspace*{\fill}\subfloat{\centering{}\includegraphics[width=0.45\textwidth]{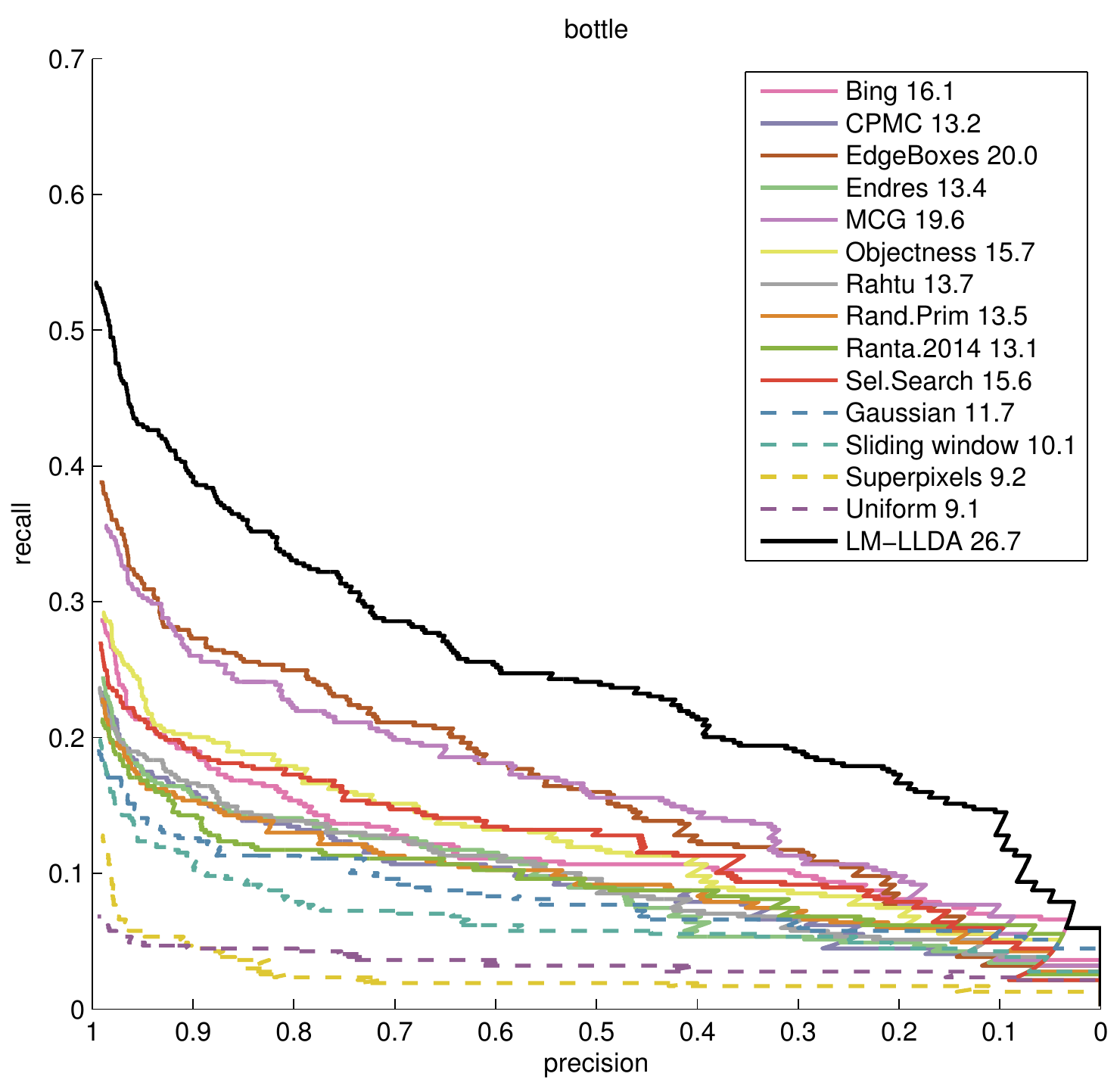}}\hspace*{\fill}\subfloat{\centering{}\includegraphics[width=0.45\textwidth]{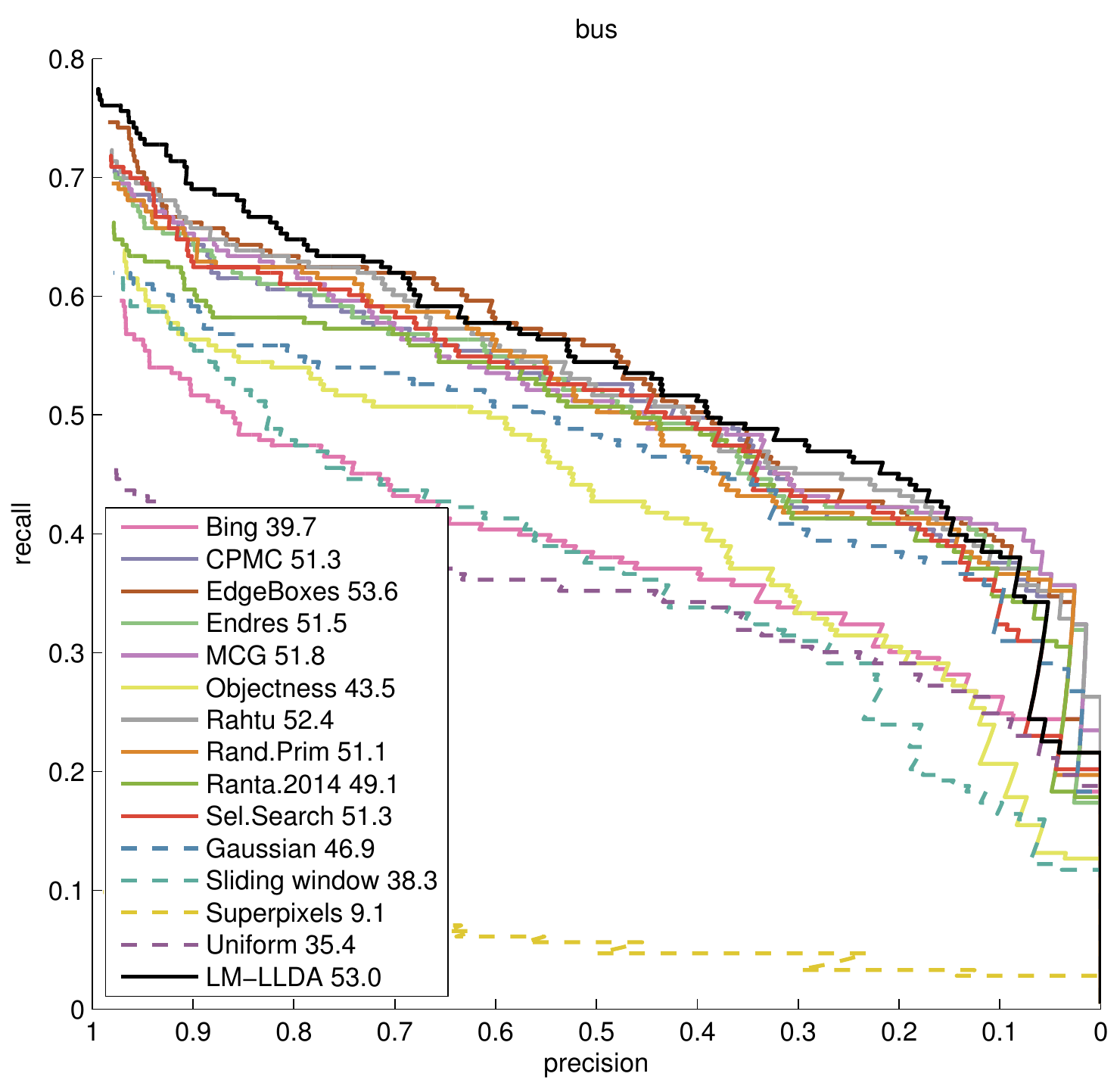}}\hspace*{\fill}
\par\end{centering}

\vspace{0.5em}

\protect\caption{\label{fig:pascal-detection-1-supp}Recall-precision curves for Pascal
VOC 2007 using different proposal methods at test time.}
\end{figure}

\begin{figure}[h]
\begin{centering}
\hspace*{\fill}\subfloat{\centering{}\includegraphics[width=0.45\textwidth]{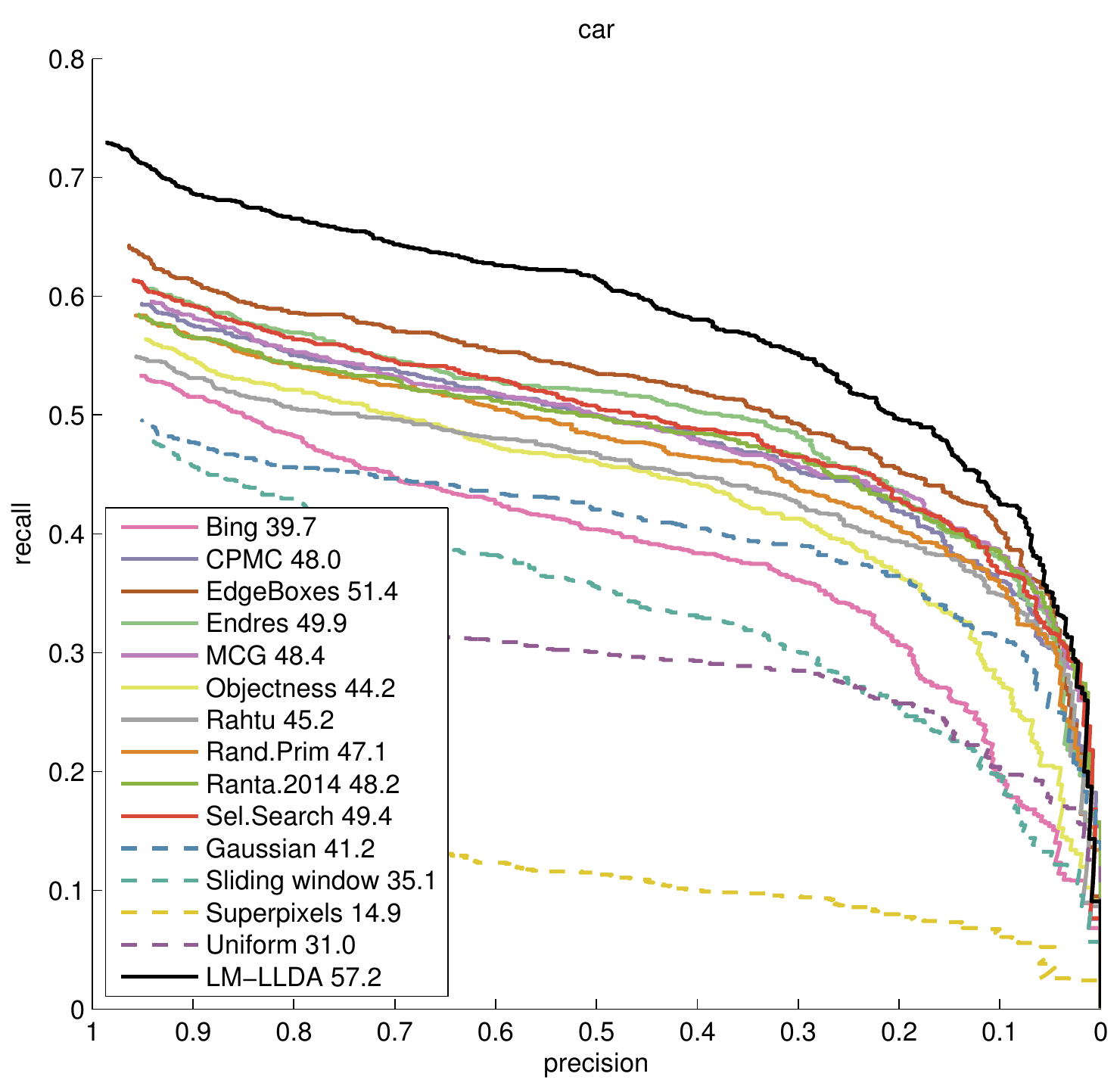}}\hspace*{\fill}\subfloat{\centering{}\includegraphics[width=0.45\textwidth]{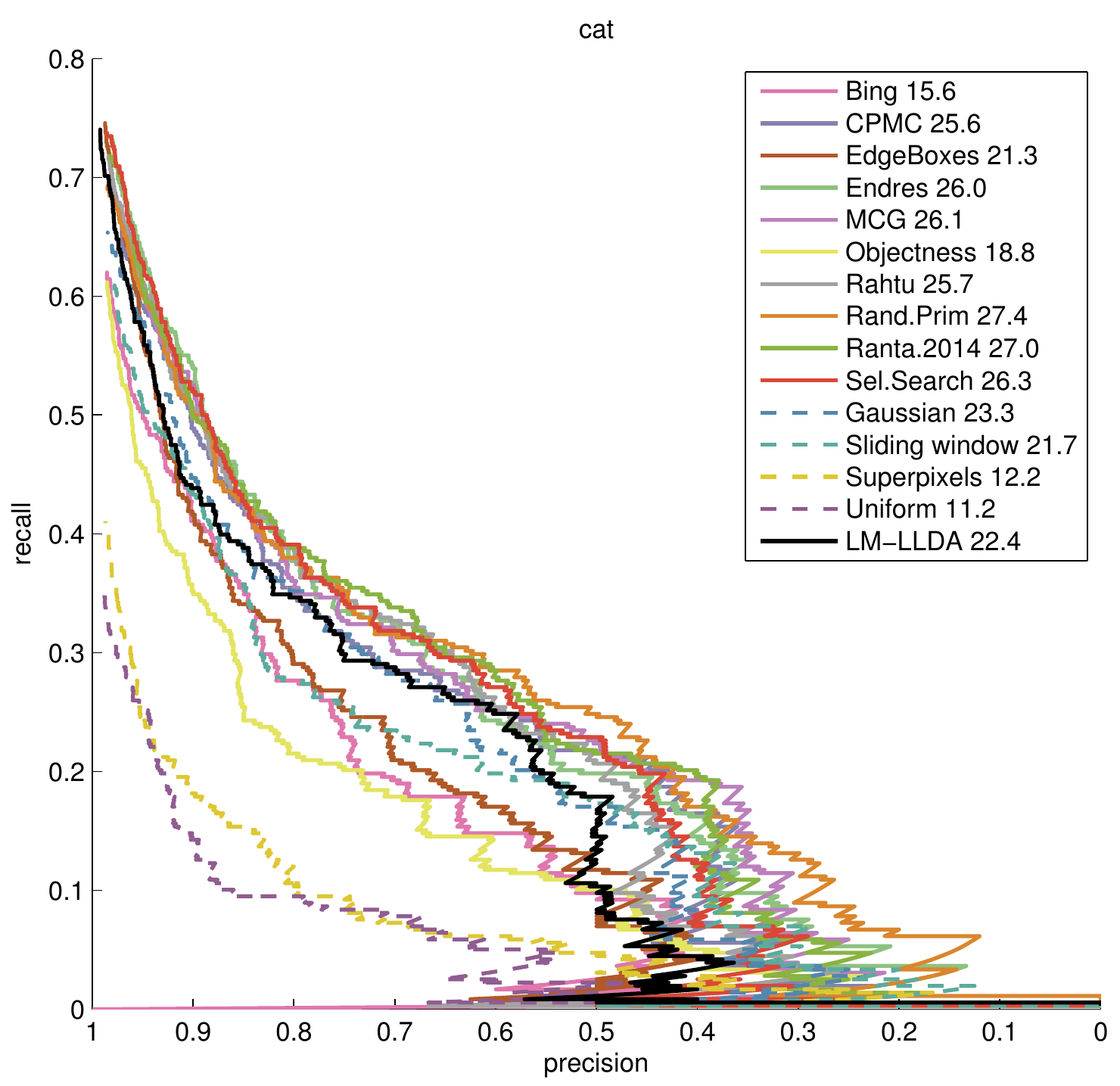}}\hspace*{\fill}
\par\end{centering}

\begin{centering}
\hspace*{\fill}\subfloat{\centering{}\includegraphics[width=0.45\textwidth]{figures/pascal_recall_precision_bird}}\hspace*{\fill}\subfloat{\centering{}\includegraphics[width=0.45\textwidth]{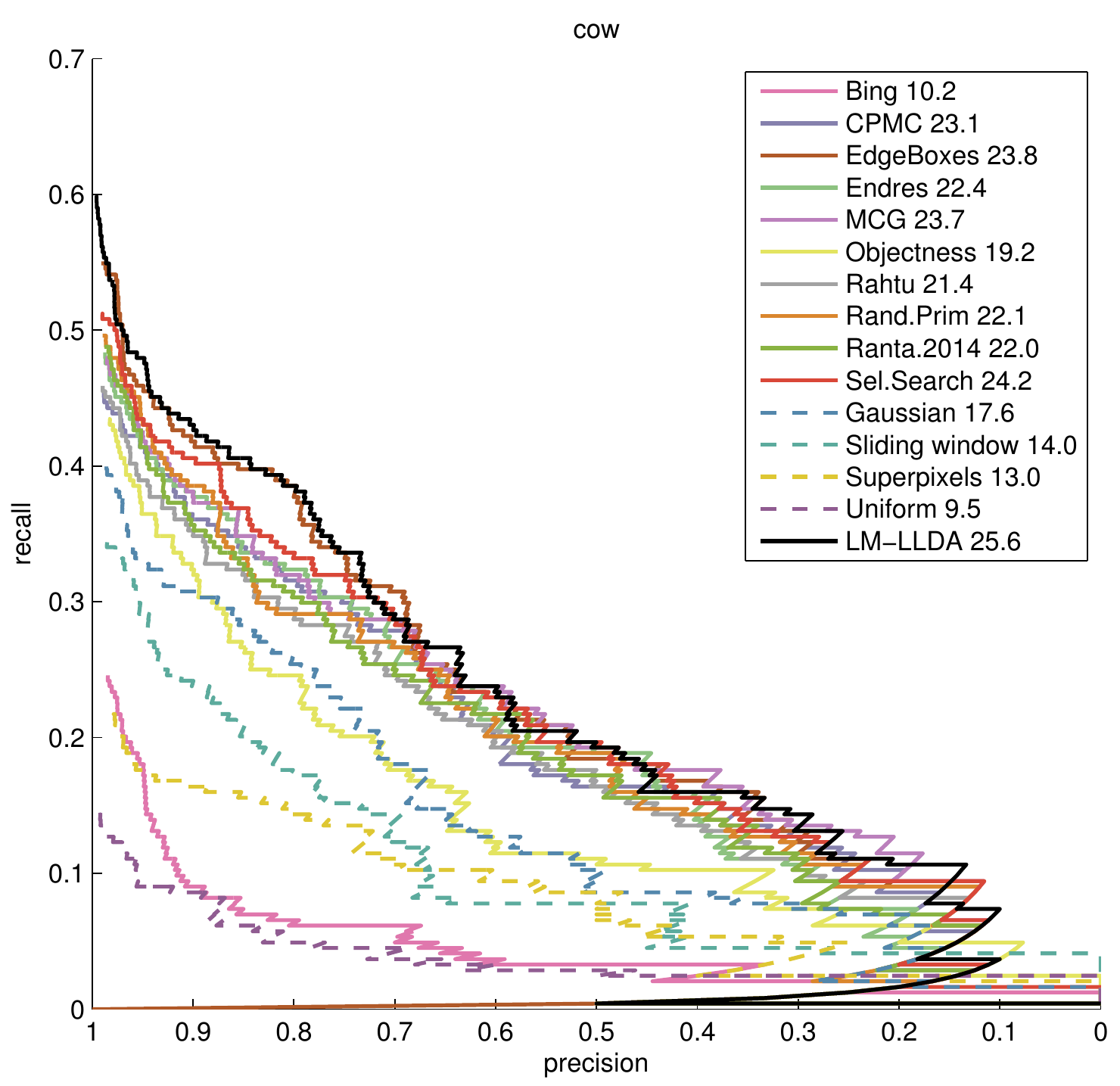}}\hspace*{\fill}
\par\end{centering}

\begin{centering}
\hspace*{\fill}\subfloat{\centering{}\includegraphics[width=0.45\textwidth]{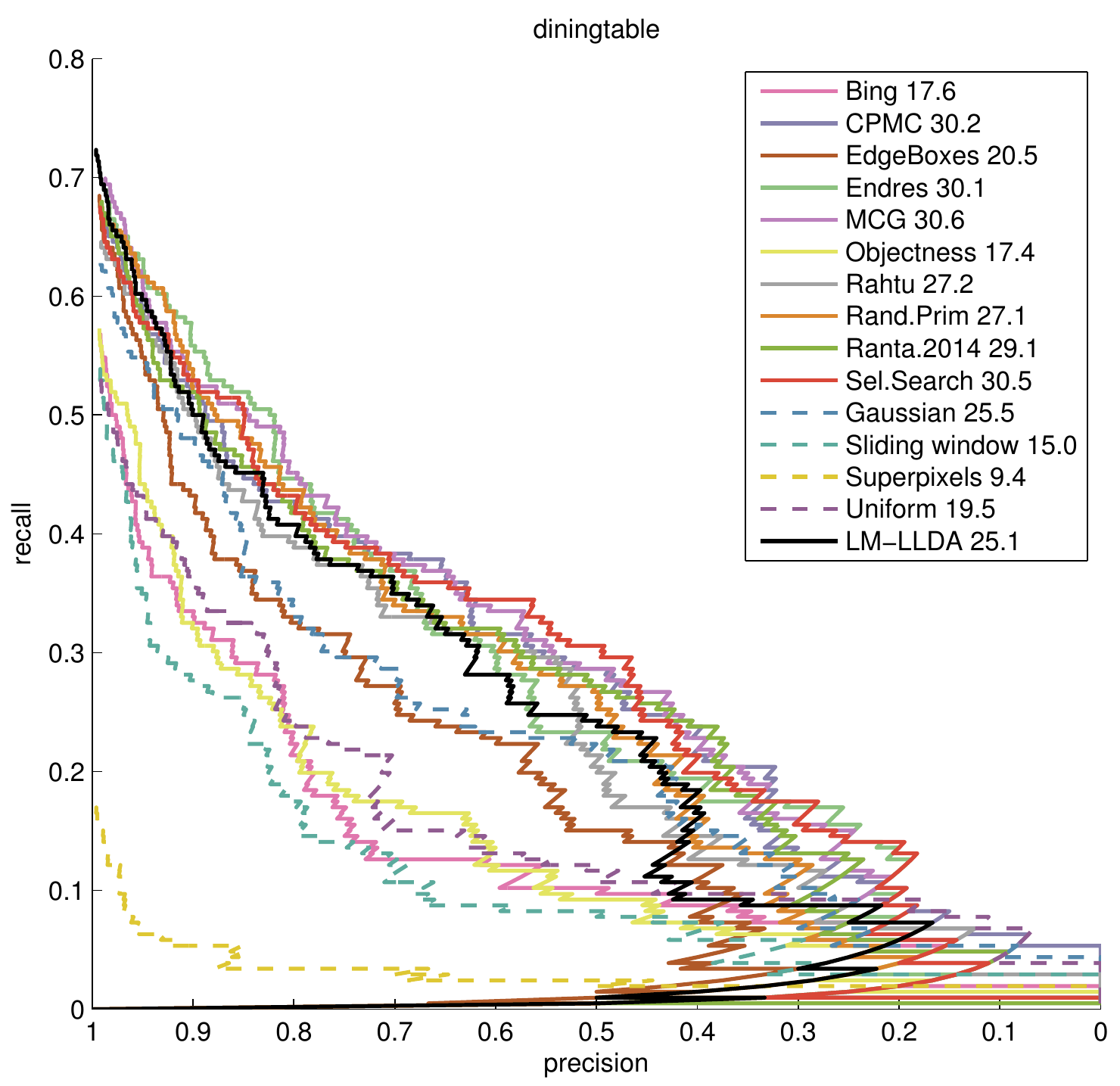}}\hspace*{\fill}\subfloat{\centering{}\includegraphics[width=0.45\textwidth]{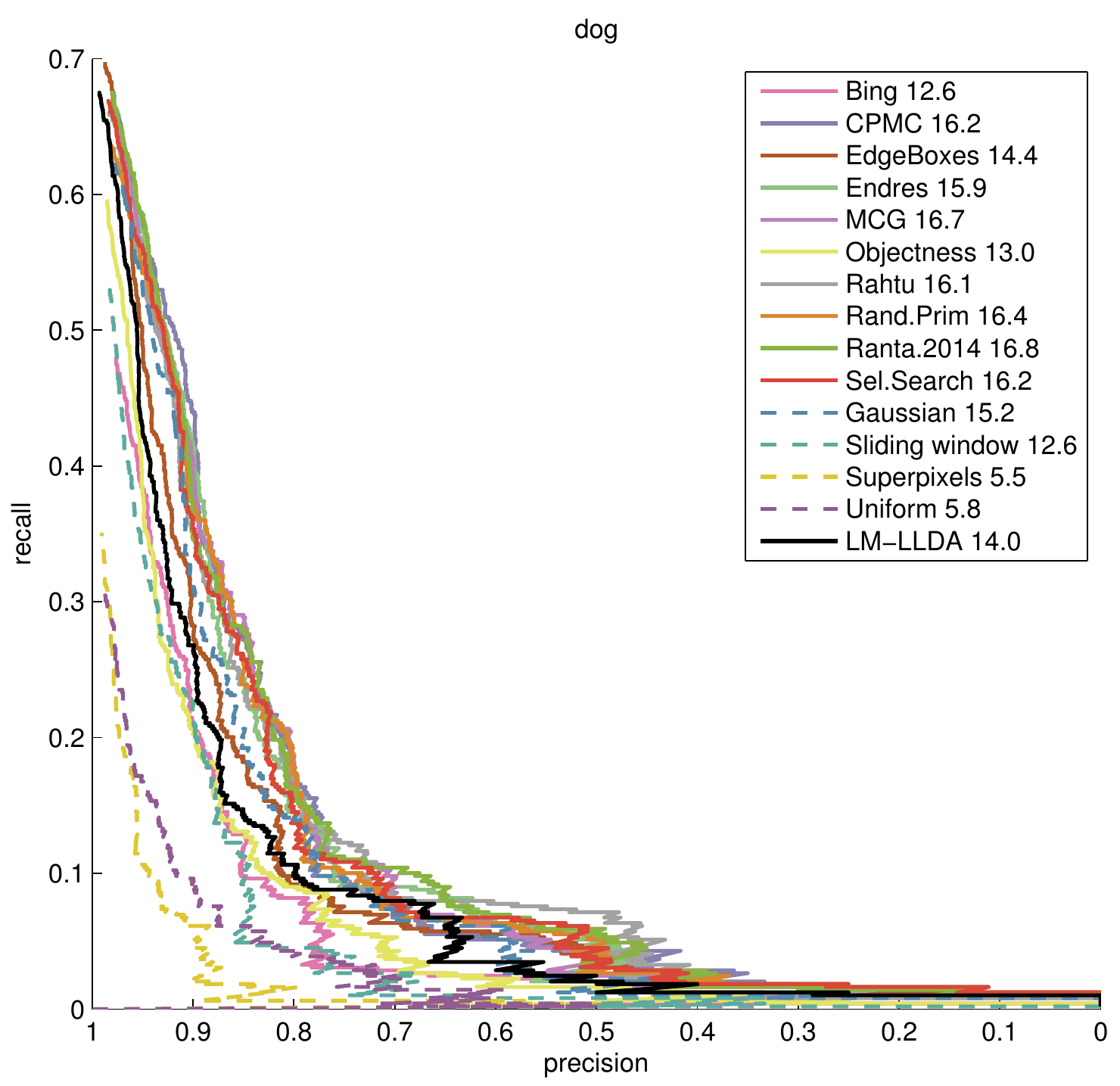}}\hspace*{\fill}
\par\end{centering}

\vspace{0.5em}

\protect\caption{\label{fig:pascal-detection-2-supp}Recall-precision curves for Pascal
VOC 2007 using different proposal methods at test time.}
\end{figure}

\begin{figure}[h]
\begin{centering}
\hspace*{\fill}\subfloat{\centering{}\includegraphics[width=0.45\textwidth]{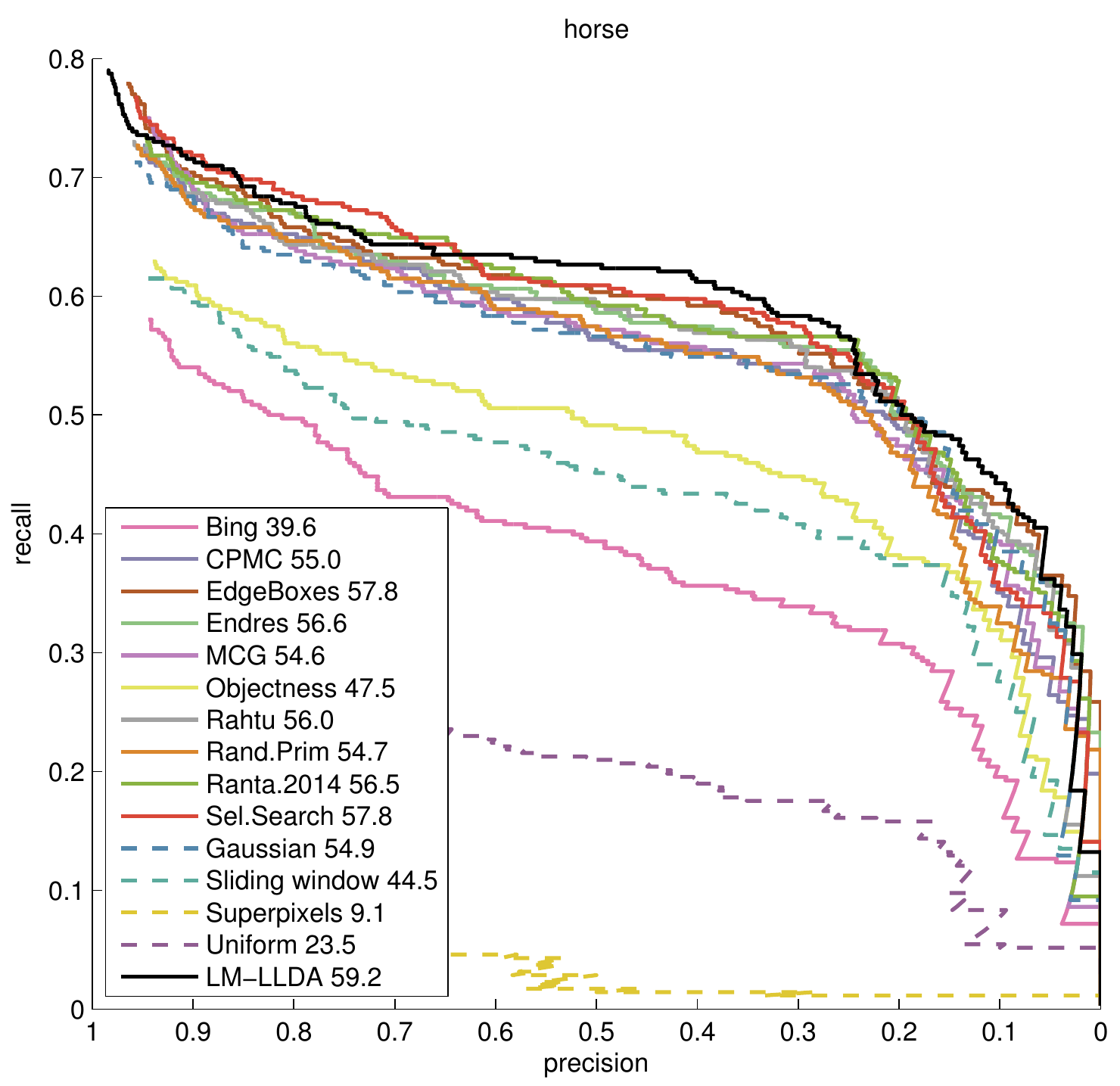}}\hspace*{\fill}\subfloat{\centering{}\includegraphics[width=0.45\textwidth]{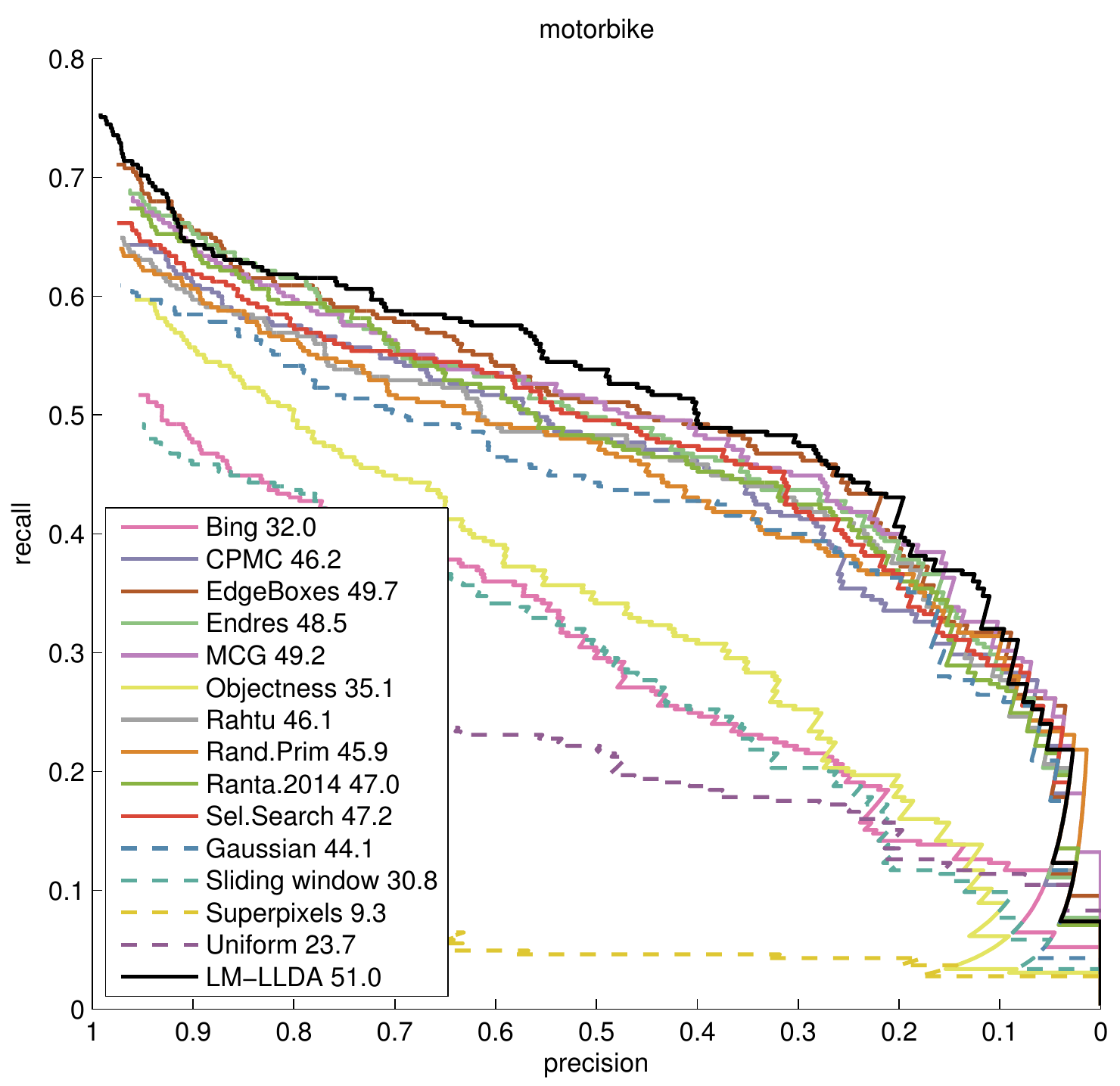}}\hspace*{\fill}
\par\end{centering}

\begin{centering}
\hspace*{\fill}\subfloat{\centering{}\includegraphics[width=0.45\textwidth]{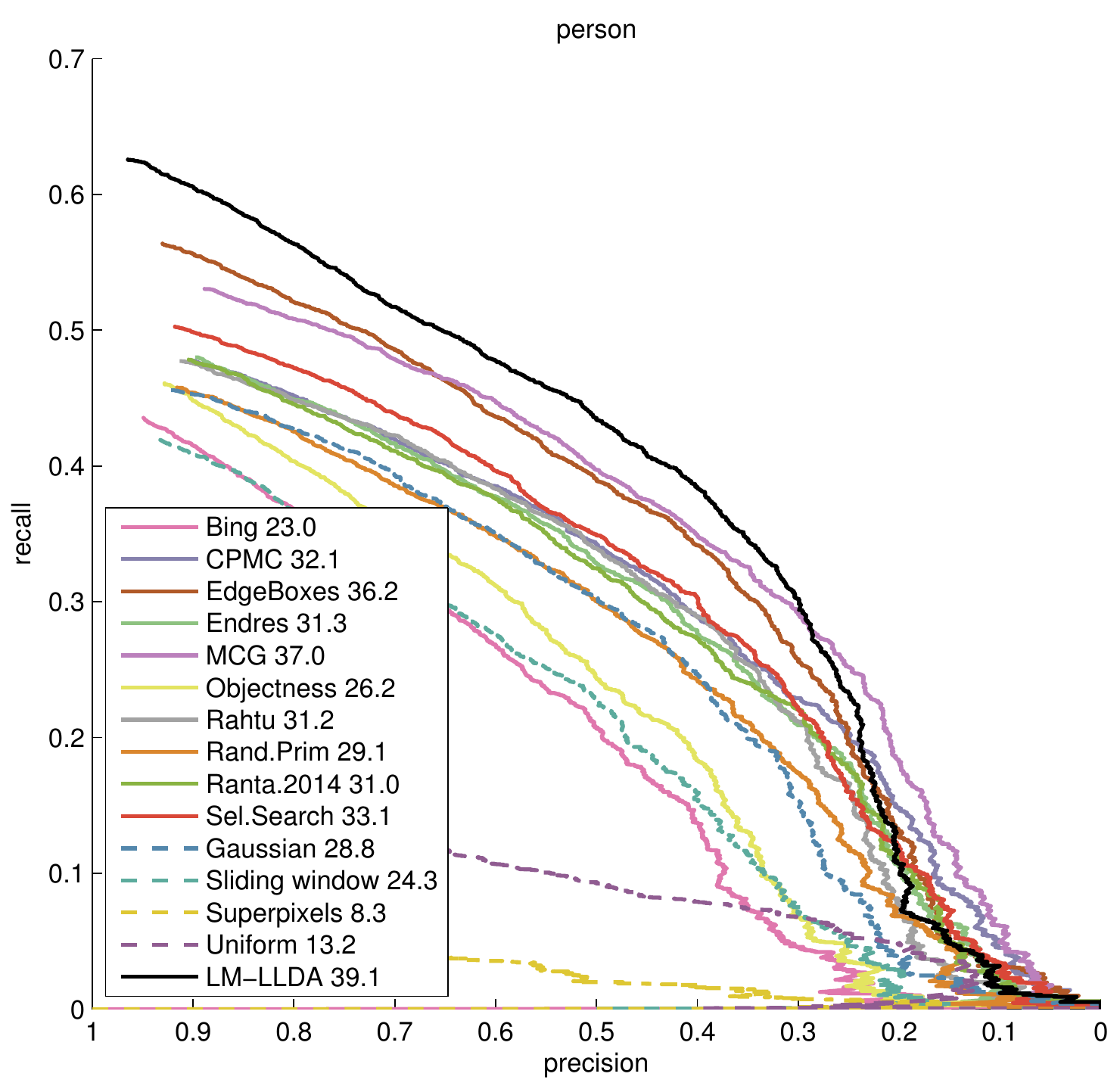}}\hspace*{\fill}\subfloat{\centering{}\includegraphics[width=0.45\textwidth]{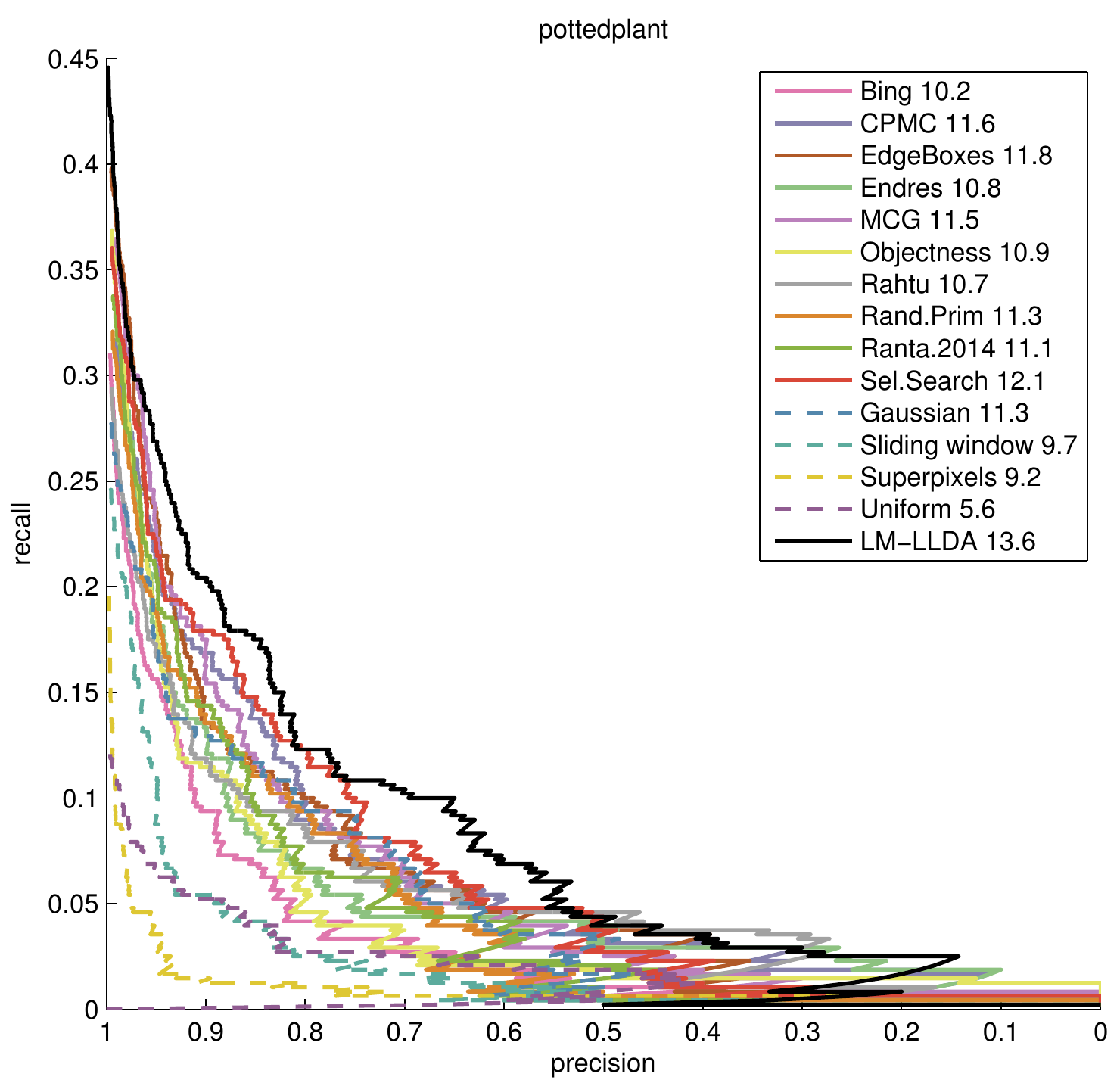}}\hspace*{\fill}
\par\end{centering}

\begin{centering}
\hspace*{\fill}\subfloat{\centering{}\includegraphics[width=0.45\textwidth]{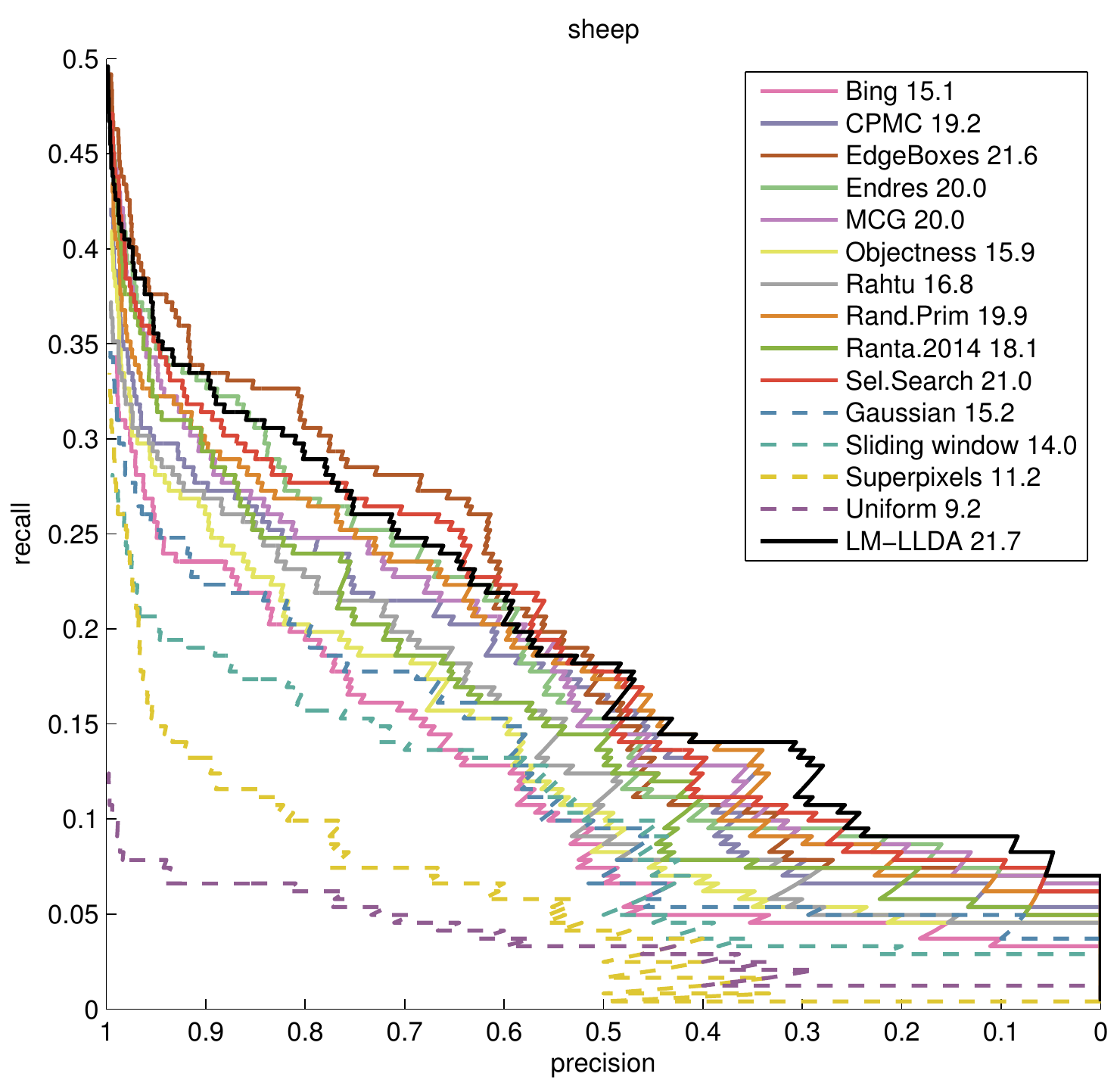}}\hspace*{\fill}\subfloat{\centering{}\includegraphics[width=0.45\textwidth]{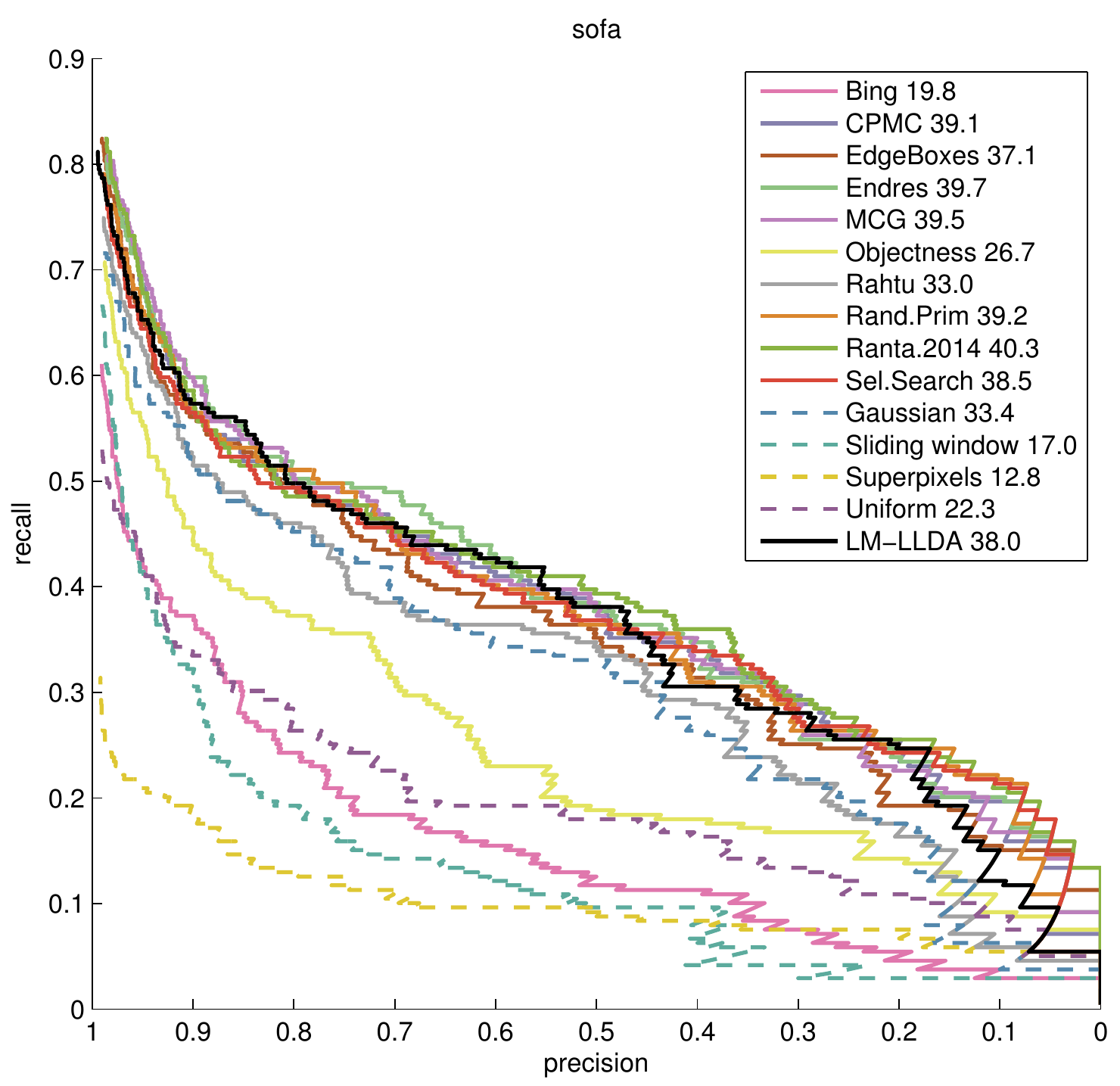}}\hspace*{\fill}
\par\end{centering}

\vspace{0.5em}

\protect\caption{\label{fig:pascal-detection-3-supp}Recall-precision curves for Pascal
VOC 2007 using different proposal methods at test time.}
\end{figure}

\begin{figure}[h]
\begin{centering}
\hspace*{\fill}\subfloat{\centering{}\includegraphics[width=0.45\textwidth]{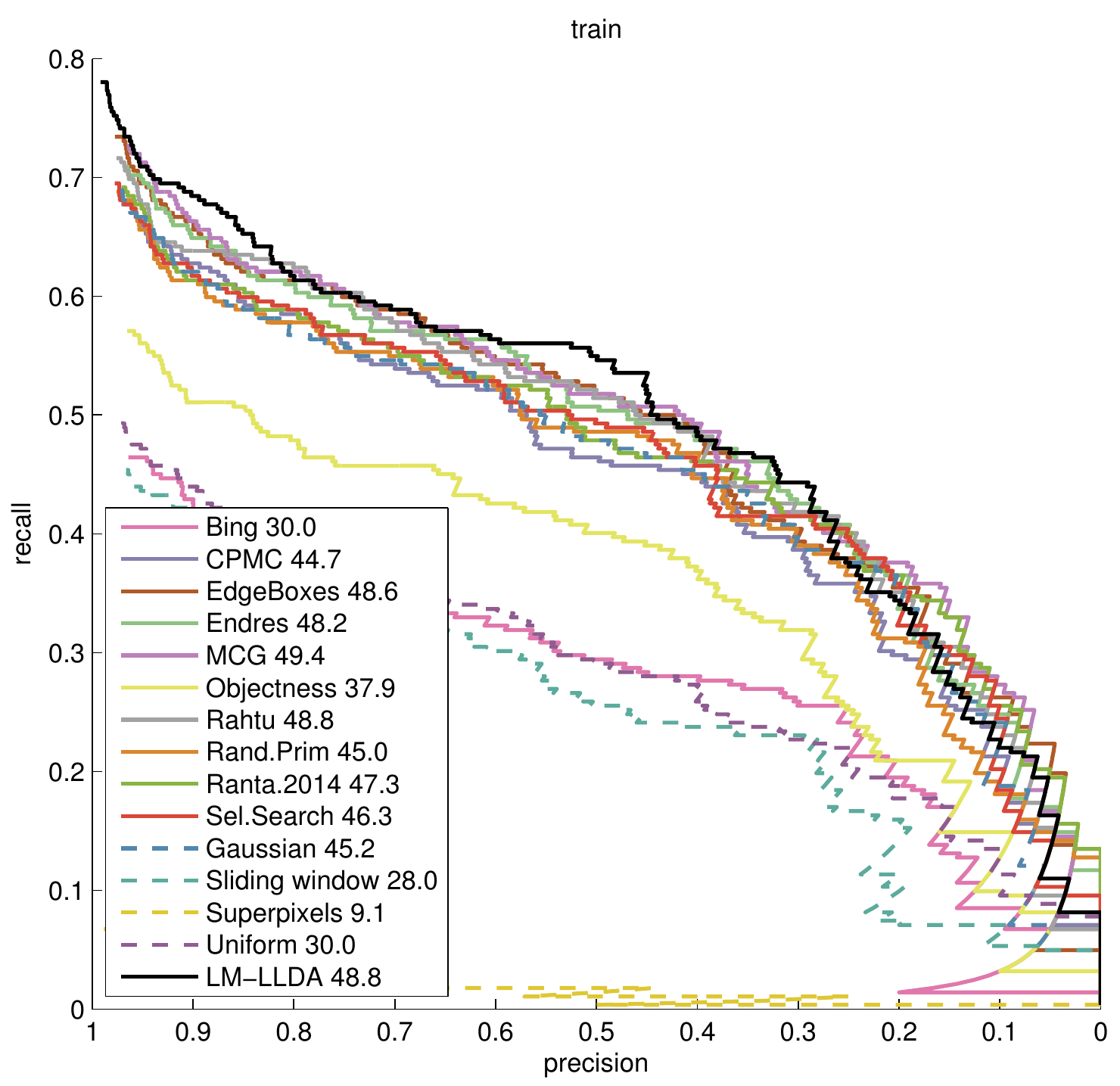}}\hspace*{\fill}\subfloat{\centering{}\includegraphics[width=0.45\textwidth]{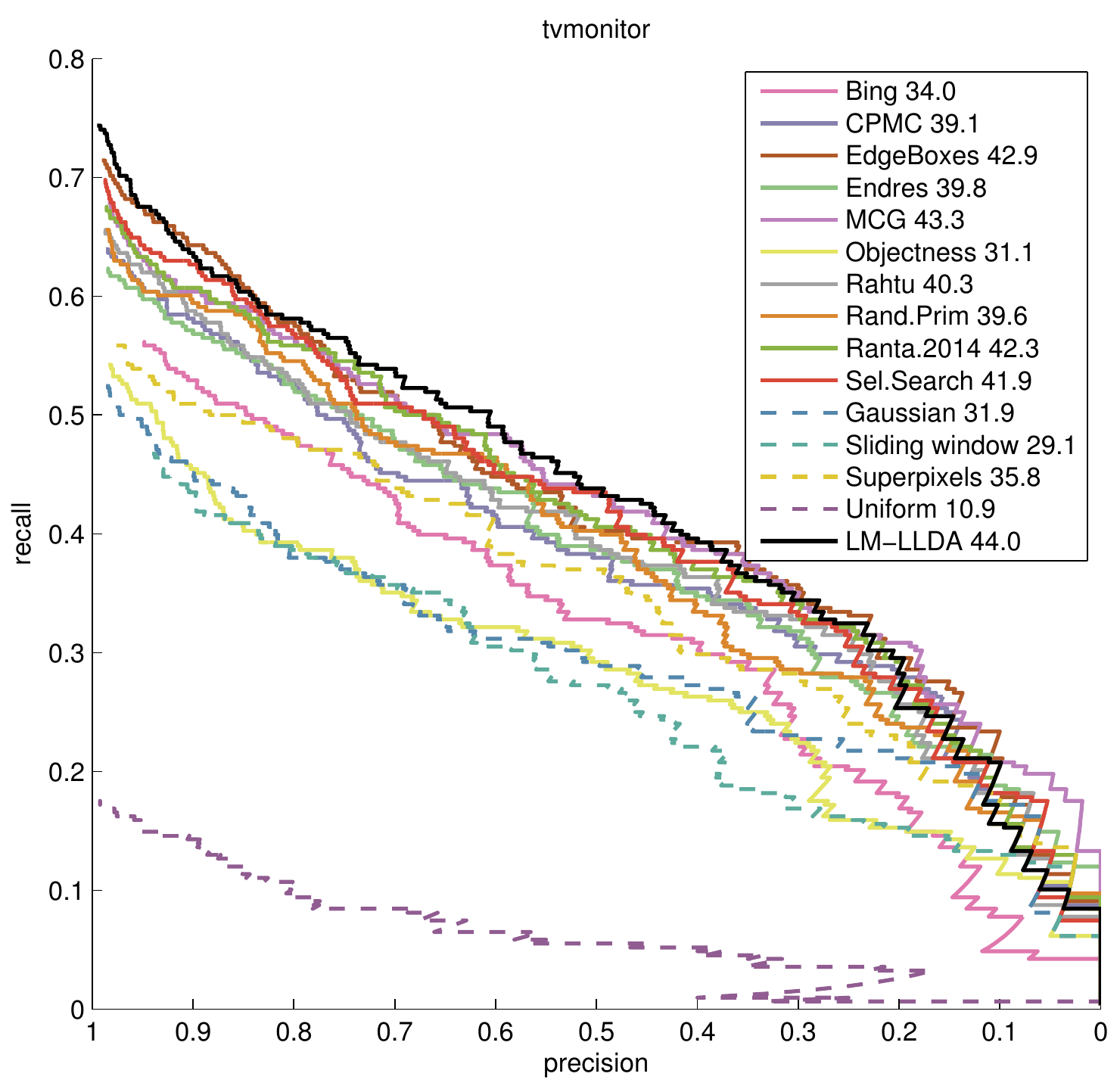}}\hspace*{\fill}
\par\end{centering}

\begin{centering}

\par\end{centering}

\vspace{0.5em}

\protect\caption{\label{fig:pascal-detection-4-supp}Recall-precision curves for Pascal
VOC 2007 using different proposal methods at test time.}
\end{figure}

In the paper, we only report the mean average precision (mAP), i.e.~results
average over all classes. As mentioned in the paper we see different
behaviour for the classes, so it is worth supplying all recall precision
curves. See figures \ref{fig:pascal-detection-1-supp}, \ref{fig:pascal-detection-2-supp},
\ref{fig:pascal-detection-3-supp}, and \ref{fig:pascal-detection-4-supp}.

\end{document}